\documentclass{article}[fontsize=12]
\usepackage{jmlr2e}
\usepackage{graphicx}
\usepackage{graphbox}
\usepackage{subfiles}
\usepackage{float}
\usepackage{cprotect}

\usepackage[letterpaper,top=2cm,bottom=2cm,left=2.5cm,right=2.5cm,marginparwidth=1.2cm]{geometry}
\usepackage{multirow,tabularx}
\usepackage{placeins}
\usepackage{subcaption}

\usepackage{enumitem}
\usepackage{amsmath}
\usepackage{tcolorbox}
\usepackage{amssymb}
\usepackage{algorithm}
\usepackage{mathtools}% http://ctan.org/pkg/mathtools
\usepackage{lipsum}
\usepackage[export]{adjustbox}
\makeatletter
\renewcommand{\slimits@}{\limits}
\renewcommand{\nmlimits@}{\limits}
\makeatother
\usepackage[noend]{algpseudocode}
\usepackage{url}
\usepackage[labelfont=bf]{caption}
\usepackage{xcolor}
\usepackage[bottom]{footmisc}
\newcommand{\mD}{\mathcal{D}}

\title{``What is Different Between These Datasets?" \\
A Framework for Explaining Data Distribution Shifts}

\author{\name Varun Babbar$^*$ \email varun.babbar@duke.edu \\
       \addr Department of Computer Science\\
       Duke University
       \AND
       \name Zhicheng Guo$^*$ \email zhicheng.guo@duke.edu \\
       \addr Department of Electrical and Computer Engineering\\
       Duke University       \AND
       \name Cynthia Rudin \email cynthia@cs.duke.edu\\
       \addr Department of Computer Science\\
       Duke University}

\editor{Mingyuan Zhou}

\usepackage{lastpage}
\jmlrheading{26}{2025}{1-\pageref{LastPage}}{3/24; Revised
2/25}{5/25}{24-0352}{Varun Babbar, Zhicheng Guo, Cynthia Rudin}
\ShortHeadings{title}{Babbar, Guo, and Rudin}

\begin{document}
\maketitle

\maketitle

\begin{abstract}
    The performance of machine learning models relies heavily on the quality of input data, yet real-world applications often face significant data-related challenges. A common issue arises when curating training data or deploying models: two datasets from the same domain may exhibit differing distributions. While many techniques exist for detecting such distribution shifts, there is a lack of comprehensive methods to explain these differences in a human-understandable way beyond opaque quantitative metrics. To bridge this gap, we propose a versatile framework of interpretable methods for comparing datasets. Using a variety of case studies, we demonstrate the effectiveness of our approach across diverse data modalities—including tabular data, text data, images, time-series signals -- in both low and high-dimensional settings. These methods complement existing techniques by providing actionable and interpretable insights to better understand and address distribution shifts.
\end{abstract}

\def\thefootnote{*}\footnotetext{These authors contributed equally to this work.}
\newpage
\tableofcontents
\newpage
\section{Introduction}
\begin{figure}[H]
    \centering
    \includegraphics[width=0.82\textwidth]{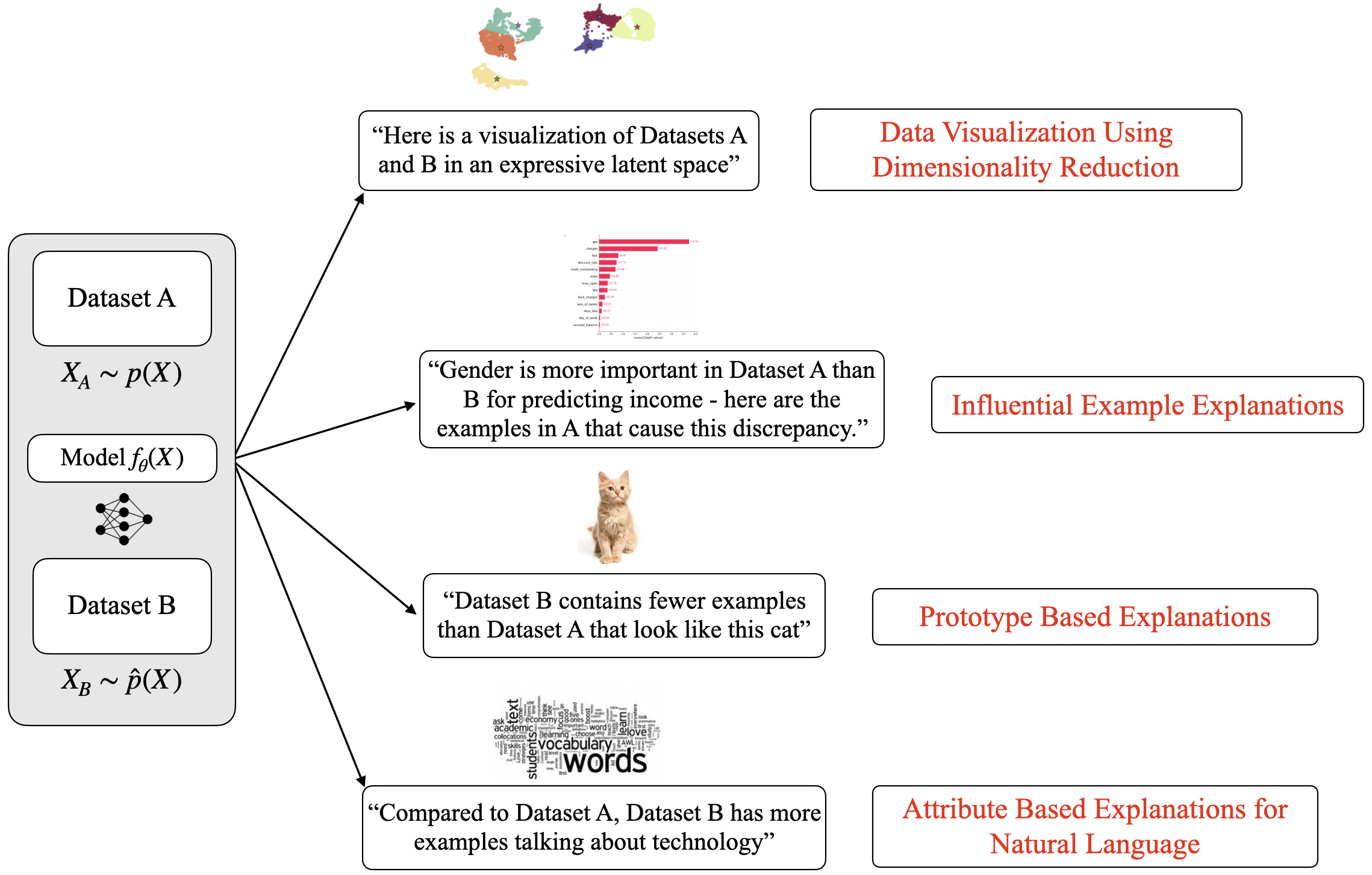}
    \caption{An illustration of our dataset explanation framework. This repertoire of explanations enables the user to gain insights on differences between distribution shifted datasets, with applicability across different modalities. Notably, these explanations do not all require machine learning models trained on the datasets.}
    \label{fig:dataset_explanation_examples}
\end{figure}
Some of the most serious challenges facing the data revolution involves data itself: it is often hard to acquire, hard to share, hard to generate, and hard to troubleshoot. If we generate more data, how do we know it follows the same distribution as our original dataset? If we obtain datasets from different sources, how do we know what is different between them? These questions about data generation and comparisons are important: they arise when we generate medical datasets to protect patient privacy \citep{medical_synthdata1, medical_synthdata2, medical_synthdata3}, generate larger synthetic datasets to augment small true datasets, study data from multiple related sources, or try to determine whether distribution shift has occurred \citep{generalization1, generalization2, generalization3, generalization4}. 
Thus, it is important to be able to understand the \textit{differences between datasets}.

Most previous works in this direction studied distribution shift, focusing on detecting whether or not distribution shift has occurred, as well as detecting differences in statistical features between datasets (e.g., mean, median, and variance, etc.) We claim that knowing whether changes have occurred is not good enough, nor is viewing the data through a few basic statistical measurements such as Wasserstein distance and KL divergence. Understanding the true nature and extent of the changes can help human operators make more informed and effective decisions.

In this work, we propose an explainable AI framework for examining and comparing the differences between two distribution shifted datasets, providing detailed and actionable information. We provide approaches for several data modalities, including high-dimensional complex data, with examples in audio, time series signal, image, and text data. Our framework is summarized in Figure~\ref{fig:dataset_explanation_examples}. It encompasses a variety of explanation types, including prototype explanations (e.g., ``Dataset B contains fewer examples that look like this''), explanations that involve feature importance (``these examples are why feature $K$ is more important for Dataset $A$ than $B$''), and explanations that compare interpretable attributes of natural language datasets. Most explanations are accompanied by visualizations that allow users to examine high-dimensional data and samples. Note that we are not aiming to provide an \textit{exhaustive} list of methods, as there are an infinite number of ways one could examine the difference between two datasets, and sorting through these could easily be overwhelming; instead, we aim for a small set of good methods that will suffice in most cases.

\begin{figure}[H]
    \centering
    \includegraphics[width=0.83\textwidth]{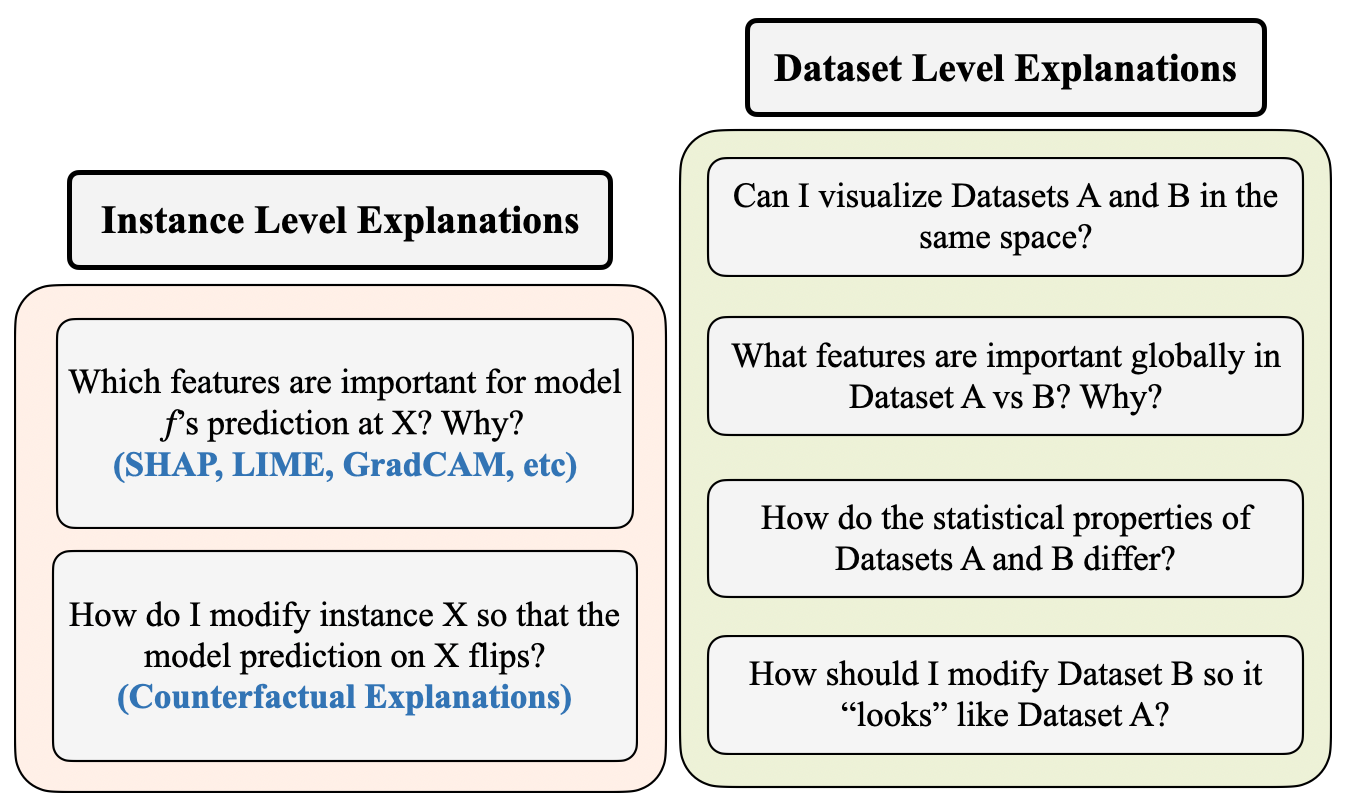}
    \caption{Highlights of the differences between explanations at the instance-level vs$.$ those at the dataset-level.}
    \label{fig:dataset_explanations_vs_instance_explanations}
\end{figure}

In Figure~\ref{fig:dataset_explanations_vs_instance_explanations}, we illustrate the distinction between traditional explainable AI (XAI) approaches and our specific task. Existing XAI methods primarily focus on elucidating the reasoning behind a specific model's decisions on an individual sample basis, as depicted in the left examples of Figure~\ref{fig:dataset_explanations_vs_instance_explanations}. 
However, such methods are inherently tied to model behavior and are not well-suited for explaining dataset-level differences. Since dataset differences do not necessarily depend on a specific model, analyzing them purely through model-based explanations can be limiting. In contrast, our goal is to develop model-agnostic methods that provide dataset-level explanations—capturing systematic differences directly from the data itself — as illustrated in the rightmost examples of Figure~\ref{fig:dataset_explanations_vs_instance_explanations}.

\section{Related Works}

We introduce and discuss several related previous works to this study.

\subsection{Distribution Shift}
% similar to https://arxiv.org/pdf/2210.10275.pdf
Our study is adjacent to distribution shift analysis, though our focus is broader: we do not focus on any specific type of distribution shift (such as covariant shift, \citealt{sugiyama2007covariate}, or label shift, \citealt{zhang2013domain}), rather we focus on the changes between datasets with no particular assumptions on the type of shift. 
Previous efforts have largely focused on the detection and analysis of the shifts \citep{distshift_detect1, distshift_detect2, distshift_detect3}, and the improvement of model generalizability to alleviate the effects of distribution shift \citep{distshift_generalize1, distshift_detect2, distshift_generalize3, distshift_generalize4}. However, to the best of our knowledge, most works have not explored explaining distribution shifts in a human-understandable manner.
The closest work to ours from this literature is possibly that of \citet{whymodelfail}, who proposed an approach to attribute model performance changes due to distribution shifts based on Shapley values \citep{shapley1953value}. We focus more broadly on explaining differences between datasets, with no requirements of prior knowledge or a task-related model.

\subsection{Instance-level Explanations}
The conventional instance-level explanation literature has largely focused on post-hoc analysis, i.e., analysing a prediction from a trained model. Some well-known work \citep{lime, anchors} has focused on learning simpler explanation functions that approximate the model around the neighbourhood of a point. The output of these functions is a score for each feature representing its contribution to a given prediction. The feature importance-based explanation literature has also examined methods that compute the gradient of the prediction with respect to the input \citep{shap,gradcam,saliency,smoothgrad,axiom}. Another line of research focuses on counterfactual explanations \citep{mace,recourse,clue,recourse_hima}, which provide changes to a given instance so that the model flips its prediction (or, in the case of \citet{clue}, becomes certain of its prediction). As far as we can tell, none of these types of approaches can be applied to explaining the difference between two datasets; instead, they all explain a model.
\subsection{Dataset-level Explanations}
To the best of our knowledge, there is very limited literature on dataset-level explanations. The most relevant work on dataset-level explanations is that of \citet{towards_explain}, which uses optimal transport \citep{OT} maps to explain mean shifts in distributions of the datasets (or individual clusters). The user is provided with the original clusters and the transported clusters and can visually inspect the difference between the two to derive insights. However, their method focuses exclusively on mean shifts between clusters and requires both datasets to be of the same size, which can be limiting (see Section~\ref{sec:failure} in the appendix for an example). \citet{page} provides dataset-level explanations for graph classification tasks by comparing examples in a dataset to salient sub-graph prototypes frequently observed in the dataset. \citet{gsclip} introduce natural language explanations for visual datasets. In particular, for each attribute or class in the dataset, the explanation consists of the $K$ most salient image samples in dataset $\mD$, their shifted versions in dataset $\mD^\prime$, and a natural language description of their differences. This method depends on having a 1-to-1 correspondence between items in $\mD$ and $\mD'$ that are not usually available.

For textual data, \citet{elazar2023whats} explores properties of several large-scale text corpora to uncover insights on the relative presence of attributes such as toxicity, level of contamination, and n-gram statistics.
\subsection{Synthetic Data}
One major application of our work lies in explaining the difference between real data and synthetic datasets. Most frameworks for evaluating real and synthetic data focus on the statistical properties of the datasets or evaluate the quality of the generative models. \cite{Livieris2024} construct an evaluative framework for synthetic data generating models, providing metrics that quantify statistical faithfulness. However, while these metrics are very useful, they only provide a limited picture of the synthetic dataset. \cite{neto2024massivelyannotateddatasetsassessment}, on the other hand, create annotated attributes for known real and synthetic face recognition datasets and compare the data along these attributes. While this emulates our philosophy of providing interpretable dataset explanations, their main findings are specific to the domain of face recognition (including the attributes they picked) and may not be directly applicable to other kinds of datasets. In this work, we show that our approaches are general enough to uncover underlying intricacies of synthetic data that distinguish it from real data, such as the quality of cluster substructures and properties of influential groups of datapoints.

\subsection{Prototype Learning}
In recent years, ProtoPNet \citep{protopnet}, a type of prototype network, has been introduced as an inherently interpretable neural network capable of providing explanations through case-based reasoning for its predictions. Specifically, ProtoPNet has been built into a popular framework where images are classified by comparing specific parts of an image to prototypical parts associated with each class. Subsequent developments have expanded on the original ProtoPNet algorithm \citep{protopnet}, focusing primarily on enhancing the components of ProtoPNet itself \citep{donnelly2022deformable, nauta2021neural, rymarczyk2022interpretable, rymarczyk2021protopshare, wang2023learning, ma2024looks, wang2021interpretable, nauta2021looks}, refining the training regimen \citep{rymarczyk2023icicle, nauta2023pip, willard2024looks}, or adapting ProtoPNets for high-stakes applications \citep{yang2024fpn, barnett2021case, choukali2024pseudo, wei2024mprotonet, interpEEG}. Although we utilize the underlying prototype learning mechanism, our focus differs significantly from traditional applications in the prototype learning literature. These methods have the ultimate goal of performing classification given the underlying task. Our proposed approach aims to compare distribution shifted datasets, and can operate on both labeled and unlabeled datasets. This fundamental shift highlights one of the unique challenges and goals of our methodology.
    
\subsection{Rashomon Effect}
\label{sec:rashomon_effect}
Our research leverages the Rashomon Effect to evaluate feature importance. This phenomenon is the existence of multiple, diverse models that achieve similar predictive performance for the same task \citep{breiman2001statistical}. The Rashomon Effect presents both challenges and opportunities: it gives rise to predictive multiplicity -- where different models yield varying predictions for the same instance \citep{marx2020predictive, watson2023multi, kulynych2023arbitrary, hsu2022rashomon, watson2023predictive}. Also, the set of high-quality models for a given dataset can disagree on which variables are important \citep{fisher2019all,dong2019variable,smith2020model}. Interestingly, compiling all these good models yields something better than what can be obtained with any single model -- a robust, model-agnostic method for assessing variable importance, called the Rashomon Importance Distribution \citep{donnelly2023rashomon}. In this study, we focus on the Rashomon set, the collection of highest-performing models, to construct reliable and robust feature importance measures that help distinguish between datasets.

\section{The Dataset Explanation Framework}
\label{sec:dataset_exp_framework}
\subsection{Overview of Methodology}
In this paper, we aim to illuminate the differences between distribution shifted datasets $\mD$ and $\mD^\prime$ consisting of features $X$ and possibly labels $Y$, i.e., $\mD = \{(X_i,Y_i)\}_{i=1}^N$ and $\mD^\prime = \{(X^\prime_i,Y^\prime_i)\}_{i=1}^{N^\prime}$. $Y$ is not always required - our framework contains methods to deal with both supervised and unsupervised data. Our primary assumption in this work is that $\mD$ and $\mD^\prime$ belong to the same domain (e.g., both consist of animal images), but other properties of the datasets and their corresponding task models may differ, such as feature and class distributions, (latent) cluster structure, and model performance metrics. While these aspects of datasets are relatively easy to capture, what is not trivial is producing actionable insights into dataset differences. For instance, using our explanation framework, we can reveal that $\mD^\prime$ lacks examples of a certain archetype that are more prevalent in $\mD$, how structural properties of the datasets differ, and certain intrinsic biases in either dataset that cause differing feature importances between $\mD$ and $\mD^\prime$.

\begin{figure}[H]
    \centering
    \includegraphics[width=\textwidth]{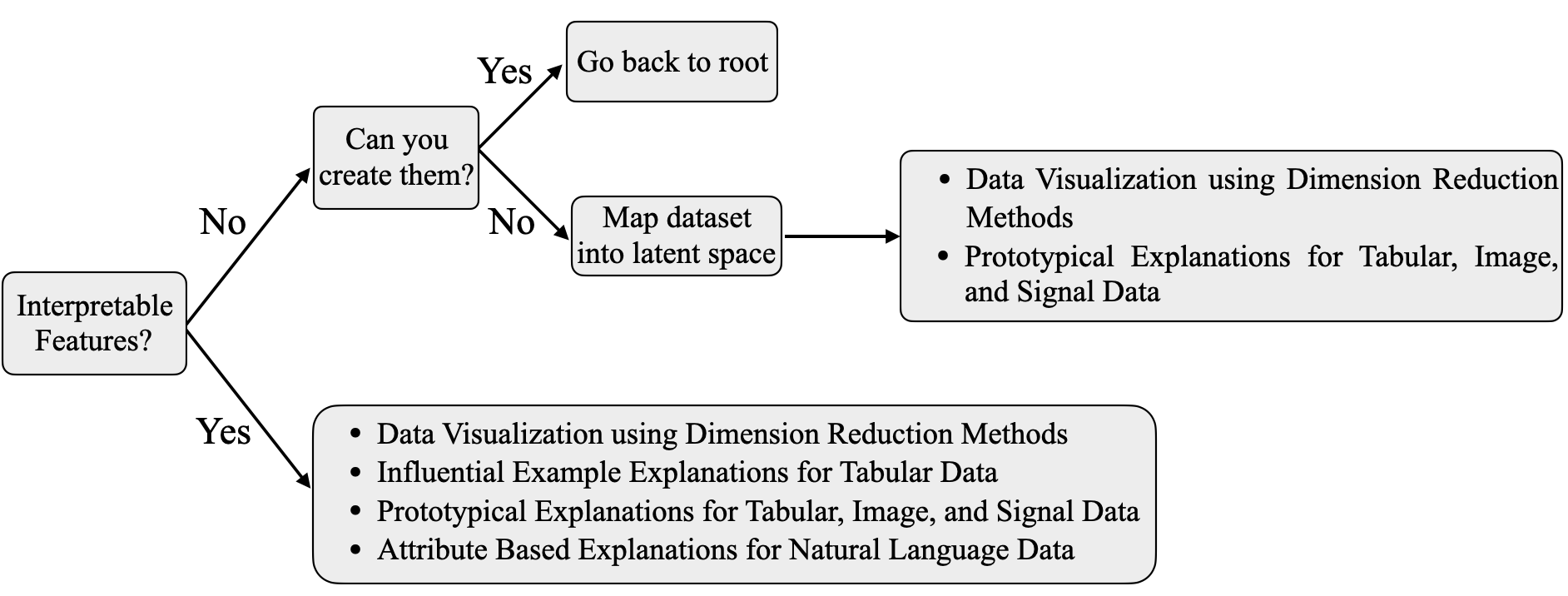}
    \caption{Pipeline for our explanation framework. We provide explanation methods that are applicable across many data modalities. Some of our methods leverage interpretable features and explain dataset differences in terms of those features. When the features are uninterpretable (e.g., individual tokens in natural language), one could potentially create proxy attributes that are interpretable and explain the datasets in terms of those attributes, or use prototypical explanations and dimension reduction projections. We use PacMAP \cite{pacmap} as the dimension reduction method in this paper, as it currently offers state of the art performance.
    }
    \label{fig:dataset_explanation_tree}
\end{figure}

Our pipeline for exploring the differences between datasets is illustrated in Figure~\ref{fig:dataset_explanation_tree}, which includes dimension reduction for data visualization, as well as three novel algorithms:
\begin{itemize}
    \item Influential example-based explanations are discussed in Section~\ref{sec:meth:featimp}. These explanations help uncover subgroups in datasets $\mD$ and $\mD^\prime$ that cause differences in their feature importances. Applicable for supervised data.
    \item Prototype-based explanations. See Sections~\ref{sec:protoneigh_explainmethod} and \ref{sec:summarizationprotoexp}. These explanations help compare local neighborhoods between datasets $\mD$ and $\mD^\prime$. They are accompanied by comparative visualizations using dimensionally reduction methods. They can be used for both supervised and unsupervised data.
    \item Large Language Model (LLM)-based explanations using interpretable attributes, see Section~\ref{sec:nlpattributemethod}. Can be used to compare text corpora.
\end{itemize}

The first two explanations involve generating, analysing, and comparing salient samples and their features in either dataset. The final explanation involves creating interpretable attributes for each dataset and examining the dataset in terms of those attributes.

\subsection{Overview of Paper Structure}

Different data modalities and tasks require different types of explanations. For instance, using influential example-based explanations is appropriate for tabular data, where the feature values are interpretable. However, for image and signal data with non-interpretable features (e.g., a pixel value or a signal value at time $t$), feature importance would not be as interpretable for humans. In the following subsections, we describe each method in our framework and provide several supporting case studies of different tasks and data modalities utilizing the proposed approaches. 
\subsection{Explanations Based on Influential Examples}\label{sec:meth:featimp}
\subsubsection{Introduction}
\begin{figure}[H]
    \centering
    \includegraphics[width=\textwidth]{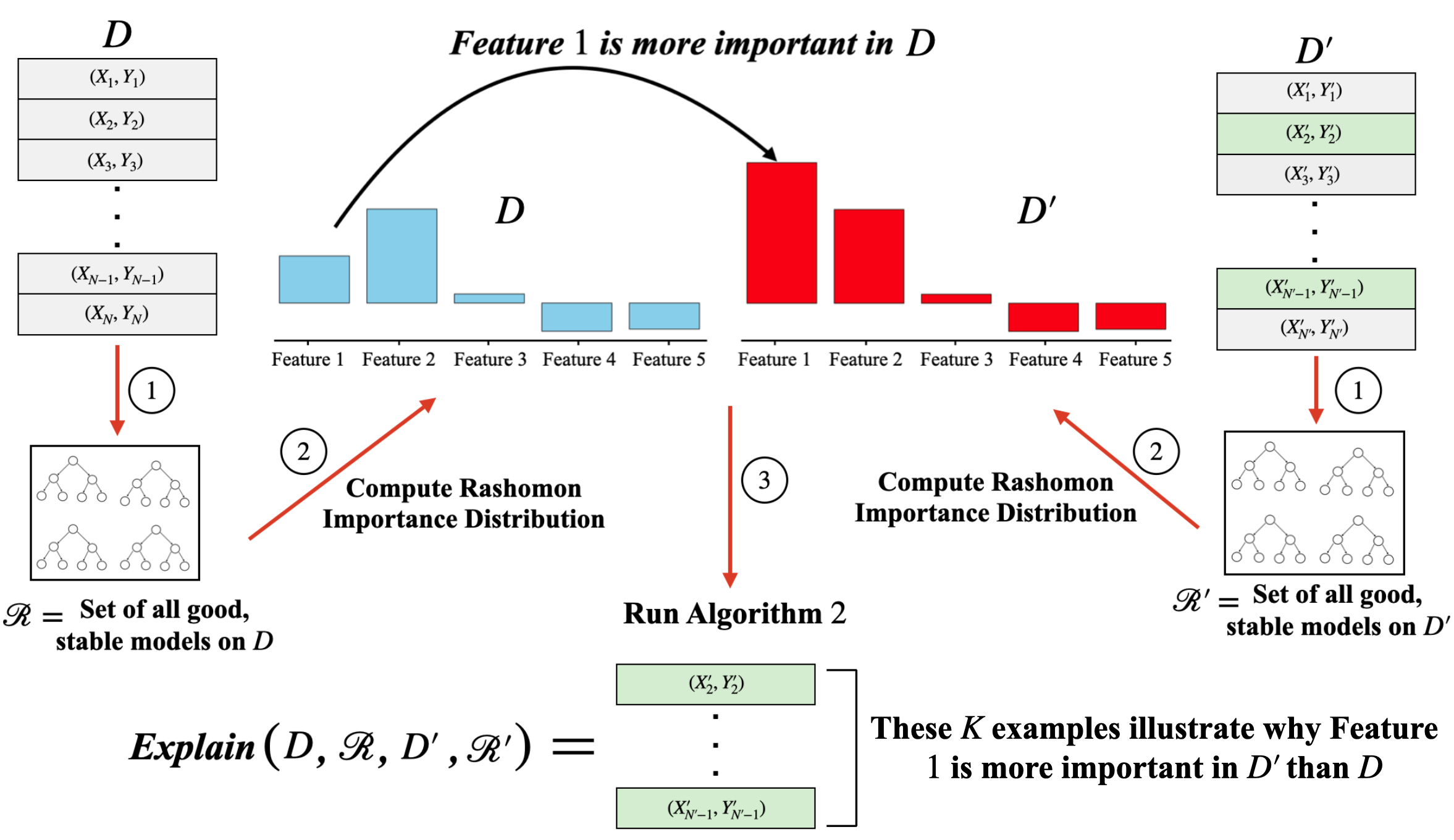}
    \caption{An illustration of influential example explanations. Given datasets $\mD$ and $\mD^\prime$ and a feature importance metric, our explanation gives us $K$ influential examples (the user can choose whether they are from $\mD$ or $\mD^\prime$) that are most responsible for the feature importances being different between $\mD$ and $\mD^\prime$. These feature importances are computed from the set of all nearly optimal, stable decision trees (where stability means that the model is also nearly optimal for perturbations of $\mD$) -- we show how to compute these below. A practitioner can uncover specific patterns that distinguish these $K$ examples -- we illustrate this in Sections~\ref{sec:fi_exp_heloc} and \ref{sec:fi_exp_adult}.}
    \label{fig:fi_explanations_visualization}
\end{figure}
\label{sec:featureimportancemethod}

This section examines explanations that take into account differences between datasets by considering \textit{which features are intrinsically important in both datasets relative to the underlying task}. An intrinsically important feature is one whose importance for the underlying data distribution remains stable across multiple well-trained models and perturbations of the dataset. \citet{donnelly2023rashomon} show that not considering this model-agnostic representation of feature importance can cause researchers to arrive at multiple equally valid -- yet contradictory -- conclusions about the data.  After determining the intrinsic importance of features, we ask the question: \textit{Given Datasets $\mD$ and $\mD^\prime$, which $K$ examples from Dataset $\mD^\prime$ should I remove so that the intrinsic importance of features in both datasets for the underlying task are as similar as possible?} To the best of our knowledge, this is a novel way of looking at two datasets while taking into account an underlying task (e.g., classification). To determine intrinsic feature importances for a labeled dataset, we employ the Rashomon Importance Distribution (RID) framework of \citet{donnelly2023rashomon}. \cite{donnelly2023rashomon} show that this method, which quantifies the importance of a feature across the set of all good models in a class, results in feature importances that are highly robust to dataset perturbations. Given a dataset $\mD$, a hypothesis class $\mathcal{F}$, regularization strength $\lambda$, and tolerance $\epsilon$, the Rashomon set $\mathcal{R}$ is defined as the set of all models in $\mathcal{F}$ whose empirical losses are within $\epsilon$ of the minimum empirical loss \citep{rashomon_set}:
\begin{equation}
    \mathcal{R}(\mD, \epsilon, \mathcal{F},\lambda) = \{f \in \mathcal{F}: \ell(f,\mD,\lambda) \leq \min_{f^\prime \in \mathcal{F}} \ell(f^\prime, \mD, \lambda) + \epsilon \}, \textrm{ where}
\end{equation}
\begin{equation}
    \ell(f, \mD, \lambda) = \frac{1}{|\mD|} \sum_{Z=(X,Y) \in \mD} L(Z, f) + \lambda R(f) 
\end{equation}
is the regularized empirical loss for the dataset, with loss function $L$ and model sparsity $R(f)$ (e.g. the number of leaves in a decision tree). Our intrinsic variable importance will average a variable importance metric over Rashomon sets constructed on bootstrap samples.

\subsubsection{Definitions of Relevant Importance Measures}
Before we introduce the method to compute the intrinsic feature importances, we first define the following terms:
\begin{definition}[\textbf{Local Feature Importance Measure  -- LFIM}]
\label{def:lfim}
    Given a predictor $f$ from a hypothesis class $\mathcal{F}$ and a dataset $\mD$ with $M$ features, a local feature importance measure is a function $\phi(f, X, Y): \mathcal{F} \times \mathcal{X} \times \mathcal{Y} \rightarrow \mathbb{R}^{M}$ that outputs a vector representing the relative contribution of each feature towards the output prediction $Y$ for a specific input $X$. A lot of work has been devoted to the development of faithful feature importance measures \citep{lime,shap,donnelly2023rashomon} -- in principle, any of these can be used in our explanation framework. We assume that this feature importance measure is a property of the dataset and the model in question.
\end{definition}
\begin{definition}[\textbf{Global Feature Importance Measure -- GFIM}]
\label{def:gfim}
     Given a predictor $f$ from a hypothesis class $\mathcal{F}$ and a dataset $\mD$ with $M$ features, a global feature importance measure $\phi_g(f, \mathcal{D}): \mathcal{F} \times \mathcal{D} \rightarrow \mathbb{R}^{M}$ will provide a similar vector as an LFIM, except that it represents the predictive power of each feature in the entire dataset.
    In this paper, we consider GFIM to be the average LFIM vector across all examples in a dataset, i.e., $\phi_g(f,\mathcal{D}) = \mathbb{E}_{(X,Y) \in \mD}[\phi(f,X,Y)]$.
\end{definition}

\begin{definition}[\textbf{Local Intrinsic Feature Importance Measure  -- LiFIM}]
\label{def:lifim}
    Given a dataset $\mD$ with $M$ features, a local intrinsic feature importance measure $\phi(X,Y,\mathcal{D}): \mathcal{X} \times \mathcal{Y} \times \mathcal{D} \rightarrow \mathbb{R}^M$ for an example $(X,Y) \in \mD$ computes the importance of each feature in $(X,Y)$ by aggregating the LFIMs of well-trained, stable models in $\mathcal{F}$. This involves computing Rashomon sets of bootstrapped samples from $\mD$, storing models associated with each set and aggregating their LFIMs. The precise technique is detailed below in this section. 
\end{definition}
\begin{definition}[\textbf{Global Intrinsic Feature Importance Measure -- GiFIM}]
\label{def:gifim}
      Given a dataset $\mD$ with $M$ features, a global feature importance measure $\phi_g(\mathcal{D}): \mathcal{D} \rightarrow \mathbb{R}^{M}$ will provide a similar vector as an LiFIM, except that it represents a holistic summary of the intrinsic predictive power of each feature across an entire dataset. In this paper, we consider GiFIM to be the average of LiFIMs across all examples in a dataset, i.e., $\phi_g(\mathcal{D}) = \mathbb{E}_{(X,Y) \in \mD}[\phi(X,Y,\mathcal{D})]$.
\end{definition}

Under the framework of \citet{donnelly2023rashomon}, we can compute the LiFIM and GiFIM of models in the following manner:
\begin{itemize}
    \item Bootstrap the dataset $\mD$ $B$ times. 
    \item For each bootstrapped dataset $\mD_{i}$, compute its Rashomon set $\mathcal{R}(\mD_i, \epsilon, \mathcal{F},\lambda)$. For decision trees, this can be done using TreeFARMS \citep{treefarms}.
    \item Compute the LFIMs of each example under each model in each Rashomon set using any method in literature  \citep[here, we use SHAP of][]{shap}. Under computational constraints, a random sample of models from each Rashomon set can also be used.
    \item The LiFIM $\phi(X,Y,\mD)$ for an example $(X,Y) \in \mD$ is computed by taking the mean (over bootstraps and Rashomon sets) of feature importances. That is: 
    \begin{equation}
        \phi(X,Y,\mD) = \frac{1}{B}\sum_{i=1}^B \frac{1}{|\mathcal{R}(\mD_i,\epsilon,\mathcal{F},\lambda)|}\sum_{f \in \mathcal{R}(\mD_i,\epsilon,\mathcal{F},\lambda)} \phi(f, X,Y).
    \end{equation}
    If a model appears more than once across different Rashomon sets, this results in that model's feature importance vector having a larger contribution to the final LiFIM. 
    \item The GiFIM $\phi_g(\mD)$ for the dataset $\mD$ is the average of LiFIMs in the dataset, i.e.\\ $\phi_g(\mD) = \mathbb{E}_{(X,Y) \in \mathcal{\mD}}[\phi(X,Y, \mD)]$.
\end{itemize}

In this paper, given $\mD$ and $\mD^\prime$, the influential example explanation provides the following information to the user: \textit{A set of $K$ examples (from either $\mD$ or $\mD^\prime$ that), if removed from the dataset, would align the GiFIMs of $\mD$ and $\mD^\prime$ the most.} Concretely, let $\phi_g(\mD)$ and $\phi_g({\mD^\prime})$ be the GiFIMs on $\mD$ and $\mD^\prime$. We aim to find the set $S$ of $K$ examples $S = \{(X_{[1]}, Y_{[1]}), .. (X_{[K]}, Y_{[K]})\}$ in $\mD^\prime$ such that $d\big(\phi_g(\mD), \phi_g(\mD^\prime \backslash S)\big)$ is minimized, where $d(.,.)$ is the Euclidean distance metric between two vectors. That is, $\mD$ and $\mD^\prime \backslash S$ will have more aligned intrinsic feature importances. Figure \ref{fig:fi_explanations_visualization} illustrated the underlying intuition behind these explanations. We now explain how we obtain these $K$ influential examples. 

\subsubsection{Determining the Influential Examples}
In order to provide influential example explanations, we first define the notion of \textit{influence} for a test loss function. 

\begin{definition}[Influence Function for Test Loss \citep{influence}]
    Given the following: 
    \begin{itemize}
        \item training and test datasets $\mathcal{D}_{\textrm{train}} = \{Z^{\textrm{train}}_i = (X^{\textrm{train}}_i,Y^{\textrm{train}}_i)\}_{i=1}^{N_{\textrm{train}}}$, and \\ $\mathcal{D}_{\textrm{test}} = \{Z^{\textrm{test}}_i = (X^{\textrm{test}}_i,Y^{\textrm{test}}_i)\}_{i=1}^{N_{\textrm{test}}}$,
        \item a trained, parameterized model $m_{\theta}(x)$,
          \item the minimizer of the training loss: $\hat{\theta} = \textrm{argmin}_\theta \frac{1}{N_{\textrm{train}}}\sum_{i=1}^{N_{\textrm{train}}}L(Z^{\textrm{train}}_i,m_{\theta})$,
        \item the empirical test loss $L_{\textrm{test}}(m_{\hat{\theta}}) = \frac{1}{N_{\textrm{test}}}\sum_{i=1}^{N_{\textrm{test}}}L(Z^{\textrm{test}}_i,m_{\hat{\theta}})$, 
    \end{itemize}
    an influence function for training point $(X^{\textrm{train}}_i,Y^{\textrm{train}}_i)$ estimates the theoretical change in the test loss $L_{\textrm{test}}(m_{\hat{\theta}})$ if the model $m_\theta$ is trained using $\mathcal{D}_{\textrm{train}} \backslash (X^{\textrm{train}}_i,Y^{\textrm{train}}_i)$. By applying techniques from \citet{influence}, we can write this as: 
    \begin{equation}
        I(Z^{\textrm{train}}_i,\mathcal{D}_{\textrm{test}}, m) = \sum_{j=1}^{N_{\textrm{test}}} \frac{1}{N_{\textrm{test}}}\nabla_\theta L(Z^{\textrm{test}}_j,m_{\hat{\theta}})^TH_{\hat{\theta}}^{-1}\nabla_\theta L(Z^{\textrm{train}}_i,m_{\hat{\theta}}).
        \label{eqn:influence}
    \end{equation}
    where $H_{\hat{\theta}}$ is the Hessian of the parameters $\theta$ evaluated at $\theta = \hat{\theta}$. This is essentially an approximation of the following form:
    \begin{equation}
        I(Z^{\textrm{train}}_j,\mathcal{D}_{\textrm{test}}, m) \approx L_{\textrm{test}}(m_{\hat{\theta}}) - L_{\textrm{test}}(m_{\hat{\theta}_{-Z^{\textrm{train}}_j}})
    \end{equation}
    where 
    \begin{equation}
        \hat{\theta}_{-Z^{\textrm{train}}_j} = \textrm{argmin}_\theta \Big(\Big(\frac{1}{N_{\textrm{train}}}\sum_{i=1}^{N_{\textrm{train}}}L(Z^{\textrm{train}}_i,m_\theta)\Big)- \frac{1}{N_{\textrm{train}}} L(Z^{\textrm{train}}_j,m_\theta)\Big)
    \end{equation}
    is the set of parameters that minimize the loss on all training examples except $Z^{\textrm{train}}_j$.
\end{definition}

\begin{algorithm}[H]
    \caption{Influential Example Dataset Difference Explanations Based on Feature Importance}
    \begin{algorithmic}[1]
    % Change 
        \Require $\mathcal{D} = \{(X_i, Y_i)\}_{i=1}^N$, $\mathcal{D}^\prime = \{(X_i^\prime, Y^\prime_i)\}_{i=1}^{N^\prime}$
        \State Let $\mD_\phi = \{(\phi(X,Y,\mD), 1) \ \textrm{if} \ (X,Y) \in \mathcal{D} \textrm{ else} \ (\phi(X,Y,\mD^\prime), 0), \  \forall (X,Y) \in \mathcal{D} \cup \mD^\prime\}$ be the dataset of LiFIMs and corresponding labels computed from both datasets $\mD$ and $\mD^\prime$ (using \cite{donnelly2023rashomon})
        \State Train a logistic regression model $m_\theta(X)$ to classify $\mathcal{D}$ vs $\mathcal{D}^\prime$ using the dataset $\mD_\phi$
        \State Scores $= \emptyset$
        \For {each example $Z^\prime \in \mD_\phi$}
        \State $s_{Z^\prime} = I(Z^\prime, \mD_\phi, m_\theta)$ \Comment{This is computed using Equation~\ref{eqn:influence}}
        \State Add $s_{Z^\prime}$ to Scores \\
        \EndFor
        \Return The $K$ examples in $\mathcal{D^\prime}$ with the highest $s_{Z^\prime}$ in Scores
    \end{algorithmic}
    \label{alg:fi_alignment}
\end{algorithm}
Algorithm~\ref{alg:fi_alignment} finds the examples that are most detrimental to the performance of the discriminator (i.e., have the highest positive influence value $I(Z^\prime, \mathcal{D}^\prime, m_\theta)$). Because the discriminator learns to distinguish between $\mD$ and $\mD^\prime$ based on their respective feature importance measures, removing the examples found by our algorithm will make the remaining feature importances look more indistinguishable. That is, once we find the set $S \in \mD$ of examples to remove, $d\big(\phi_g(\mD), \phi_g(\mD^\prime\backslash S)\big)$ will become smaller -- we also demonstrate this through empirical studies later. Knowledge of these influential examples can be valuable to the end user, not only to precisely understand the properties of `culprit' examples that make $\mD$ and $\mD^\prime$ different, but also to design ways to remediate this difference by generating or removing certain examples.

\subsubsection{Case Study 1: Low Dimensional Tabular Data - Adult Dataset}
\label{sec:fi_exp_adult}
The Adult dataset contains demographic information from the $1994$ US Census database. In particular, each data point corresponds to information on age, sex, education levels, marital status, race, and occupation of an individual. The underlying task is to predict if the annual income of the individual is $\geq \$50$K. In this section, we aim to explain the difference between Adult male and females. This analysis sheds light on potential biases in the dataset, which may affect model predictions and subsequently influence decision-making processes.

\paragraph{Dataset $\mD$} $\mD$ corresponds to the dataset of all males, but with the same subset of features as \citet{towards_explain} -- age, education, and income. The income feature is encoded as $1$ if the annual income is $\geq \$50k$ and $0$ otherwise.
\paragraph{Dataset $\mD^\prime$} This is the dataset of all females, preprocessed in the same manner as $\mD$.
Thus, we will be examining differences between the two ``sex'' datasets. We followed the procedure as outlined in Section~\ref{sec:meth:featimp} for the Adult male ($\mD$) and female datasets ($\mD^\prime$). We identified $N = 50$ influential examples in $\mD^\prime$ and examined their characteristics -- these correspond to only $\approx 1\%$ of the dataset. In order to use GOSDT \citep{gosdt} decision trees, we first binarised the age and education-num features by thresholding, using the method of \citet{gosdt_guesses}.

\begin{figure}[H]
    \centering
    \includegraphics[width=0.8\textwidth]{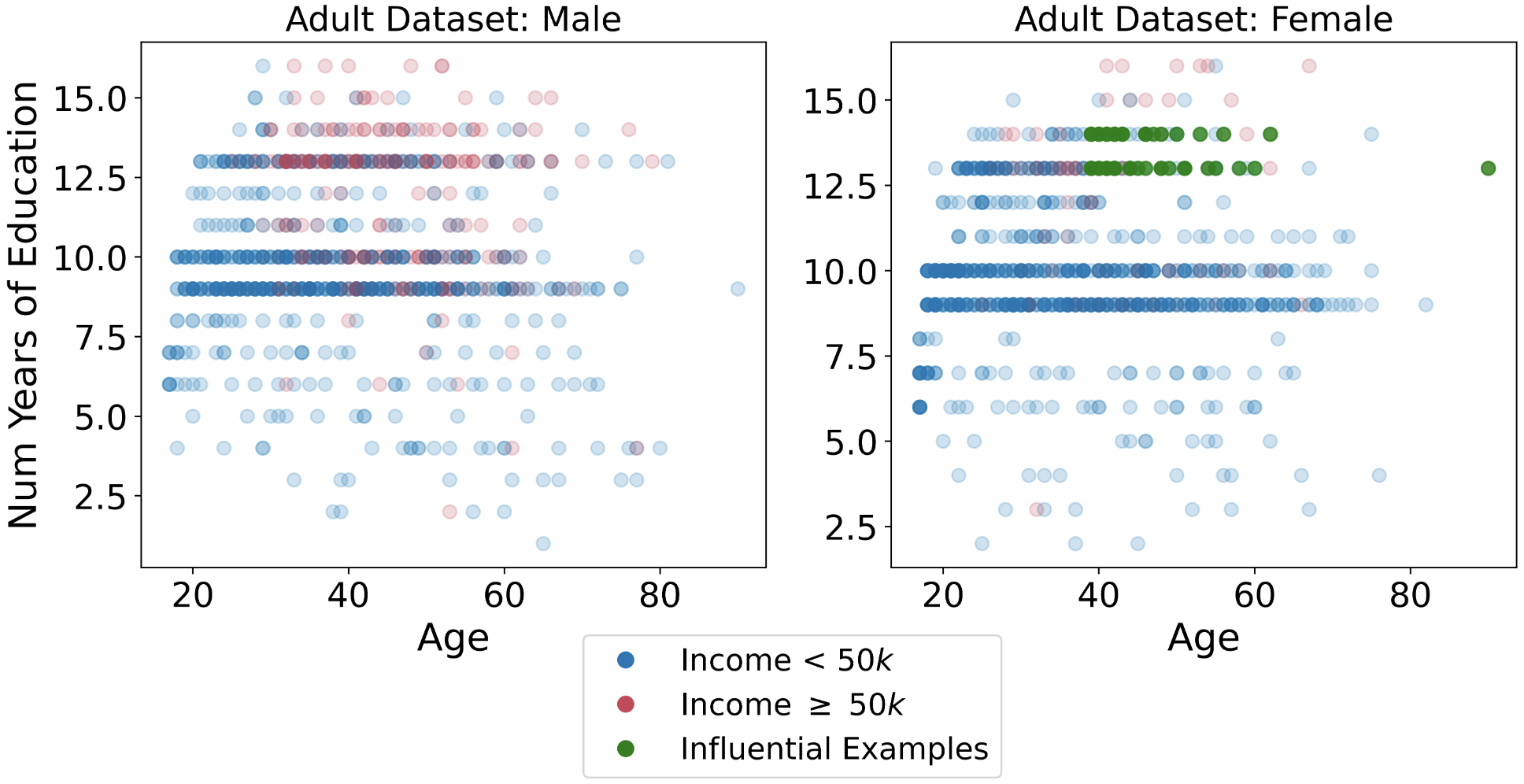}
    \caption{Visualization of the Adult male ($\mD$) and female datasets ($\mD^\prime$) with the influential examples for $\mD^\prime$ overlaid. Our explanation aims to show that these examples are a big reason why the feature importances in the female dataset are different from the male dataset - hence, we only highlight influential examples in the female dataset and 'fix' the male dataset. These influential examples are seen to be localised to a specific part of the feature space. In particular, they are examples of young to middle-aged women with many years of education. We place this into context in the analysis below.}
    \label{fig:adult_influential_examples}
\end{figure}

For visualization purposes, Figure~\ref{fig:adult_influential_examples} shows non-binarised features and the respective influential examples. 
\begin{figure}[H]
    \centering
    \includegraphics[width=0.8 \textwidth]{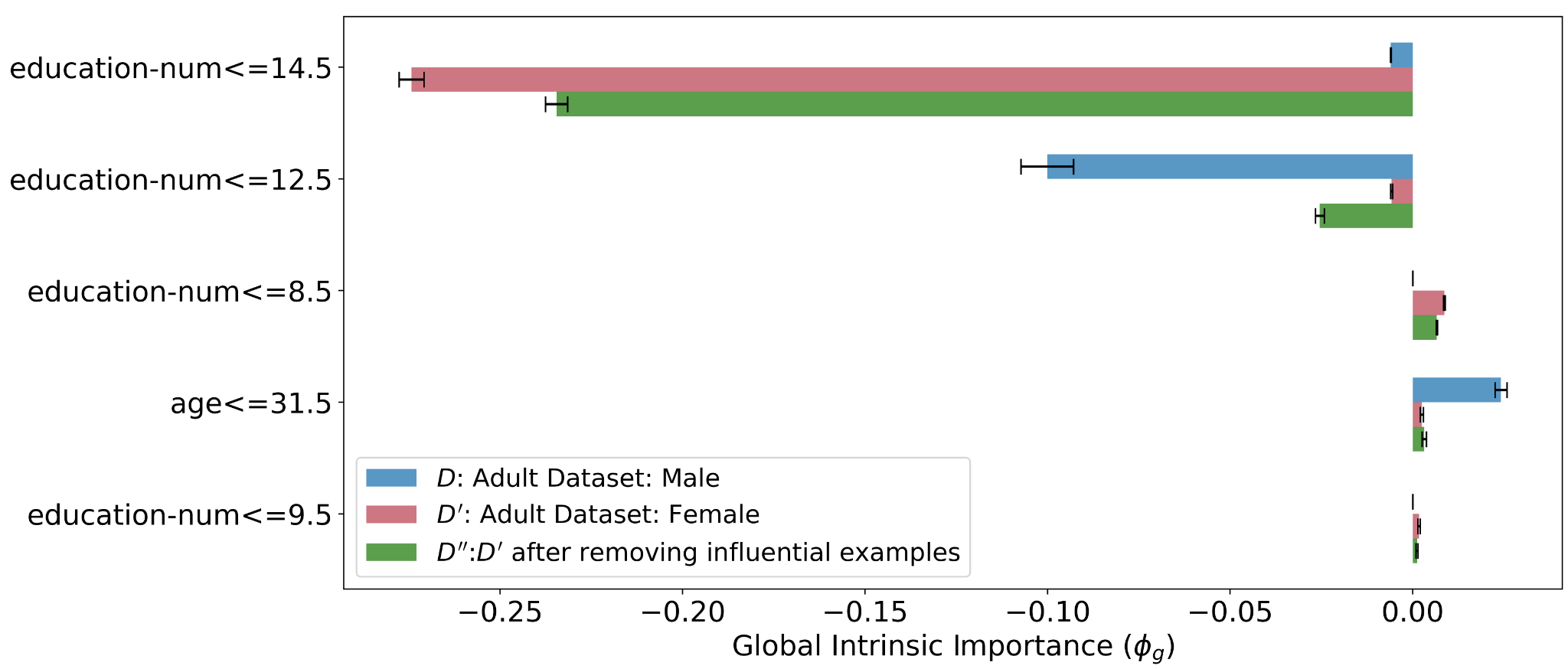}
    \caption{Global intrinsic feature importances for Adult males ($\mD$), Adult females ($\mD^\prime$), and Adult females after removing $20$ influential examples (i.e., $\mD^{\prime\prime}$). The task is to predict whether the annual income is $\geq 50$k from binarised age and education features. We show features whose importances in $\mD^{\prime\prime}$ are most aligned. In particular, note how the blue and the green bars in the plot (corresponding to $\mD$ and $\mD^{\prime\prime}$) are closer than the blue and red bars (resp.$\mD$ and $\mD^\prime$). Compared to men, women who have less than $14.5$ years of education are disproportionately more likely to have lower income -- this is the most affected feature. However, removing a small number of influential examples from the female dataset closes this gap -- we discuss the implications of this below. }
    \label{fig:adult_fi_influential}
\end{figure}
\begin{table}[H]
\centering
\resizebox{\textwidth}{!}{
\begin{tabular}{c|cc|clcc}
\hline
\textbf{Dataset}                     & \textbf{Age} & \textbf{Num Education Years} & &\textbf{$\#$ Income $\leq \$50k$} & \textbf{$\#$ Income $\geq \$50k$} \\ \hline
Adult male: $\mD$  & 39.86 $\pm$ 0.42   & 10.10 $\pm$ 0.08 &   & 709                           & 291 \\
Adult female: $\mD^\prime$   & 36.13 $\pm$ 0.44       & 9.91 $\pm$ 0.07 &                         & 899                             & 101 \\
Influential Examples in $\mD^\prime$ & \textbf{46.12 $\pm$ 1.21} & \textbf{13.38 $\pm$ 0.07}  &                           & \textbf{38}           & \textbf{12} \\ \hline
\end{tabular}
}
\caption{Mean value of the features ($\pm$ standard error) \texttt{Age} and \texttt{Num Education Years} alongside the class balance of $\mD$, $\mD^\prime$, and the influential examples. We see that the average influential examples all have similar characteristics -- they include older women who are highly educated but are mostly not commanding a high income. Removing instances of these examples better aligns the intrinsic importances of features in the male and female datasets.}
\label{tab:influential_examples_adult}
\end{table}

Given the above information in Table \ref{tab:influential_examples_adult} and Figure \ref{fig:adult_fi_influential}, we can posit one dataset explanation: \textit{Women who have less than $14.5$ years of education are more likely to have lower income than men. This association is driven in large part due to a few highly educated $(\sim 13.4$ years), middle-aged women, most of whom are not earning well}. Thus, analysing the properties of influential examples in datasets can not only uncover insights as to why $\mD$ and $\mD^\prime$ differ in their intrinsic feature importances for the given task, but also highlight specific biases that may exist within the data.
%%%%%
\subsubsection{Case Study 2: High Dimensional Tabular Data - HELOC Dataset}
\label{sec:fi_exp_heloc}
This dataset, which was used in the Explainable Machine Learning Challenge, contains information from the credit reports of around $12000$ people. In particular, it contains features relating to trade characteristics (e.g., total trades, overdue trades, etc), consolidated risk indicators (external risk estimate, longest delinquency period, etc), and miscellaneous indicators (e.g., length of credit history). The task is to predict whether an applicant for a loan will repay it back within $2$ years. Following \citet{towards_explain}, we generate two separate datasets corresponding to low risk and high risk individuals. This is done by splitting the HELOC dataset on the variable \verb|ExternalRiskEstimate|.  

\paragraph{Dataset $\mD$} This is the low risk dataset. Concretely, $\mD = \{(X,Y)| \verb|ExternalRiskEstimate|(X) \leq 70\}$. \verb|ExternalRiskEstimate| is a black-box metric computed by external agencies that estimates the risk of defaulting. We chose to split the data on this feature because it is likely that there is a distribution shift between individuals with high and low \verb|ExternalRiskEstimate|. 
\paragraph{Dataset $\mD^\prime$} This is the high risk dataset. $\mathcal{D}^\prime = \{(X,Y)| \verb|ExternalRiskEstimate|(X) > 70\}$.
\\

\begin{figure}[H]
    \begin{subfigure}[b]{0.52\textwidth}
        \centering
        \includegraphics[width=\textwidth]{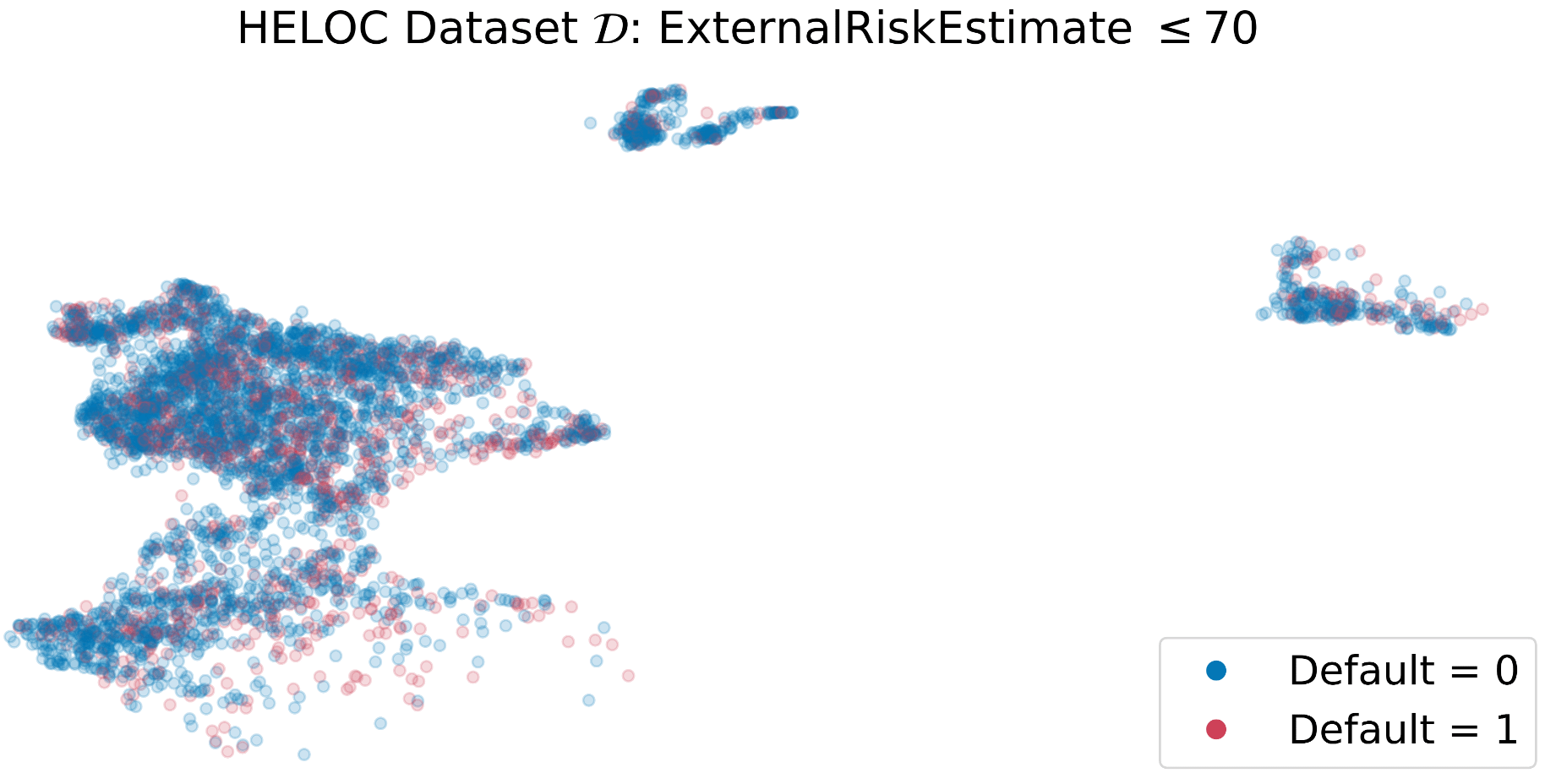}
        \caption{HELOC Dataset $\mD$: \texttt{ExternalRiskEstimate} $\leq 70$\\}
    \end{subfigure}
    \begin{subfigure}[b]{0.5\textwidth}
        \centering
        \includegraphics[width=\textwidth]{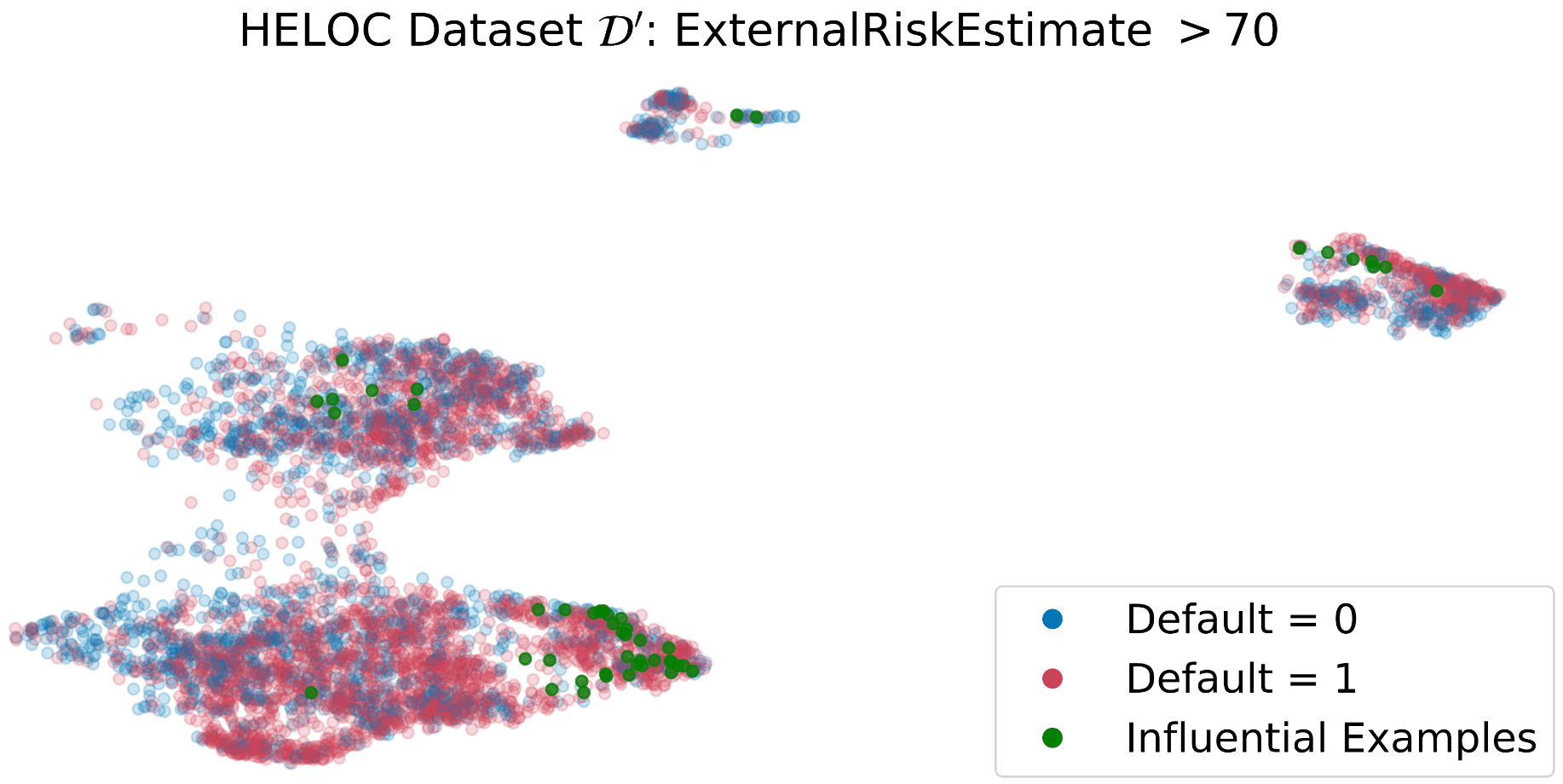}
        \caption{HELOC Dataset $\mD^\prime$: \texttt{ExternalRiskEstimate} $> 70$\\}
    \end{subfigure}
    \caption{PaCMAP \citep{pacmap} projection of HELOC datasets $\mD$ and $\mD^\prime$ in a common $2$-D space, but with the influential examples for $\mD^\prime$ overlaid. The most influential examples are seen to be localised to a specific part of the feature space. From Section~\ref{sec:featureimportancemethod}, these are the examples that, if removed from $\mD^\prime$, would most likely align the feature importances of $\mD$ and $\mD^\prime$. We examine this further below.}
    \label{fig:heloc_fi_explanation}
\end{figure}

\begin{figure}[H]
    \centering
    \includegraphics[width=\textwidth]{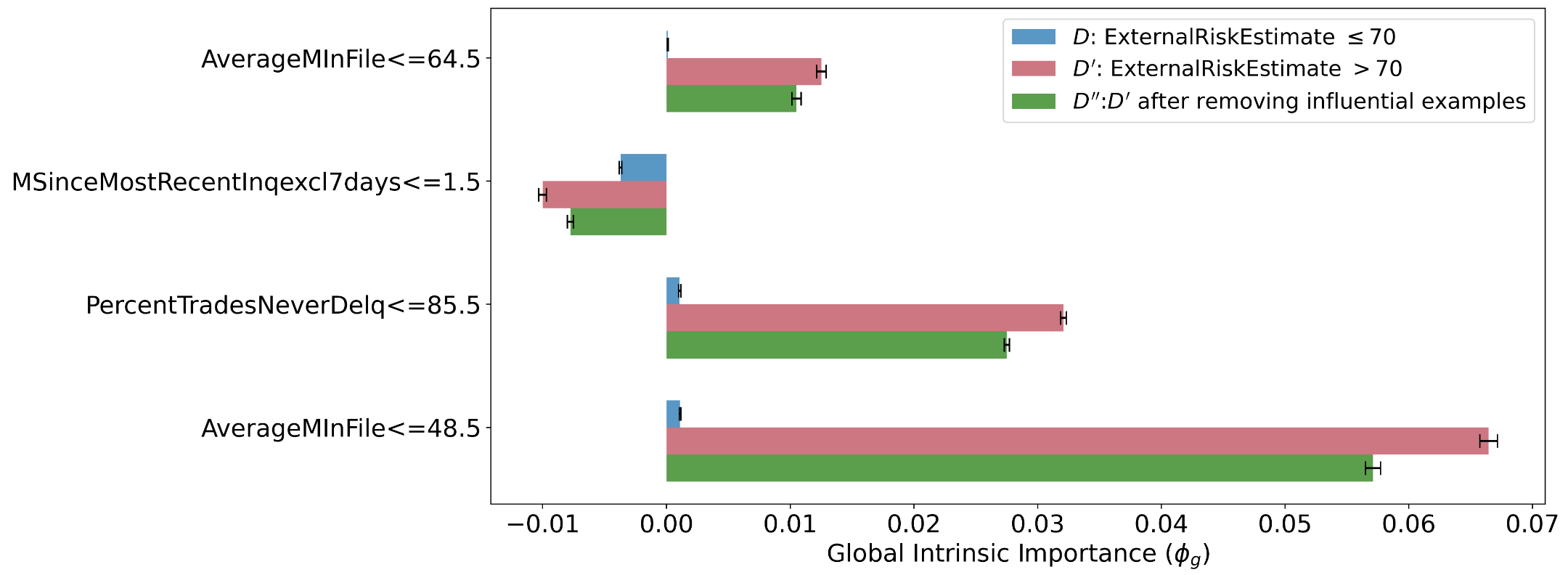}
    \caption{Global intrinsic feature importances for datasets $\mD$, $\mD^\prime$, and $\mD^{\prime}$ after removing influential examples (i.e., $\mD^{\prime\prime}$). \texttt{AverageMinFile} = Length of credit history. \texttt{MSinceMostRecentInq} = Months since most recent credit inquiry. \texttt{$\%$ TradesNeverDelq} = $\%$ of non-delinquent trades. We show binarised features (e.g. \texttt{AverageMinFile} $\leq 64.5$) that have the greatest change in feature importance between $\mD^{\prime\prime}$ and $\mD^\prime$ . In particular, note how the blue and the green bars in the plot (corresponding to $\mD$ and $\mD^{\prime\prime}$) are closer than the blue and red bars (resp. $\mD$ and $\mD^\prime$). We examine the properties of examples removed from $\mD^\prime$ to see why this is the case (see Table~\ref{tab:influential_examples_heloc}.)}
    \label{fig:fi_heloc}
\end{figure}

We now attend to influential example-based explanations for HELOC. We use the Rashomon Importance Distribution (RID) \cite{donnelly2023rashomon} as the feature importance measure (see Section~\ref{sec:featureimportancemethod} for details). As with Section~\ref{sec:tab_adult_prototype}, we first binarized the features in $\mD$ and $\mD^\prime$ using threshold guessing \cite{gosdt_guesses} as this is required as input to GOSDT and the RID framework. Let $\phi_g(\mD)$ and $\phi_g(\mD^\prime)$ be the global intrinsic feature importance measures (GiFIM) for the datasets $\mD$ and $\mD^\prime$ respectively (see Definition~\ref{def:gifim}).
\begin{itemize}
    \item We first identified the $N = 50$ \textit{influential} examples in $\mD^\prime$ using Algorithm~\ref{alg:fi_alignment}. Because $\mD^\prime$ is of size $\approx 4500$, these influential examples correspond to only $\approx 1\%$ of the dataset. Figure \ref{fig:heloc_fi_explanation} shows these examples highlighted in the original dataset.
    \item We then removed these examples from the dataset $\mD^\prime$. Call this new dataset $\mD^{\prime\prime}$.
    \item Lastly, we recomputed the LiFIMs and GiFIMs on $\mD^{\prime\prime}$.
\end{itemize}
We now show the resulting feature importances in Figure~\ref{fig:fi_heloc}. We then look at the features whose importances were most affected by this removal.

\begin{table}[H]
\centering
\resizebox{\textwidth}{!}{%
\begin{tabular}{cclccccc}
\hline
\textbf{Dataset}                     & \textbf{AverageMinFile} & \textbf{MSinceMostRecentInq} & \textbf{$\%$TradesNeverDelq} &  & \textbf{$\#$ Default = 0} & \textbf{$\#$ Default = 1} \\ \hline
$\mD$                                & 67.00 $\pm$ 0.47       & 0.1 $\pm$ 0.07 & 91.01 $\pm$ 0.21 &  & 1746                          & 3566                           \\
$\mD^\prime$                         & 86.76 $\pm$ 0.46       & 0.70 $\pm$  0.09 & 97.10  $\pm$ 0.08 &  & 3390                          & 1169                           \\
Influential Examples in $\mD^\prime$ & \textbf{106.92 $\pm$ 4.99}       & NaN & \textbf{99.16 $\pm$ 0.28} &  & \textbf{6}                            & \textbf{44}                              \\ \hline
\end{tabular}
}

\cprotect\caption{Average value $\pm$ standard error of some original (non-binarised) important features and number of examples of each class (Default = $0$ and Default $=$ 1) in $\mD$, $\mD^\prime$, and the influential examples. We see that the influential examples correspond to individuals with high \texttt{AverageMinFile} and \texttt{$\%$TradesNeverDelq} and no known recent inquiry (\texttt{MSinceMostRecentInq} is NaN -- these are given a special value of -$8$ in the dataset). This corresponds to individuals with longer credit histories who have almost no delinquent trades and no credit inquiries on their profile. Despite these positive indications, most of these individuals have defaulted on their loans in the last $2$ years ($44$ out of $50$ samples with Default $=1$).}
\label{tab:influential_examples_heloc}
\end{table}
We can now compare the two datasets by considering the properties of influential examples in Table~\ref{tab:influential_examples_heloc} and the GiFiMs of important features in Figure~\ref{fig:fi_heloc}. The dataset difference explanation therefore tells us the following: \textit{The binary features \texttt{TradesNeverDelq} $\leq 85.5$, $\texttt{AverageMinFile} \leq 48.5$, $\texttt{AverageMinFile} \leq 64.5$, and $\texttt{MSinceMostRecentInq} \leq 1.5$ are considered to be unusually important in the higher risk dataset $\mD^\prime$ compared to $\mD$. However, this is in large part due to a few individuals in $\mD^\prime$ who mostly defaulted on their loan in the last $2$ years despite having $\approx 99\%$ non-delinquent trades, longer credit history, and no recent credit inquiries.}

\subsection{Prototype-Neighbourhood-Based Explanations}
\label{sec:protoneigh_explainmethod}
\subsubsection{Introduction}
\label{sec:intro_proto_quant}
\begin{figure}[H]
    \begin{subfigure}[b]{0.5\textwidth}
        \centering
        \includegraphics[width=\textwidth]{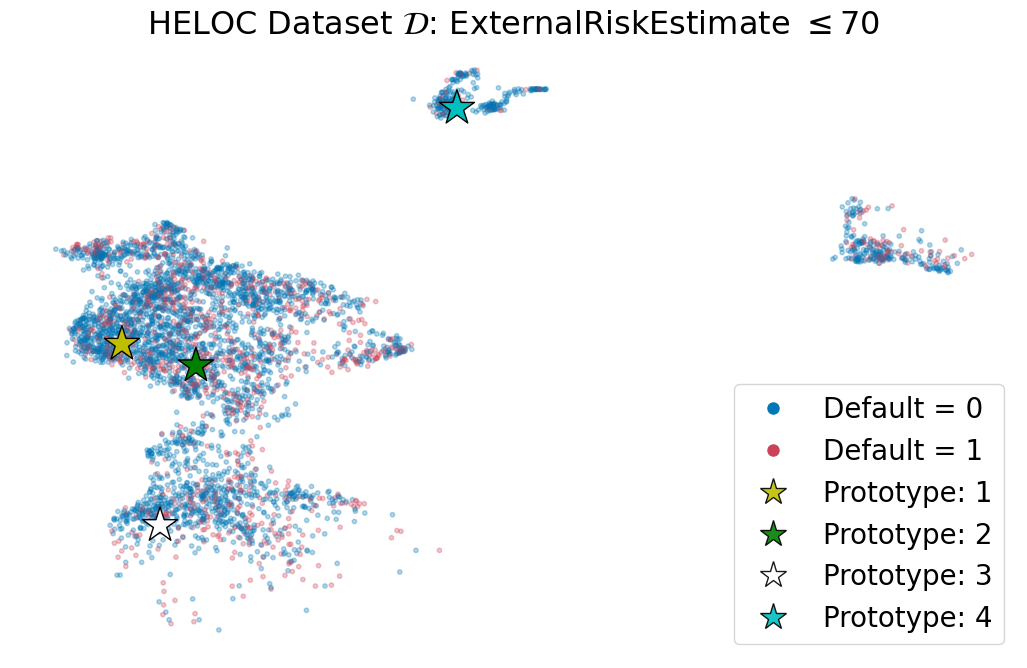}
        \caption{HELOC Dataset $\mD$: \texttt{ExternalRiskEstimate} $\leq 70$\\}
    \end{subfigure}
    \begin{subfigure}[b]{0.5\textwidth}
        \centering
        \includegraphics[width=\textwidth]{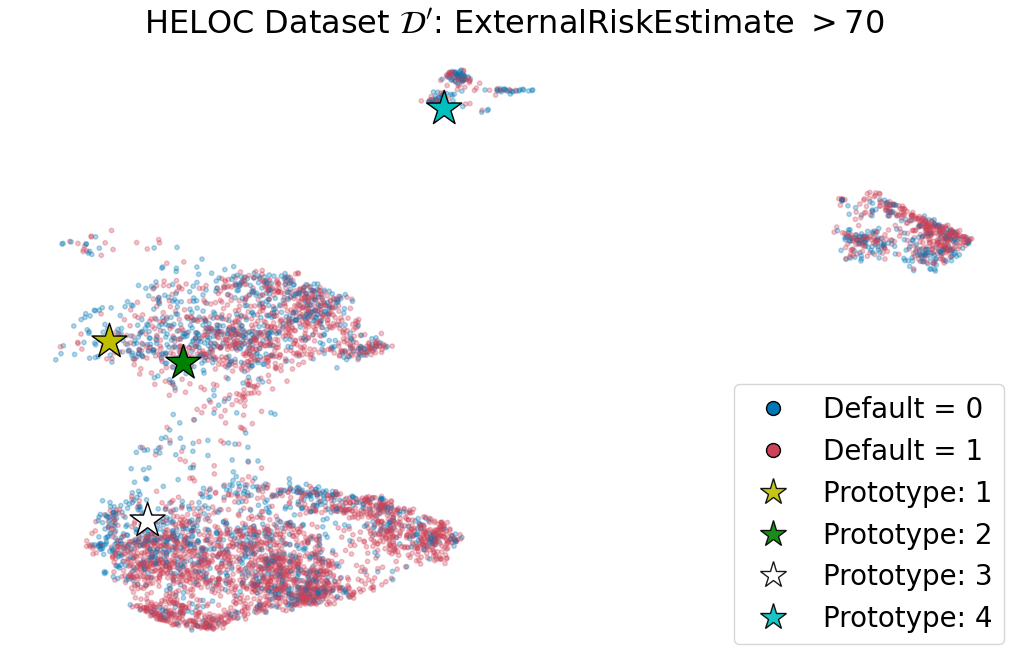}
        \caption{HELOC Dataset $\mD^\prime$: \texttt{ExternalRiskEstimate} $> 70$\\}
    \end{subfigure}
    \caption{PaCMAP projection of HELOC datasets $\mD$ and $\mD^\prime$ in a common $2$-D space. The color of each point corresponds to its class label. The same prototypes that were learned on $\mD$ (left) are being visualized on $\mD^\prime$ (right). Both datasets are normalized using the mean and standard deviation of features from $\mD$.}
    \label{fig:heloc_figs}
\end{figure}

Given two datasets $\mD$ and $\mD^\prime$, prototype-based explanations compare these datasets using a set of prototypical samples $P = \{p_1, p_2, p_3, ..., p_n\}$. Each of these prototypes is considered to be a meaningful and faithful representation of its neighboring samples when $\mD$ and $\mD^\prime$ are projected to a latent space. By comparing the neighborhood sample distribution between $\mD$ and $\mD^\prime$, we could provide insights into the differences between two datasets. There are multiple ways to create prototypes:

\begin{itemize}
    \item First, we can choose prototypes manually with domain knowledge. We show an example of this for explaining the difference between males and females in the Adult dataset in Section~\ref{sec:tab_adult_prototype}, i.e., in Figures~\ref{fig:adult_pacmap} and \ref{fig:adult_neighbouring_sample_distances}. 
    \item Second, cluster centers from clustering methods such as $k$-means as the cluster centers  \citep[similar to][]{towards_explain} can be seen as prototypes. This is illustrated in Section~\ref{sec:tab_heloc_prototype} to explain the difference between low and high risk examples in the HELOC dataset, i.e., in Figures~\ref{fig:heloc_figs} and \ref{fig:heloc_prototype}. 
    \item Third, prototypes and their surrounding latent space can be learned in a neural network in a supervised and end-to-end fashion, where the encoder $f$, prototype set $P$, and the final classifier layer are the learnable-components. We use this approach for explaining the differences between real and synthetic PPG data and human and machine generated audio. For this last approach, we adapt ProtoPNet \citep{thislookslikethat} and its variant \citep{interpEEG} to project both $\mD$ and $\mD^\prime$ into the same latent space of the learned encoder, as illustrated in Figure~\ref{fig:protosetp2}. ProtoPNet tends to have similar accuracy to its non-interpretable counterparts despite being trained to use case-based reasoning, thus providing assurance of the quality of the learned latent space from a performance perspective. In this latent space, we make quantitative comparisons between the learned prototypes $P$ and their neighborhoods in $\mD$ and $\mD^\prime$. 
\end{itemize} 

\subsubsection{Quantitative Comparison Between neighborhoods}
\label{sec:method_nspdnsdd}
Once the prototypes corresponding to $\mD$ are generated, we can use two metrics to analyze the differences between $\mD$ and $\mD^\prime$: 
\begin{definition}[\textbf{Neighboring Sample Proportion Difference -- NSPD}]
\label{def:nsdd}
    The \textbf{neighboring samples} for prototype $p_i$ are defined as the samples that have $p_i$ as their closest prototype. The neighboring sample distribution difference for $p_i$ is calculated as the difference between the percentage of $p_i$'s neighboring samples in $\mD$ and the percentage of $p_i$'s neighboring samples in $\mD^\prime$.
\end{definition}

\begin{figure}[H]
    \begin{subfigure}[b]{0.48\textwidth}
        \centering
        \includegraphics[width=0.9\textwidth]{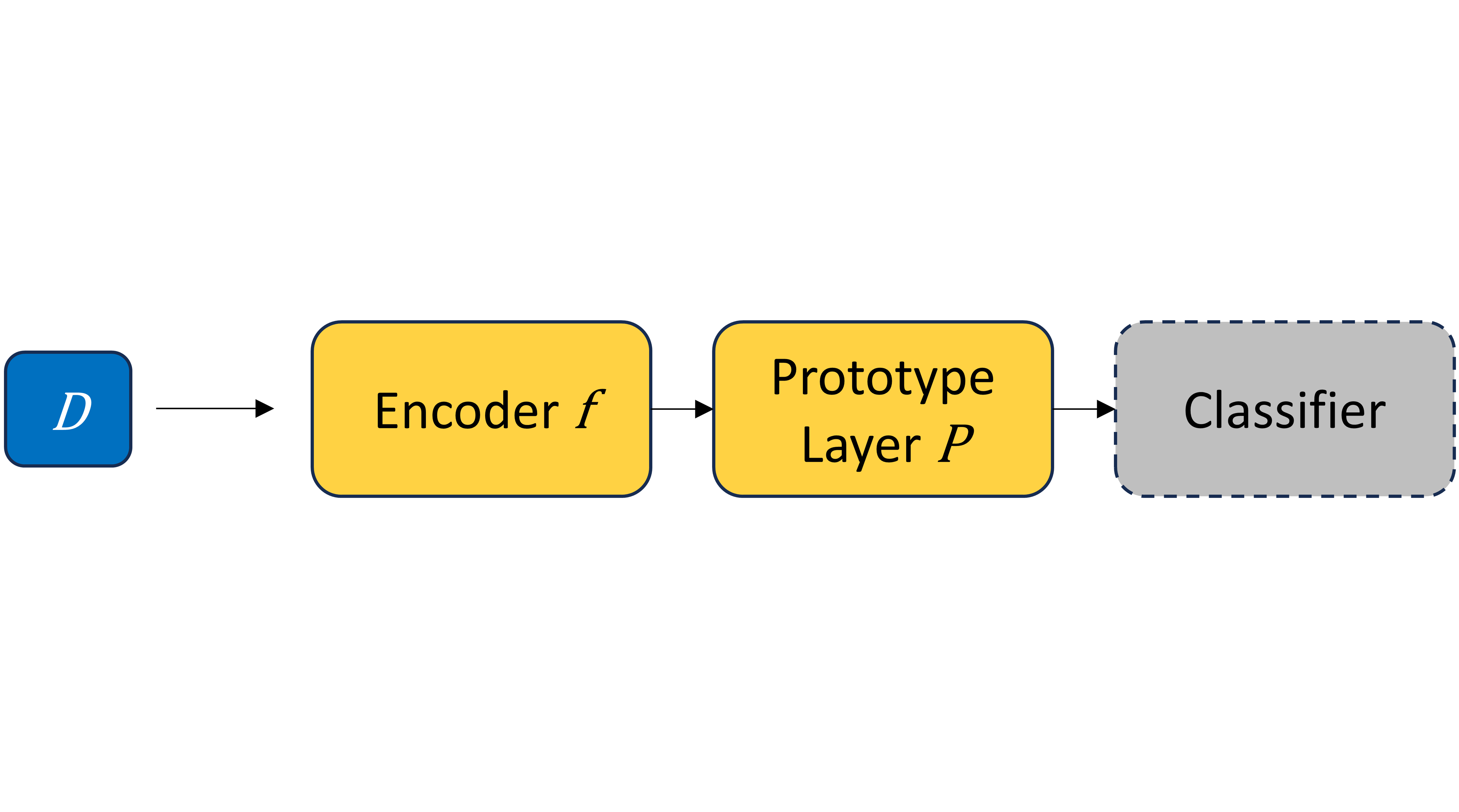}
        \caption{Prototypical explanation Step 1: train encoder.}
        \label{fig:protosetp1}
    \end{subfigure}
    \begin{subfigure}[b]{0.48\textwidth}
        \centering
        \includegraphics[width=0.9\textwidth]{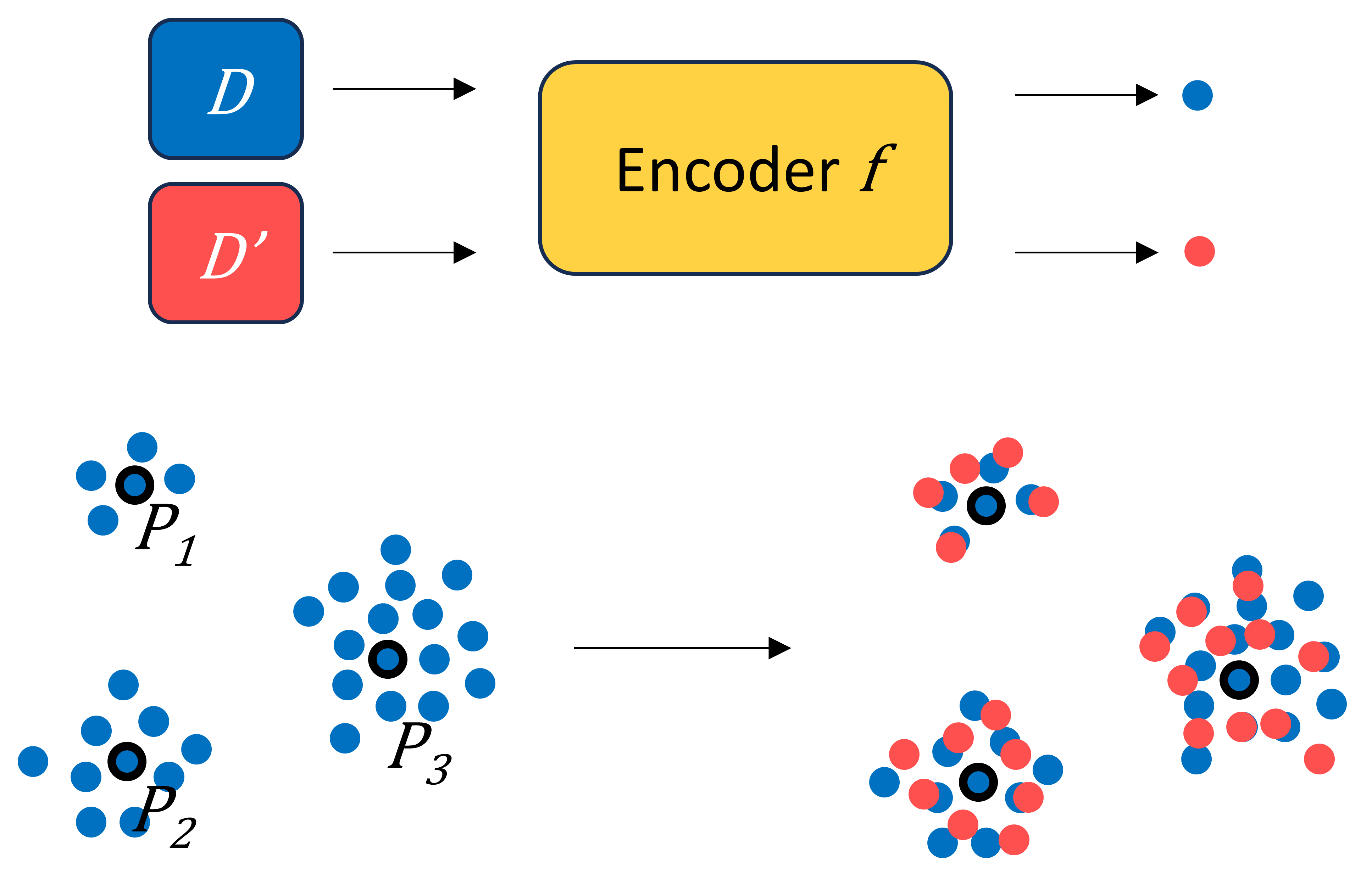}
        \caption{Prototypical explanation Step 2: Send both datasets through the encoder.}
        \label{fig:protosetp2}
    \end{subfigure}
    \newline
    \begin{subfigure}[b]{0.255\textwidth}
        \centering
        \includegraphics[width=\textwidth]{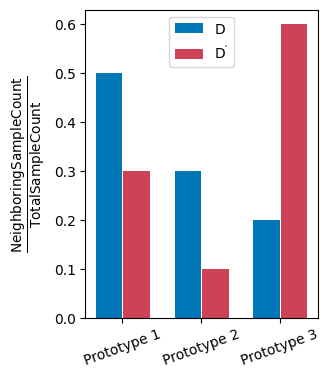}
        \caption{$\mD$ and $\mD^\prime$ have different NSPD}
        \label{fig:protoex_badp1}
    \end{subfigure}
    \begin{subfigure}[b]{0.255\textwidth}
        \centering
        \includegraphics[width=\textwidth]{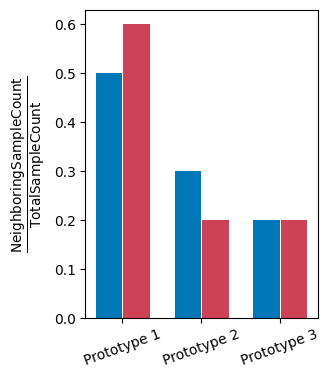}
        \caption{$\mD$ and $\mD^\prime$ have aligned NSPD}
        \label{fig:protoex_goodp1}
    \end{subfigure}
    % \newline
    \begin{subfigure}[b]{0.235\textwidth}
        \centering
        \includegraphics[width=\textwidth]{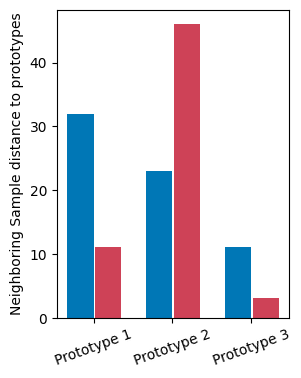}
        \caption{$\mD$ and $\mD^\prime$ have different NSDD}
        \label{fig:protoex_badp2}
    \end{subfigure}
    \begin{subfigure}[b]{0.235\textwidth}
        \centering
        \includegraphics[width=\textwidth]{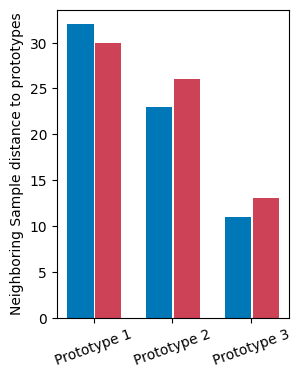}
        \caption{$\mD$ and $\mD^\prime$ have aligned NSDD}
        \label{fig:protoex_goodp2}
    \end{subfigure}
    \newline
    \begin{subfigure}[b]{0.33\textwidth}
        \centering
        \includegraphics[width=0.45\textwidth]{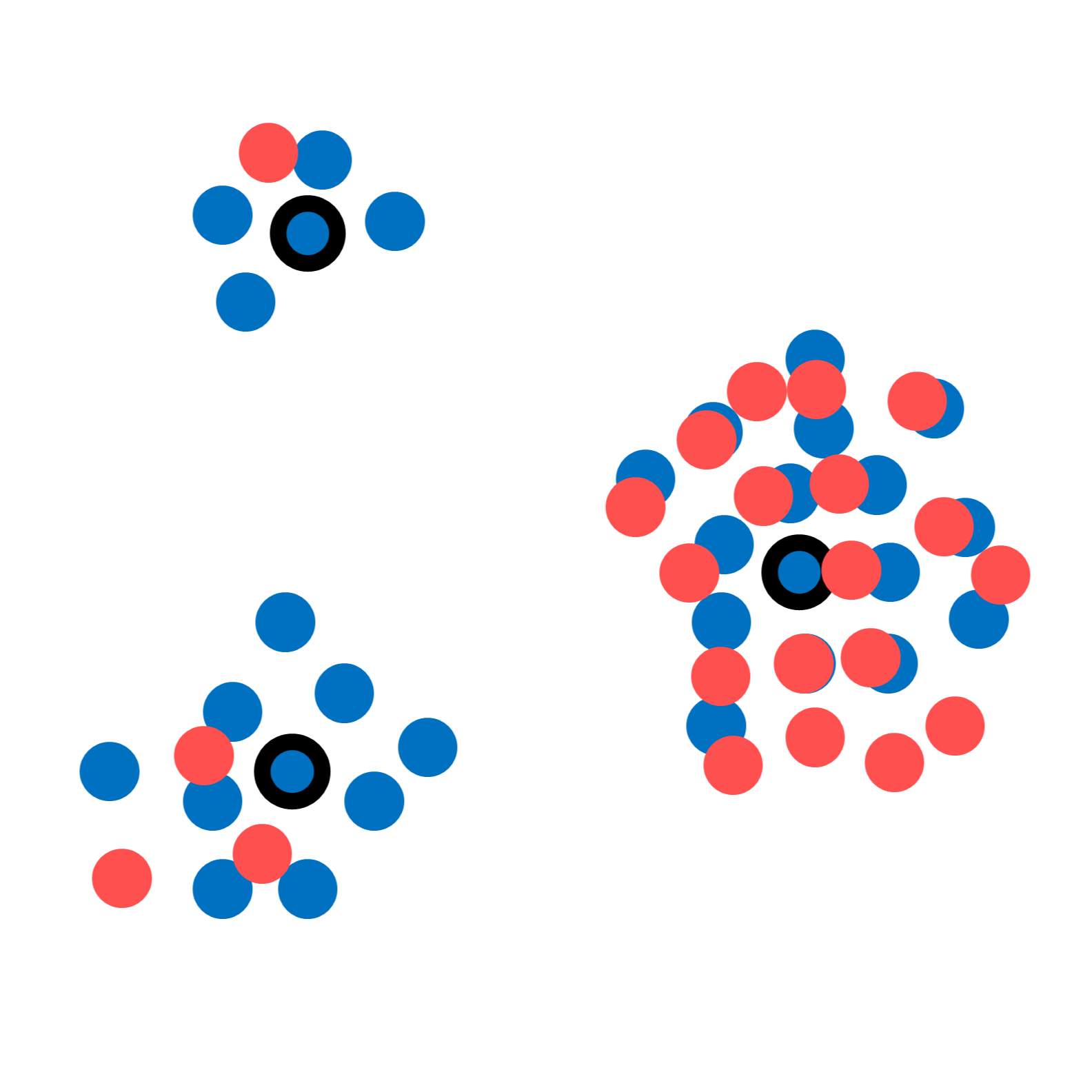}
        \caption{High NSPD \& Low NSDD}
        \label{fig:proto_two_def_example_p1}
    \end{subfigure}
    \begin{subfigure}[b]{0.33\textwidth}
        \centering
        \includegraphics[width=0.45\textwidth]{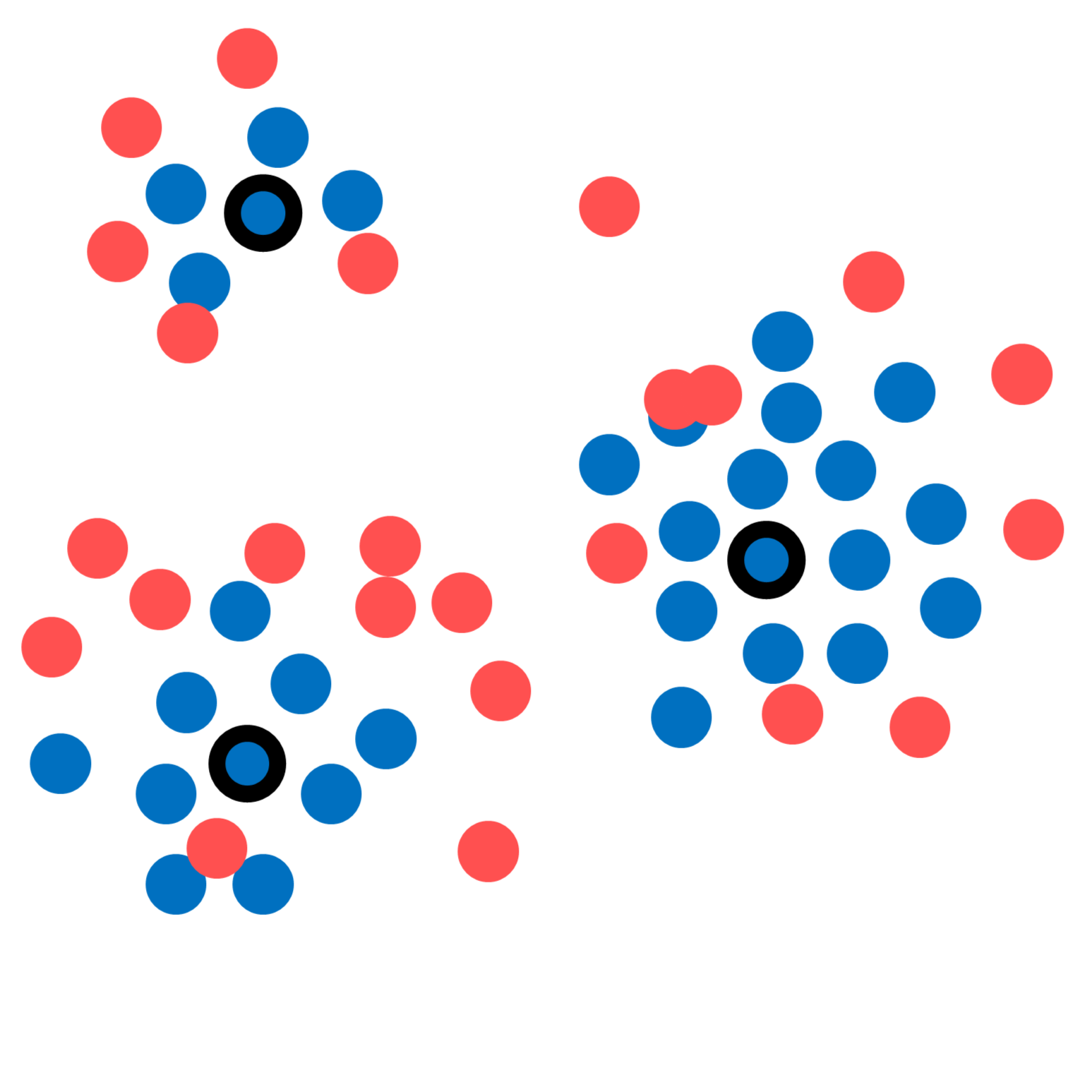}
        \caption{Low NSPD \& High NSDD}
        \label{fig:proto_two_def_example_p2}
    \end{subfigure}
    \begin{subfigure}[b]{0.33\textwidth}
        \centering
        \includegraphics[width=0.45\textwidth]{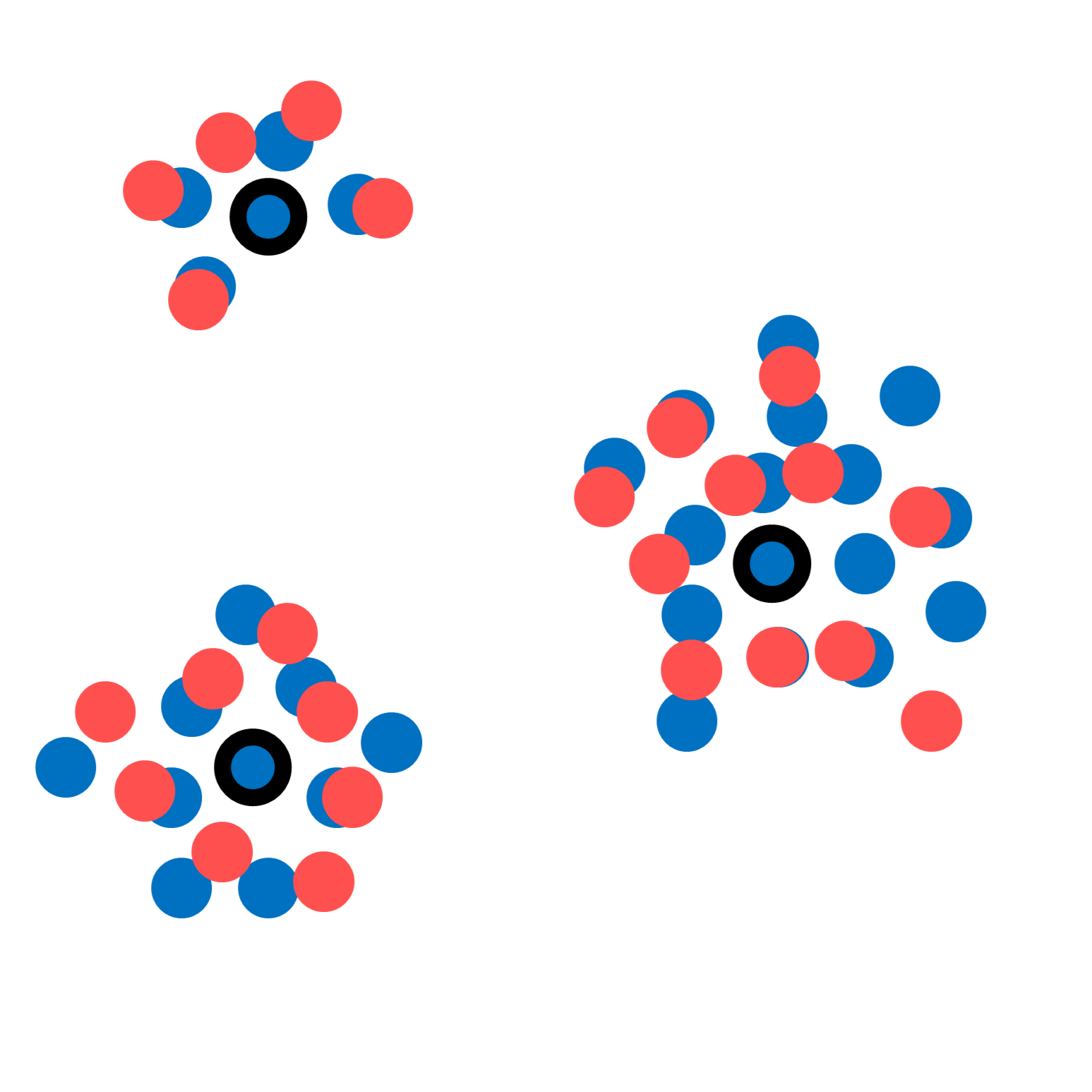}
        \caption{Low NSPD \& Low NSDD}
        \label{fig:proto_two_def_example_p3}
    \end{subfigure}
    \caption{In Step 1, we follow the approach of \citet{interpEEG} to learn both an encoder and a set of faithful prototypes. In Step 2, both $\mD$ and $\mD^\prime$ are encoded by the learned encoder in Step 1. The encoded features of samples in $\mD^\prime$ are projected into the sample latent space of $f$, the encoder learned on $\mD$. The samples in $\mD^\prime$ are compared against the prototypes $P$ in the same latent space, users can see which parts of $\mD^\prime$ are not evenly distributed in the latent space. In (c), (d), (e), (f), we show four examples of P-1 and P-2 evaluation results. The NSPD and NSDD are based on Definitions~\ref{def:nsdd} and \ref{def:nsdd2} respectively.} 
\end{figure}

\begin{definition}[\textbf{Neighboring Sample Distance Difference -- NSDD}]
\label{def:nsdd2}
    The neighboring sample distance difference for $p_i$ is calculated as the difference between the average neighboring sample distance to $p_i$ in $\mD$ and the average neighboring sample distance to $p_i$ in $\mD^\prime$. The distance between a sample's feature and a prototype in the latent space is calculated using cosine distance.
\end{definition}
We can compute these differences either in the original feature space or project the prototypes to a latent space using a learned encoder. Figures~\ref{fig:protosetp1} and \ref{fig:protosetp2} illustrate the process of obtaining prototypes in latent space. In Section~\ref{sec:prototypical_explanations_appendix_analysis} in the appendix, we examine how adjusting the number of prototypes influences the balance between the explanation's complexity and its faithfulness. We also discuss in Section~\ref{sec:practical_guide} practical justifications for the number of prototypes in the explanations, as it is an important design choice. 

We show an example of two toy examples with high NSPD and low NSDD in Figure~\ref{fig:proto_two_def_example_p1}, another pair of toy examples with high NSDD but low NSPD in Figure~\ref{fig:proto_two_def_example_p2}, and a pair of toy examples with both low NSDD and low NSPD in Figure~\ref{fig:proto_two_def_example_p3} to illustrate Def.~\ref{def:nsdd} and Def.~\ref{def:nsdd2} in practice. In addition to quantitative comparisons, users can also inspect each prototype and perform visual comparisons with samples in $\mD$ and $\mD^\prime$. We show examples of this throughout the paper. 

\subsubsection{Comparing Prototype Neighborhoods in $\mD$ and $\mD^\prime$ in high dimensions}
\label{sec:partialprotoexplain}
For tabular datasets with a large number of features, it is useful to use only a subset of relevant variables within the NSDD and NSPD calculation. As we will see, using a good subset will allow a high-quality approximation of the full NSDD and NSPD (see Section~\ref{sec:partial_prototypes_analysis} in the appendix), with a much sparser feature set. Our partial prototypical explanation provides the NSDD and the NSPD for the prototype along with the $K$ most relevant features of the prototype for the user to focus on. The notion of a relevant feature is based on two desiderata: \textit{value stability }and \textit{rank stability}.

 \begin{definition}[\textbf{Value Stability}] The $K$ chosen features must vary less around the prototype neighborhood, i.e., given a prototype $X_p$ from dataset $\mathcal{D}$ and $K$ chosen feature indices $\{m_1, ...m_k\}$, we want to ensure \\ $\mathbb{E}_{X^\prime \in \mathcal{D}^\prime | d(X^\prime,X_p) \leq \delta}\Big[d\Big(X_p[m_1,...m_k], X^\prime[m_1,...m_k]\Big)\Big]$ is small, where $\mD$ is a distance metric that can compare vectors of the same dimension (e.g., $\ell_2$, $\ell_1$, or distances that use inner products). \end{definition}
 If the dataset is labeled, we can optionally one-hot encode the labels and append them to the example vector before computing distances. While we chose examples in $\mD^\prime$ that are in an overall $\delta$ neighborhood of the prototype, there may be some features whose values in the neighborhood vary less than others. Thus, if we choose only $K$ features due to interpretability constraints, we are best off choosing important features whose values are most stable in the neighborhood. One important clarification: for the NSDDs and NSPDs of partial prototypes to remain approximately similar to the original prototypes, we want to preserve the structure of the prototype neighborhood as much as possible. Selecting more features will preserve neighborhood structure better but will lead to a loss in interpretability -- this tradeoff is illustrated in the appendix (Section~\ref{sec:explanation_quality_degradation}).

\begin{definition}[\textbf{Rank Stability}]
 If the datasets are labeled, the $K$ features selected should capture as much of the true model behavior as possible, i.e., \textit{they should be important for the \textbf{prototype in $\mD$} and similarly important for its \textbf{neighbors in both $\mD$ and $\mD^\prime$}}. 
 \end{definition}
 This helps the end user reason about neighboring sample distribution and distance differences only in terms of features that are equally important for both datasets. To this end, we generate the LiFIM $\phi(X,Y,\mD)$ using the Rashomon Importance Distribution (RID) method that can return a vector containing intrinsic feature importance scores for each feature in $(X,Y) \in \mD$. The equivalent LiFIM for $\mD^\prime$ is $\phi(X,Y,\mD^\prime)$. We further break down rank stability into two components below:  To enable this, we propose a score for each feature, and we will use the top scoring features within the partial prototype explanation:

    \begin{itemize}
    \item \begin{definition}[\textbf{Rank Difference Penalty}] If feature $j$ is deemed to be very important for the prototype $(X_p, Y_p) \in \mD$ (according to the local intrinsic importance $\phi(X,Y,\mD)$), but this feature is not so important for prototype neighbors in either $\mD$ or $\mD^\prime$, it is assigned a high penalty score -- this feature is less likely to be one of the $K$ selected. This penalty therefore penalizes the relative rank differences in the importance of feature $j$ in predicting the label for a prototype in $\mD$ and its neighbors in $\mD^\prime$ and $\mD$.
     \end{definition}
     \item \begin{definition}[\textbf{Absolute Rank Penalty}] The above mechanism could result in features which are less important for both the prototype and its neighborhood being selected (as only the relative rank difference is penalized). However, the chosen features should be important for both the prototype and the neighborhood. The absolute rank penalty aims to ensure that a chosen feature that has low rank difference penalty is also an important feature.
          \end{definition}
    \end{itemize}
As these forces can be opposing, we propose a score function for each feature that is based on a user-defined tradeoff between rank stability and value stability. Given feature $j$, datasets $\mD$ and $\mD^\prime$, an example $(X^\prime,Y^\prime)\in \mD^\prime$, and LiFIM $\phi(X^\prime,Y^\prime,\mD^\prime)$, let $U^\phi_j(X^\prime,Y^\prime, \mD^\prime) = \textrm{rank}\Big(\phi(X^\prime,Y^\prime,\mD^\prime)[j]\Big)$ be the rank of the importance of the feature (i.e., if $j$ is the $3^{rd}$ most important feature, then $U^\phi_j(X^\prime,Y^\prime,\mD^\prime) =3$). Then, the scoring function for feature $j$ given example $(X^\prime,Y^\prime) \in \mD^\prime$ and prototype $(X_p,Y_p) \in \mD$ is:
\begin{align}
\label{eqn:scoring_fn}
\begin{split}
    s_j(\mD,\mD^\prime,X_p,Y_p, X^\prime, Y^\prime) =
    &\underbrace{c_1\underbrace{\Big(\Big|U^\phi_j(X_p,Y_p,\mD)  - U^\phi_j(X^\prime,Y^\prime,\mD^\prime)\Big|\Big)}_{\textrm{Rank Difference Penalty}} + c_2\underbrace{\Big(0.5 U^\phi_j(X_p,Y_p,\mD) + 0.5 U^\phi_j(X^\prime,Y^\prime,\mD^\prime)\Big)}_{\textrm{Absolute Rank Penalty}}}_{\textrm{Rank Stability}} \\
    &+ c_3\underbrace{\Big|X_p[j] - X^\prime[j]\Big|}_{\textrm{Value Stability}}
\end{split}
\end{align}
The same scoring function can be defined for an example $(X,Y) \in \mD$. Algorithm~\ref{alg:partial_prototypes} then sums up scores across both datasets for each feature and prototype.

where the user can choose parameters $c_1$, $c_2$, and $c_3$ to weigh the relative importance of each desideratum. This naturally induces a tradeoff between value stability and rank stability, which is illustrated in Figure~\ref{fig:heloc_fsr_tradeoff_study} in the appendix. \\ \newline 
\begin{figure}[H]
    \centering
    \includegraphics[width=0.8\textwidth]{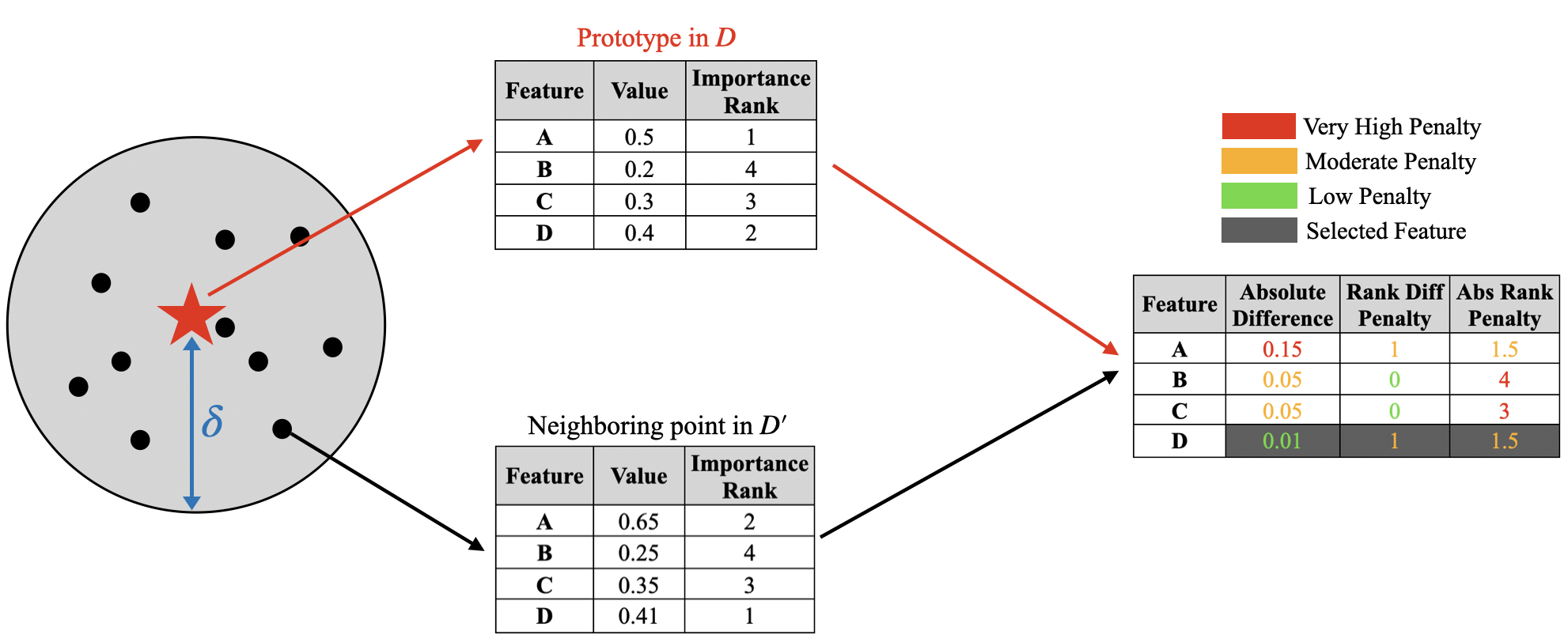}
    \caption{Simple example illustrating our interpretable partial prototype feature scoring procedure for $K = 1$ (i.e. choosing the best feature). Say we only consider two points to compute the feature scoring function -- the prototype in $\mD$ and a point in $\mD^\prime$ in the $\delta$ neighborhood of the prototype. We now compare the feature values and the feature importance values of each feature for both points: Feature $A$'s value differs a lot between the points compared to other features, and it is relatively important for predicting labels for both points. Features $B$ and $C$ are less important for both the prototype and the variable, and they do not differ as much between the two feature tables. Feature $\mD$ is very stable in value, is relatively important for prediction, and has only a moderate difference in rank between the prototype and the neighbor. Our scoring procedure therefore chooses feature $D$ as the partial prototype because it is reliably important for both $\mD$ and $\mD^\prime$. \protect }
    \label{fig:partial_protype_example_1}
\end{figure}
\begin{algorithm}[H]
    \caption{Partial Prototype-Based Explanations}
    \label{alg:partial_prototypes}
    \begin{algorithmic}[1]
    
    % Change 
        \Require $M$, $K$, $c_1$, $c_2$, $c_3$, $\delta$, $\mathcal{D} = \{(X_i, Y_i)\}_{i=1}^N$, $\mathcal{D}^\prime = \{(X_i^\prime, Y^\prime_i)\}_{i=1}^{N^\prime}$, Prototype Learning Algorithm $P$, Feature Importance Function $\phi: \mathcal{X}\times \mathcal{Y} \rightarrow \mathcal{R}^{|\mathcal{X}|}$ based on RID \cite{donnelly2023rashomon}
        \State Determine the $M$ most salient prototypes in $\mathcal{D}$ using the prototype learning algorithm $P$ 
        \For{Prototype: $Z_p = (X_p,Y_p)$ in the set of $M$ learned prototypes}
        \State $\mathcal{D}_{\delta} \rightarrow \{Z = (X,Y) \in \mD| d(Z,Z_p) \leq \delta\}$ \Comment{Examples in $\mD$ close to prototype $Z_p$}
        \State $\mathcal{D}^\prime_{\delta} \rightarrow \{Z^\prime = (X^\prime,Y^\prime) \in \mD^\prime| d(Z^\prime, Z_p) \leq \delta\}$\Comment{Examples in $\mD^\prime$ close to prototype $Z_p$}
        \State $S \rightarrow \emptyset$ 
        \For{Feature $j$ in set of features}
        % \State $s_{ij} = 0$
        % \For{$(X,Y) \in \mathcal{D}_{\delta}$}
        \State $s_{\mD}\rightarrow \mathbb{E}_{(X,Y) \in \mD_\delta}[s_j(\mD,\mD,X_p,Y_p, X, Y)]$\Comment{Equation~\ref{eqn:scoring_fn} for Dataset $\mD_\delta$ - the average score for the neighbors in $\mD$}
        \State $s_{\mD^\prime}\rightarrow \mathbb{E}_{(X^\prime,Y^\prime) \in \mD^\prime_\delta}[s_j(\mD,\mD^\prime,X_p,Y_p, X^\prime, Y^\prime)]$
        \Comment{Equation~\ref{eqn:scoring_fn} for Dataset $\mD^\prime_\delta$ - the average score for the neighbors in $\mD^\prime$}
        \State $s_{total} = s_{\mD} + s_{\mD^\prime}$ 
        \State Append score $s_{total}$ to $S$
        \EndFor
        \State Choose the array indices $[m_1, ...m_K]$ in $S$ with the $K$ lowest scores. These are the $K$ chosen features.
        \State $X_p^{partial} \rightarrow X_p[m_1, ...m_K]$. 
        \EndFor
        \\ 
        \Return $M$ partial prototypes, each with $K$ features 
    \end{algorithmic}
\end{algorithm}
We note that having a feature importance function is not strictly necessary for the scoring mechanism and may only be used if the dataset is labelled. Otherwise, one can simply set the parameters $c_1$ and $c_2$ to $0$ and work only with the value stability desiderata. In Section~\ref{sec:tab_heloc_prototype} of this paper, we will demonstrate examples of partial prototypes for a few real-world tabular datasets. In the appendix (Section~\ref{sec:partial_prototypes_analysis}), we also share recommendations for choosing an appropriate value of $K$. In particular, a large value of $K$ will provide the user with a larger prototype vector, making it less interpretable but more expressive. However, a very small value of $K$ may not necessarily preserve the NSDDs and NSPDs, degrading the quality of the explanation. 

\subsubsection{Case Study 1: Low Dimensional Tabular Data - Adult Dataset}
\label{sec:tab_adult_prototype}
In this section, we construct prototypical explanations for the Adult dataset, employing the NSPD and NSDD methods. The setup is the same as in Section~\ref{sec:fi_exp_adult} - we are comparing Adult male and female datasets. This example is only three-dimensional (so we do not require complex dimension reduction), and prototypes will be chosen in a simple heuristic manner based on feature percentiles and depth-2 decision trees. 
% We will qualitatively compare our results with those of \citet{dist_shifts},  who use an optimal transport formulation, for the same datasets. 
To construct prototypes, we first defined $3$ categories of education levels: lower, medium, and high. These correspond to the $10^{\textrm{th}}$, $50^{\textrm{th}}$, and $90^{\textrm{th}}$ percentiles of education years in the male dataset $\mD$. Note that we could have also chosen the female dataset for constructing prototypes -- there is nothing inherently special about our choice here. We categorised age in the same manner as education. $9$ prototypes were then constructed, corresponding to all possible combinations of education level and age. To construct an income feature, we trained a shallow decision tree classifier on $\mD$ to predict if income $\geq \$50k$ from age and education level. Each prototype was then passed through this decision tree and the tree's prediction (the majority vote in the leaf) was used as the income feature for the prototype.

\begin{figure}[H]
    \centering
    \includegraphics[width=0.9\textwidth]{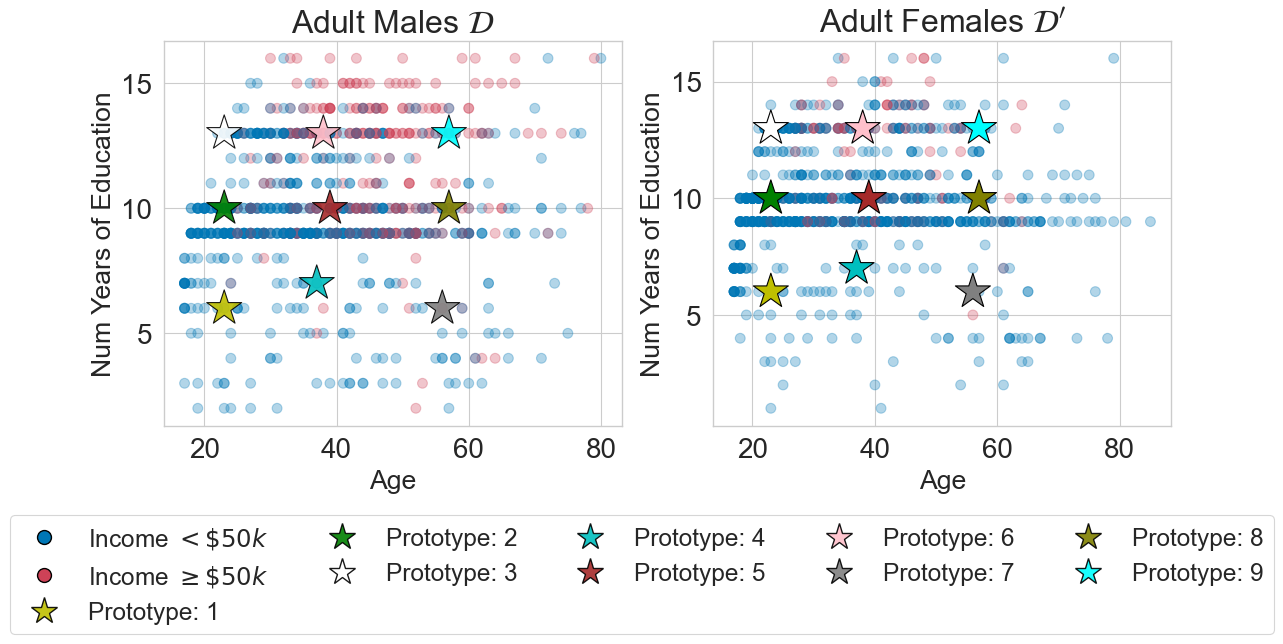}
    \caption{Visualization of the Adult datasets for males ($\mD$) and females ($\mD^\prime$) in 2-D space, where only two features -- \texttt{Age} and \texttt{Num Years of Education} -- are considered. The colors of each point correspond to its class label.}
    \label{fig:adult_pacmap}
\end{figure}
\begin{table}[H]
\centering
\resizebox{0.65\textwidth}{!}{
\begin{tabular}{ccccc}
\hline
& \textbf{Feature 1} & \textbf{Feature 2} & \textbf{Feature 3}\\ \hline
\multirow{2}{*}{\textbf{Prototype 2}} & Age & $\#$ Education Years & Income $\geq \$50k$ \\ 
& 23 & 10 & 0\\ \hline
\multirow{2}{*}{\textbf{Prototype 4}} & Age & $\#$ Education Years & Income $\geq \$50k$ \\ 
& 38 & 7 & 0\\ \hline
\multirow{2}{*}{\textbf{Prototype 6}} & Age & $\#$ Education Years & Income $\geq \$50k$ \\ 
& 40 & 13 & 1 \\ \hline
\multirow{2}{*}{\textbf{Prototype 8}} & Age & $\#$ Education Years & Income $\geq \$50k$ \\ 
& 59 & 7 & 0\\ \hline
\end{tabular}
}
\captionsetup{singlelinecheck=off}
\caption[]{A few prototypes from the Adult male dataset. The NSPD and NSDD for both datasets are computed using Euclidean distance metric over the normalized version of the datasets and the prototypes. We perform normalization of both the prototype and the datasets by using the average and standard deviation of Age and $\#$Education Years from the Male dataset. The binary income feature is not normalized. }
\label{tab:prototypes_adult}
\end{table}
\begin{figure}[H]
    \centering
    \includegraphics[width=0.6\textwidth]{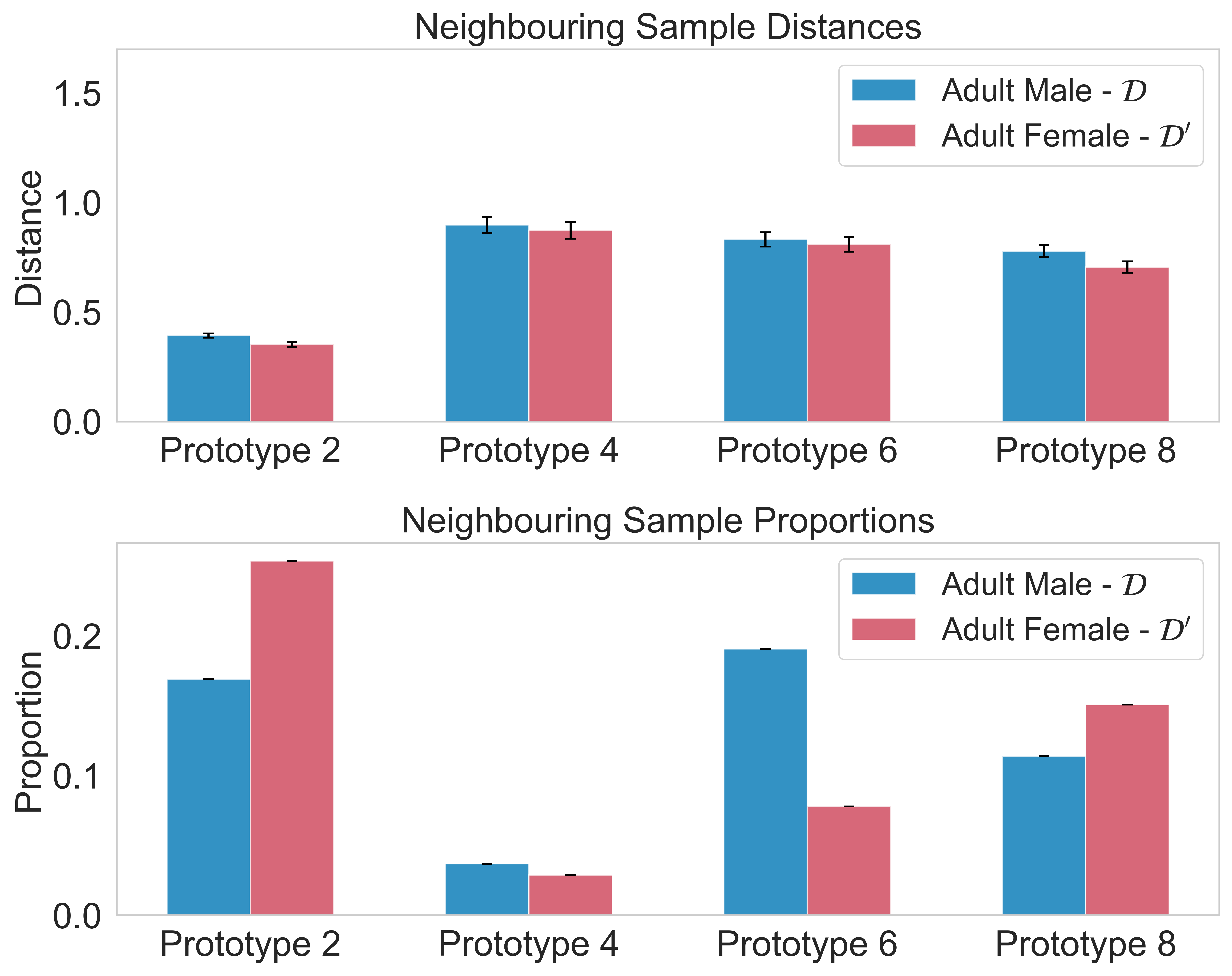}
    \caption{NSPD and NSDD for Adult male ($\mD$) and female ($\mD^\prime$) datasets. Both datasets have similar average distance to the prototypes, but the proportions of examples in the neighborhood of a given prototype are very different. For instance, there are a disproportionately high number of male examples in the neighbourhood of Prototype $6$.}
    \label{fig:adult_neighbouring_sample_distances}
\end{figure}
A visualization of the Adult male and female datasets is seen in Figure~\ref{fig:adult_pacmap}. We can now interpret the NSPD and NSDD for the datasets in terms of these prototypes. To facilitate comparison with \citet{towards_explain}, consider the prototype corresponding to middle aged individuals with a bachelor's degree who earn more than $\$50k$ (i.e., education = $13$, age = $38$, income = $1$). This is marked as Prototype $6$ in Table~\ref{tab:prototypes_adult}. Figure~\ref{fig:adult_neighbouring_sample_distances} shows that there are comparatively fewer examples of this archetype in the female dataset than in the male dataset. An explanation is therefore:
\textit{Compared to the male dataset, the female dataset contains fewer individuals who have a bachelors degree, are middle aged, and earn a high income.}
Similar comparisons can be made for other prototypes.
\subsubsection{Case Study 2: High-Dimensional Tabular Data - HELOC Dataset}
\label{sec:tab_heloc_prototype}
We now construct prototypical explanations for the HELOC dataset. Our datasets $\mD$ and $\mD^\prime$ are the same as in Section~\ref{sec:fi_exp_heloc}. Keeping $\mD$ as the reference dataset, we define the prototypes to be the cluster centers in $\mD$ obtained after K-means clustering
on the high dimensional space and projecting them to a lower dimensional space using PaCMAP \citep{pacmap}. 
To generate these PaCMAP projections, we combined $\mD$ and $\mD^\prime$, ran PaCMAP on this combined dataset, and plotted the lower dimensional datasets separately. The PaCMAP visualizations serve as explanations on their own; because PaCMAP preserves the global and local structure of datasets \citep{pacmap}, visualizing them on a common projected space enables us to understand the cluster structure and relative shifts qualitatively. Even a bird's eye view of the datasets using PaCMAP provides us with very useful information. First, both datasets have similar structures in the feature space, implying that their features are likely to take on the same range of values. Another indication is the larger presence of people who defaulted on their loan in the higher risk dataset $\mD^\prime$ (i.e. class $1$ labels).  
\begin{figure}[H]
    \centering
    \includegraphics[width=0.5\textwidth] {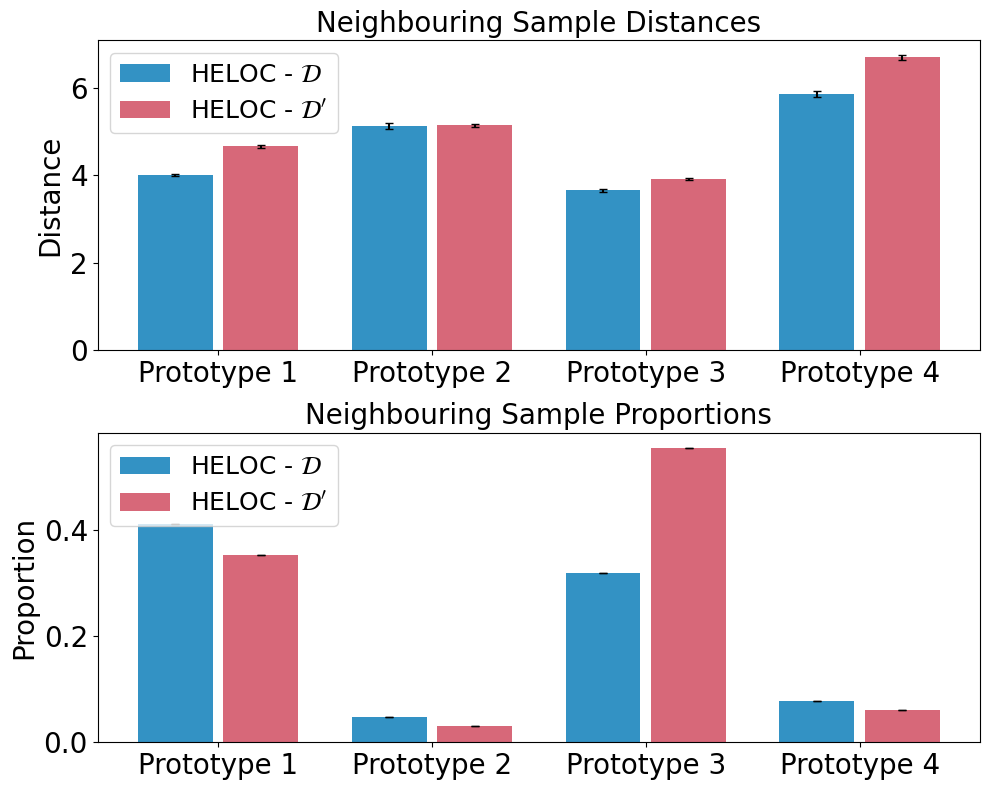} 
    \caption{NSPD and NSDD for the HELOC datasets $\mD$ and $\mD^\prime$. The distance metric used is the Euclidean distance in high dimensional space. Because PaCMAP is structure-preserving \citep{pacmap}, the distance metrics in low and high-dimensional space will be very similar. $\mD$ contains fewer examples that are close to Prototype $3$ compared to $\mD^\prime$, but the average distance to the prototype is similar. Similar types of conclusions can be made for other prototypes. This analysis enables the user to focus on certain neighbourhoods where $\mD$ and $\mD^\prime$ are most different.}
    \label{fig:heloc_prototype}
\end{figure}
\FloatBarrier

\begin{table}[H]
\resizebox{\textwidth}{!}{
\begin{tabular}{cccccc}
\hline
& \textbf{Feature 1} & \textbf{Feature 2} & \textbf{Feature 3} & \textbf{Feature 4} \\ \hline
\multirow{2}{*}{\textbf{P1}} & NumRevolvingTradesWBalance & PercentInstallTrades & NumInstallTradesWBalance & NumInqLast6M \\ 
& 3 & 0 & 1 & 1 \\ \hline
\multirow{2}{*}{\textbf{P2}} & MSinceOldestTradeOpen & NumTrades60Ever2DerogPubRec & NumTrades90Ever2DerogPubRec & NumRevolvingTradesWBalance \\ 
& 1 & 1 & 97 & 5 \\ \hline
\multirow{2}{*}{\textbf{P3}} & NumSatisfactoryTrades & NumTrades60Ever2DerogPubRec & MSinceOldestTradeOpen & PercentInstallTrades \\ 
& 1 & 1 & 7 & 0 \\ \hline
\multirow{2}{*}{\textbf{P4}} & NumTotalTrades & NumInqLast6M & MSinceMostRecentInqexcl7days & MaxDelqEver \\ 
& 0 & 2 & 2 & 18 \\ \hline
\end{tabular}
}
\captionsetup{singlelinecheck=off}
\caption[]{Understanding the $K = 4$ most salient features for each prototype in $\mD$. We can interpret this jointly with Figures~\ref{fig:heloc_prototype} and \ref{fig:heloc_figs}. Here is a dataset-level explanation in terms of Prototype $3$: the dataset of individuals with lower \texttt{ExternalRiskEstimate} (i.e., $\mD$) has a lower proportion of individuals with approximately the following profile: 
\begin{itemize}
    \item Num Satisfactory Trades $=1$
    \item 1 trade more than $60$ days past due
    \item $7$ Months since last trade
    \item No installment trades
\end{itemize}
Similar interpretations can be made for other prototypes.}
\label{tab:prototypes_heloc}
\end{table}
Given the prototypes in $\mD$, the explanation compares the NSPD and NSDD of $\mD$ and $\mD^\prime$ for these prototypes. From Figure~\ref{fig:heloc_prototype}, we can also analyze a small subset of salient features for a prototype (aka the partial prototype) to understand the properties of the prototype and its neighbourhood in $\mD$ and $\mD^\prime$ in an interpretable manner.

\FloatBarrier
\subsubsection{Case Study 3: Time Series Medical Data - Cardiac Signals}
\label{sec:cardiac_signals}
Cardiac signals are essential in clinical diagnostics and disease screening. The advancement of machine learning and deep learning has facilitated numerous studies to automate cardiac disease detection, further improving reliability and efficiency. However, the scarcity of large open-access datasets poses a challenge for practitioners and machine learning researchers. Given this context, the need for accurate and high-quality synthetic cardiac data becomes imperative. In this experiment, we aim to showcase our method by comparing synthetic data against real-world data and derive actionable items to improve synthetic data generation.
\begin{itemize}
    \item \textbf{Dataset $\mD$:} Photoplethysmography (PPG) was chosen as a representative form for time series medical signals due to its rising popularity in recent years as the medium for heart monitoring on wearable devices. In this study, we chose the Stanford PPG dataset, which was collected from subjects wearing smartwatches while performing regular daily activities \cite{stanford}. Using the dataset's signal quality labels, we sampled a subset of 16,058 25-second signals that contain a relatively small amount of noise, each accompanied by an atrial fibrillation (AF) or non-atrial fibrillation (non-AF) label. During preprocessing, signal amplitudes were normalized into the 0-1 range and resampled to have 2400 timesteps in the 25 seconds time frame. We show samples of real PPG signals in Figure~\ref{fig:synth_ppg}.
    \item \textbf{Dataset $\mD^\prime$:} A popular PPG processing and simulation tool, neurkit2 \citep{neurokit}, was used to generate a synthetic PPG dataset containing 3,000 30-second signals for this study. Synthetic signals were created with the addition of varying levels of signal noise and artifacts to mimic realistic conditions. A detailed description of the simulation parameters can be found in appendix Section~\ref{sec:PPG_simulation_parameters}; we also show a few generated synthetic PPG signals in Figure~\ref{fig:synth_ppg}. Signal amplitudes were normalized to the 0-1 range and resampled to have 2400 timesteps.
\end{itemize}

\paragraph{Forming the explanation} The comparisons were conducted using the prototypical explanation method introduced in Section~\ref{sec:method_nspdnsdd}. A 1D-ResNet-34 model is used as the encoder. To accommodate the relatively small $D$ dataset size, we first pre-trained the encoder using a multitask approach. The encoder was trained to optimize both a signal reconstruction MSE loss as part of an autoencoder, and the cross-entropy loss for the AF detection classification task (AF vs$.$ non-AF classification). The pre-trained encoder was then used to train the prototype learning model following the approach in previous work \citep{interpEEG}. 

\begin{figure}[hbt]
% \captionsetup{justification=centering,labelfont=bf}
    \centering
    \begin{subfigure}[b]{0.45\textwidth}
        \centering
        \includegraphics[width=\textwidth,valign=c]{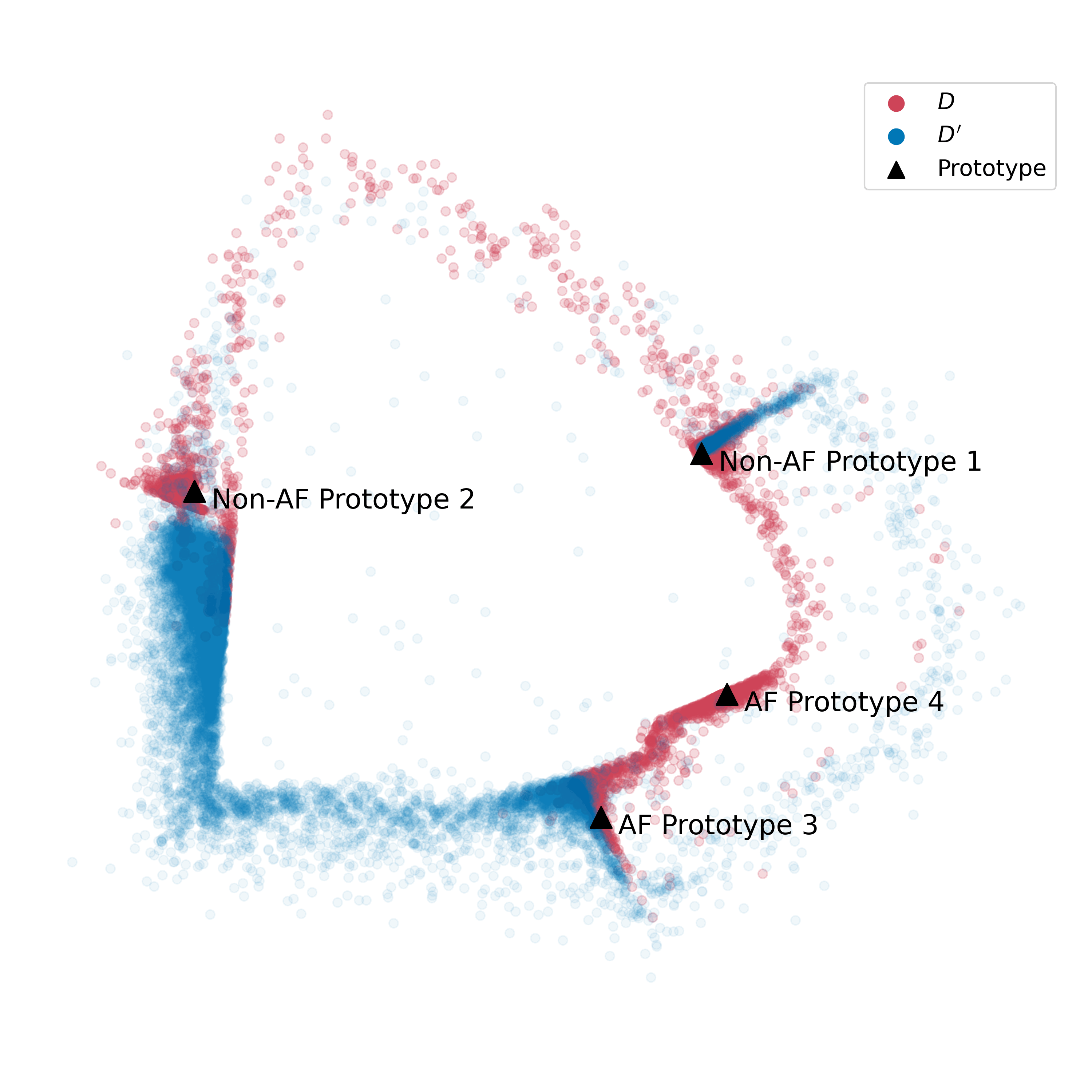}
    \end{subfigure}
    \begin{subfigure}[b]{0.45\textwidth}
        \centering
        \includegraphics[width=\textwidth,valign=c]{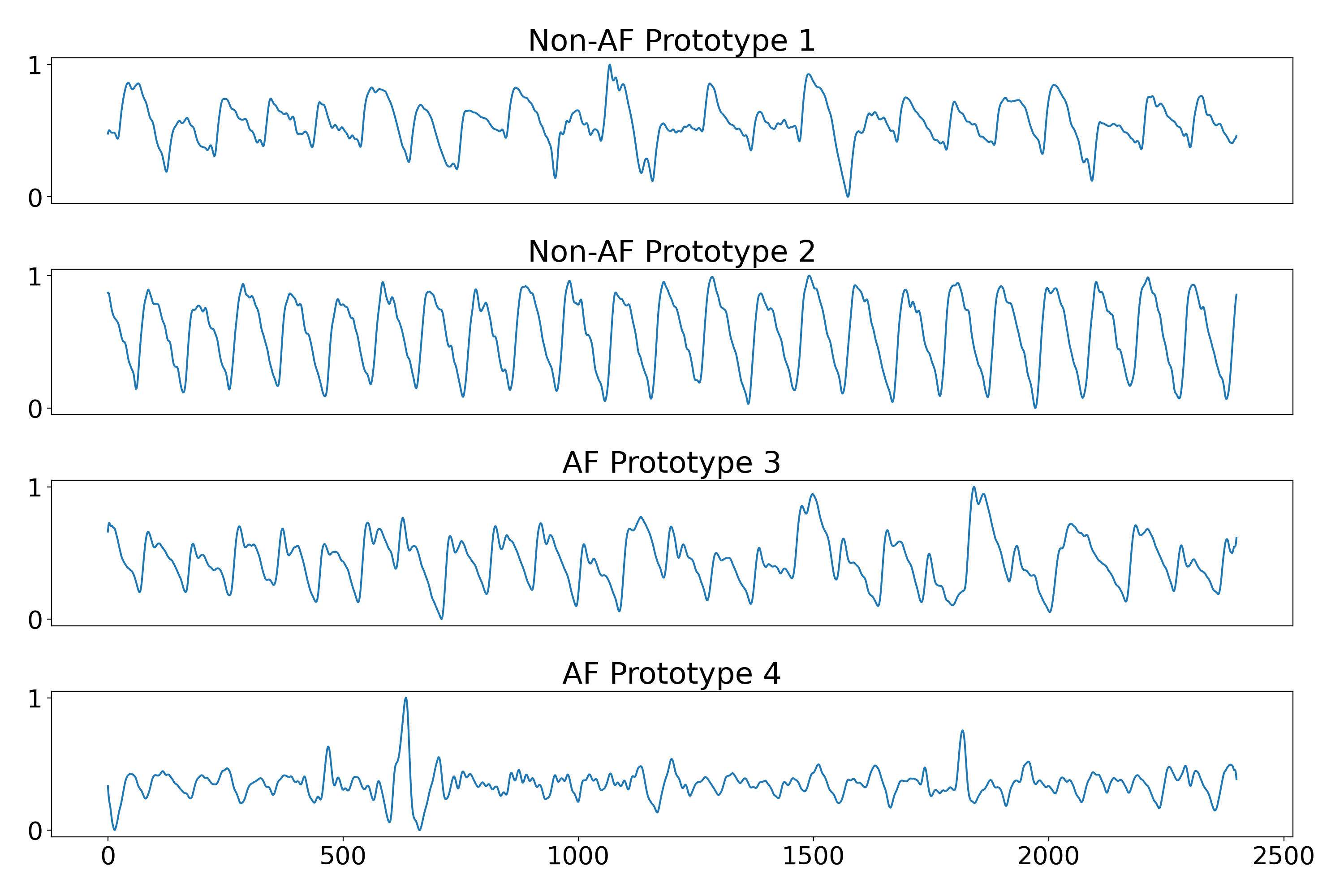}
    \end{subfigure}
    \caption{Visualization of the projections of encoded samples in both $\mD$ and $\mD^\prime$ in the same latent space, including the learned prototypes.}
    \label{fig:ucla_projections}
\end{figure}

\begin{figure}[hbt]
    \centering
    \includegraphics[width=\textwidth]{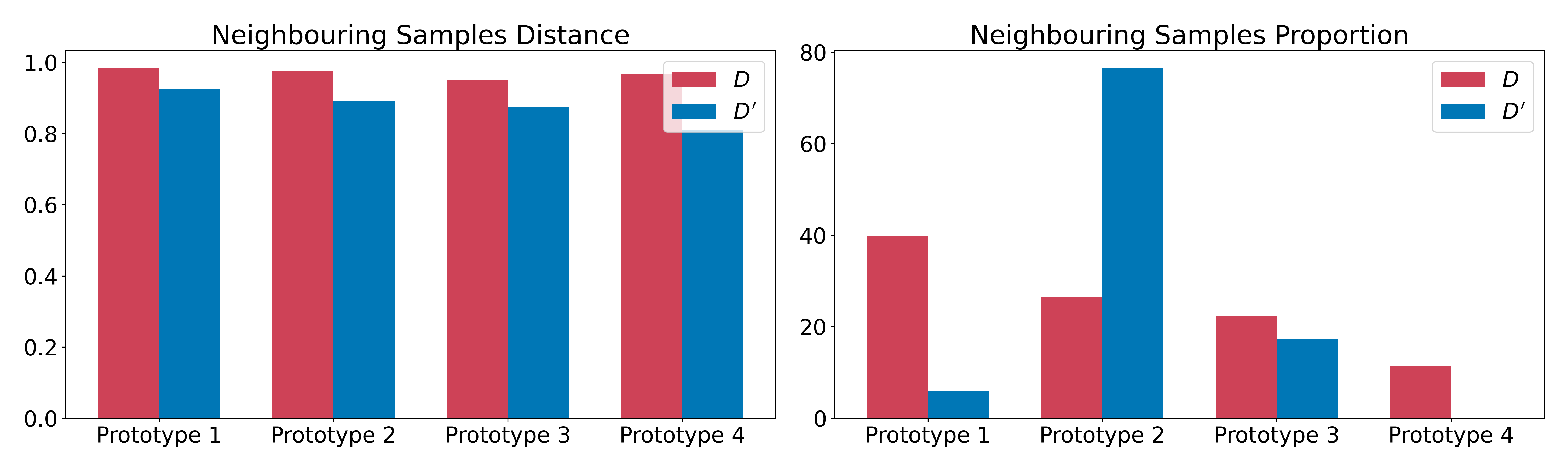}
    \caption{NSPD and NSDD comparison between $\mD$, $\mD^\prime$ sample features for all learned prototypes.  The two datasets differ in that there are very few samples in $\mD^\prime$ that are sufficiently similar to Prototypes $1$ and $4$ in $\mD$; More samples that are similar to Prototype $2$ than that in $\mD$.}
    \label{fig:ucla_syth_rsts}
\end{figure}

\paragraph{Quantitative comparison between prototypical neighborhoods} We are able to visualize the projection of encoded samples in both $\mD$ and $\mD^\prime$. We can observe the difference in coverage of $\mD^\prime$ samples to the $\mD$ samples in the latent space. The learned prototypical samples of $\mD$ are shown in Figure~\ref{fig:ucla_projections}. We calculate the quantitative difference using the NSPD and NSDD metrics defined in Section~\ref{sec:method_nspdnsdd}, and the results are shown in Figure~\ref{fig:ucla_syth_rsts}. From these results, we conclude that the synthetic data generator does generate samples similar to those of the real dataset $\mD$ in terms of latent space distance; however, there is a discrepancy between the number of certain types of signals generated in the synthetic dataset and in the real dataset. This conclusion is supported by the fact that the NSDD is relatively small, \textit{indicating a similarity in features related to AF classification between generated signals and the prototypes comparable to that between the real samples and the prototypes} (shown in Figure~\ref{fig:ucla_syth_rsts}); in addition, we observe large NSPDs for prototypes $1$, $2$ and $4$, indicating that there are \textit{insufficient samples similar to prototype $1$ and $4$}, and \textit{too many samples similar to prototype $2$}. By inspecting the learned prototypes, we could potentially improve the realism and quality of the generated signals by introducing more variable and organic noise corruptions similar to those in prototypes $1$ and $4$, in addition to those in Neurokit 2 (\cite{neurokit}).

\FloatBarrier
\subsection{Prototype-Summarization-Based Explanations}
\label{sec:summarizationprotoexp}
\subsubsection{Introduction}
In this section, we propose another type of prototypical explanation that is different from the neighbourhood-based-explanation method introduced in Section~\ref{sec:protoneigh_explainmethod}. Previously, we learned prototypes in \textit{only one} of the two datasets. Here, we learn prototypes in \textit{both} $\mD$ and $\mD^\prime$ using a modified version of ProtoPNet \citep{protopnet}. The ensuing explanation involves directly comparing prototypes unique to $\mD$ and $\mD^\prime$, providing insights into the differences between the datasets. As we will show, this approach is especially useful for visual and signal-based datasets where differences are not easily discernible through direct inspection. These prototypes represent key samples from $\mD$ and $\mD^\prime$ that best explain the distinctions between the datasets.

\begin{figure}[H]
    \centering
    \includegraphics[width=0.5\textwidth]{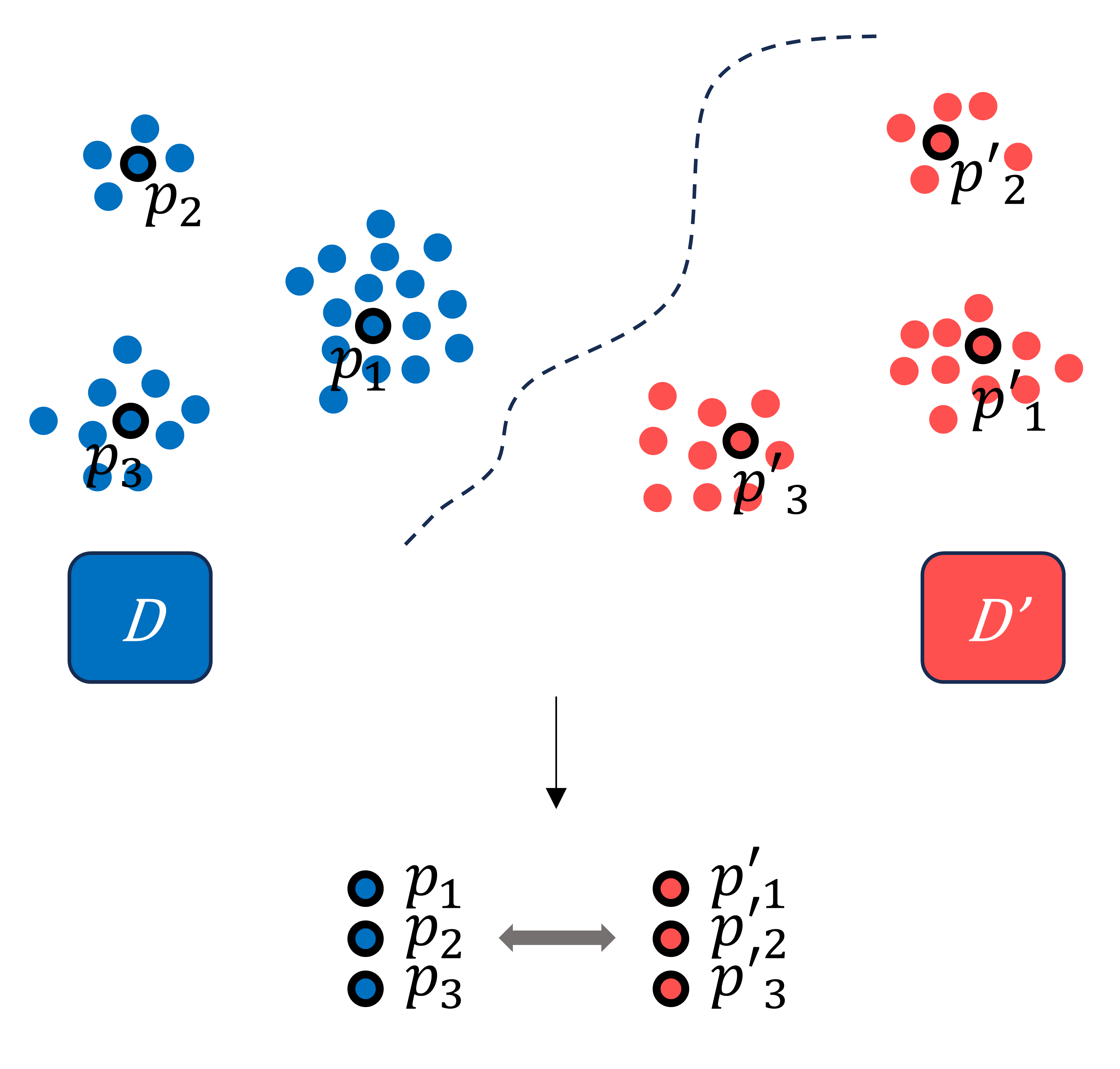}
    \caption{An illustration of the Prototype-Summarization-Based Explanations. First a prototype learning model is trained to classify between dataset $\mD$ and $\mD^\prime$. The learned prototypes $p_1$, $p_2$, $p_3$ from dataset $\mD$ can be used as a summarization of its neighbouring samples and to be compared against prototypes $p^\prime_1$, $p^\prime_2$, $p^\prime_3$ learned in dataset $\mD^\prime$, thus forming an explanation.}
    \label{fig:proto_summary}
\end{figure}

To identify these prototypes, we construct a binary classification task where samples from $\mD$ are labeled as 1 and samples from $\mD^\prime$ as 0. A ProtoPNet or a similar prototype learning network is then trained to distinguish between the two datasets, learning $n_p$ prototypes for $\mD$ and $n_{p^\prime}$ prototypes for $\mD^\prime$. These prototypes encapsulate the unique and distinguishing characteristics of each dataset, enabling users to analyze differences without examining a large number of samples. Figure~\ref{fig:proto_summary} illustrates this process. In subsequent sections, we provide examples demonstrating how prototype summarization explanations can effectively highlight distribution shifts using a small set of representative samples.

\subsubsection{Summarization Prototype Learning for Dataset Comparisons} 
In general, the feature extractor in ProtoPNet is trained by optimizing the following loss terms:
\begin{equation}
    \min_{\omega_g} \left( 
    \text{Cross Entropy} + \lambda_c ~ \ell_\text{clst} + \lambda_s ~ \ell_\text{sep} 
    \right), \textrm{ where}
\end{equation}
\begin{equation}
    \ell_\text{clst} = \frac{1}{n}\sum_{i=1}^{n} \min_{j:\text{class}(\mathbf{p}_j)=y_i} s(f_{\omega_f}(\mathbf{x}_i), p_j), \quad
    \ell_\text{sep} = -\frac{1}{n}\sum_{i=1}^{n} \min_{j:\text{class}(\mathbf{p}_j) \neq y_i} s(f_{\omega_f}(\mathbf{x}_i), p_j).
\end{equation}

Here, $\ell_\text{clst}$ ensures that each training sample is close to a prototype of its class, while $\ell_\text{sep}$ pushes samples away from prototypes of other classes. Additionally, $\ell_\text{ortho}$, an optional term, encourages prototype diversity.

The task of comparing the differences between datasets, which requires the learning of summarization prototypes, differs significantly from the existing literature's prototype learning task -- classification. Instead of associating samples with object classes, samples are labeled based on dataset membership. Although existing loss functions could work for datasets with few or single-semantic concepts (where the object label $\approx$ dataset membership label), they will falter for datasets containing diverse concepts (concepts could be objects, subcategories, art styles, etc.) In such cases, the model would reduce the loss trivially by aligning all samples regardless of their difference in concepts or subcategories within a dataset to the same prototypes, compromising the summarization power.

To address this, we propose an alternative clustering loss:
\begin{equation}
    \label{eq:clst_v2}
    \ell_\text{clst} = \frac{1}{n}\sum_{i=1}^{n} \mathbb{E}_{p_k \in P}[s(f_{\omega_f}(\mathbf{x}_i), p_k)] 
    - \min_{j:\text{class}(\mathbf{p}_j)=y_i} s(f_{\omega_f}(\mathbf{x}_i), p_j).
\end{equation}

This modified loss function encourages a greater separation between the most similar prototype and the average similarity to all prototypes within the same dataset. By doing so, it ensures that each sample strongly aligns with a single prototype while avoiding trivial alignment with others. This adjustment improves prototype summarization for datasets with diverse underlying concepts.

Additionally, to further improve the summarization power of the prototypes and coherence in the latent space, we also introduce a novel prototype affinity-based contrastive learning loss to aid the learning of the model and prototypes without explicit supervision. This learning target does not utilize any object or concept labels for contents within the dataset.

The contrastive prototype loss is designed to align similar samples while distinguishing dissimilar samples. We construct two augmented views from each input image, calculate the similarity matrices from each view's latent feature to each prototype \(P, Q \in \mathbb{R}^{n \times m}\), where \(n\) is the number of samples and \(m\) is the number of prototypes. The similarities are then normalized into an prototype affinity distribution using softmax function with a temperature scaling factor \(T\) (we default to \(T = 0.07\)):
\begin{equation}
p_{ip} = \frac{\exp(P_{ip} / T)}{\sum_{k=1}^m \exp(P_{ik} / T)}, \quad
q_{ip} = \frac{\exp(Q_{ip} / T)}{\sum_{k=1}^m \exp(Q_{ik} / T)},
\end{equation}
where \(p_{ip}\) and \(q_{ip}\) represent the affinity of sample \(i\) to prototype \(p\) in the two respective views, and \(k\) indexes over the \(m\) prototypes.

We measure how aligned two prototype affinity distributions are using cross-entropy. The pairwise cross-entropy between two samples \(i\) and \(j\) is defined as:
\begin{equation}
\text{CE}(P_i, Q_j) = -\sum_{k=1}^m p_{ik} \log(q_{jk}).
\label{eq:eq12}
\end{equation}
The pairwise cross-entropy matrix is then computed as:
\begin{equation}
\text{Pairwise\_CE}_{ij} = \text{CE}(P_i, Q_j) = -\sum_{k=1}^m p_{ik} \log(q_{jk}).
\label{eq:eq13}
\end{equation}
Finally the loss term is defined as follows:
\begin{equation}
\ell_\text{contrast} = -\frac{1}{n} \sum_{i=1}^n \log \left( \frac{\exp(\text{Pairwise\_CE}_{ii})}{\sum_{j=1}^n \exp(\text{Pairwise\_CE}_{ij})} \right).
\label{eq:eq14}
\end{equation}

Here, \(\text{Pairwise\_CE}_{ii}\) represents the similarity between the same sample across the two views, while \(\text{Pairwise\_CE}_{ij}\) for \(i \neq j\) represents the similarity between different samples. Since we want to increase the similarity between two positive pair images' prototype affinity distribution by minimizing $\ell_\text{contrast}$, we remove the negative sign in Equation~\ref{eq:eq13} when calculating $\ell_\text{contrast}$.

This formulation encourages high similarity between matching pairs with the same concepts while penalizing similarity to other samples with different concepts. By using this loss term, the model will learn a more coherent neighborhood, thus increasing the summarization power and faithfulness of the learned prototypes.

The $\ell_\text{contrast}$ and $\ell_\text{clst}$ are optional for training the summarization-based prototypes on single-semantic datasets, but it is necessary and beneficial to use these loss terms for cases with complex concepts and subcategories involved in either dataset.

\begin{figure}[H]
    \centering
    \includegraphics[width=0.5\textwidth]{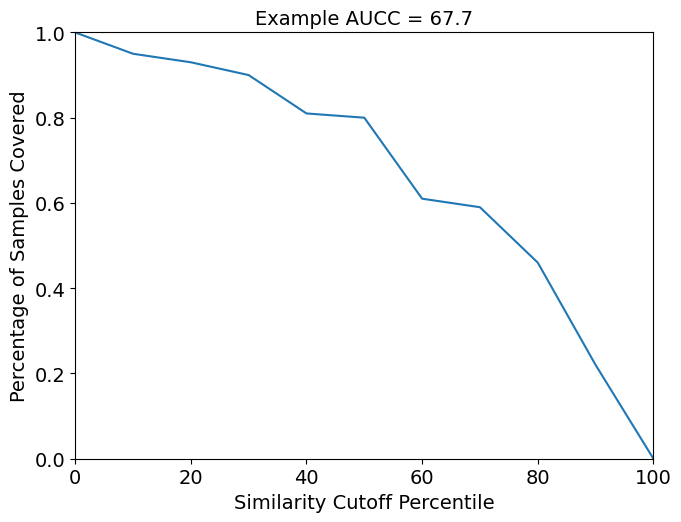}
    \caption{An example of the coverage curve and area-under-the-coverage-curve. The higher the AUCC, the more representative the given set of prototypes are of the entire dataset. However, while having too many prototypes gives higher coverage, it increases explanation complexity. We illustrate the tradeoff between explanation interpretability and prototype representativeness in Section \ref{sec:prototypical_explanations_appendix_analysis}}
    \label{fig:aucc_example}
\end{figure}

\paragraph{Quantitative metric for prototype coverage} We introduce an evaluation metric to quantify how well the learned set of summarization prototypes covers the whole dataset. For each sample, we consider it converted by prototype set, if its similarity to any one of the prototypes is higher than a certain threshold. For a prototype, we use the $X$th percentile ($X \in \{0-100\}$) of all samples' similarities to this prototype as the cutoff threshold (i.e., similarity above threshold indicates a sample is covered by the prototype). We calculate the percentage of the covered samples in both datasets $\mD$ and $\mD^\prime$ for each threshold, and derive an area-under-the-coverage-curve ($\textbf{AUCC}$) as the final coverage score for a particular learned set of summarization prototypes for $\mD$ and $\mD^\prime$. An example coverage curve and its area-under-the coverage-curve value is shown in Figure~\ref{fig:aucc_example}. The maximum AUCC score is $100$, and the minimum is $0$.

\subsubsection{Case Study 1: Time Series Data - Human vs Machine Audio}
\label{sec:exp_audio}
\begin{figure}[hbt!]
    \begin{subfigure}[b]{0.5\textwidth}
        \centering
        \includegraphics[width=0.85\textwidth]{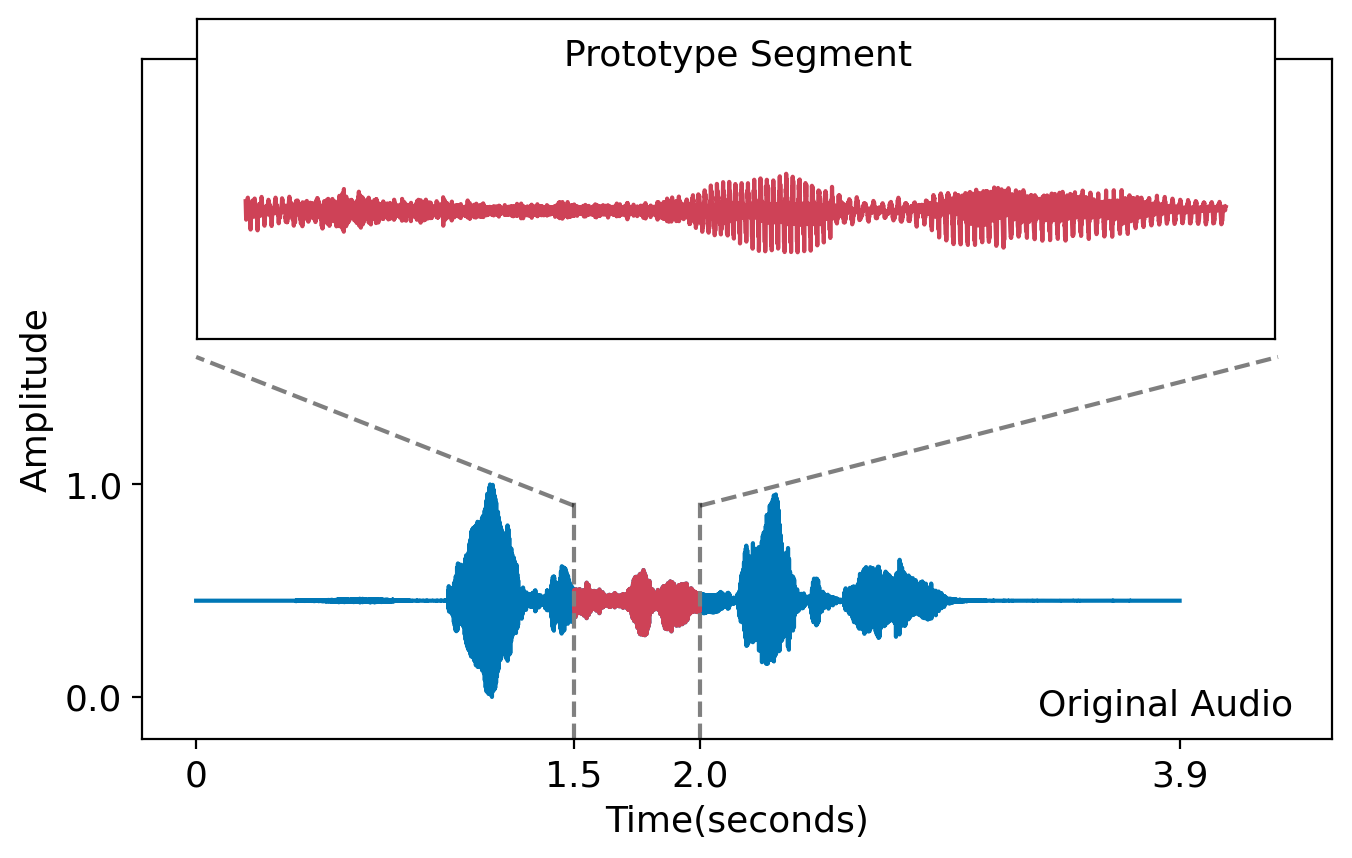}
        \caption{Human audio prototype 1 from $\mD$.}
        \label{fig:humanaudioproto1}
    \end{subfigure}
    \begin{subfigure}[b]{0.5\textwidth}
        \centering
        \includegraphics[width=0.85\textwidth]{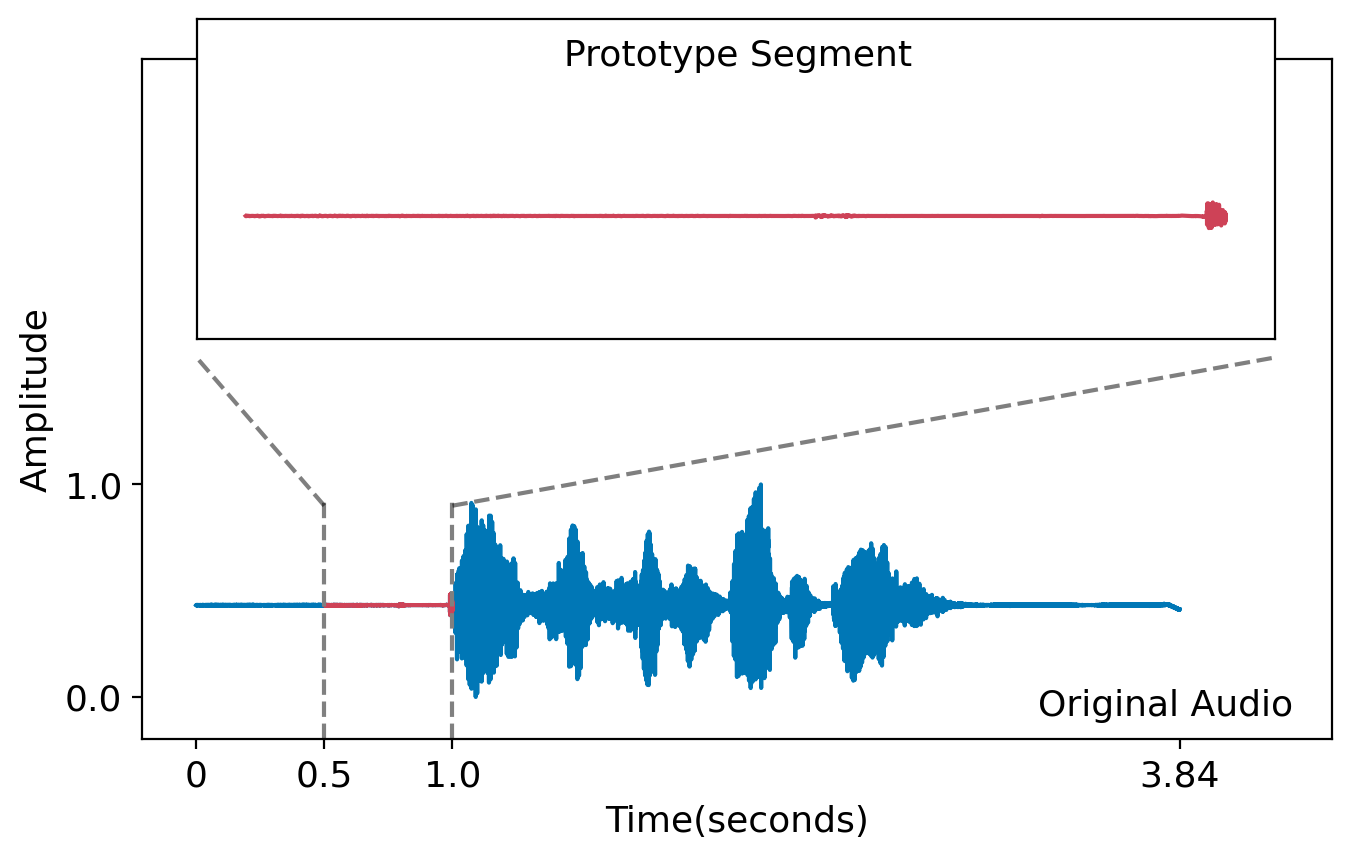}
        \caption{Human audio prototype 2 from $\mD$.}
        \label{fig:humanaudioproto2}
    \end{subfigure}
    \newline
    \begin{subfigure}[b]{0.5\textwidth}
        \centering
        \includegraphics[width=0.85\textwidth]{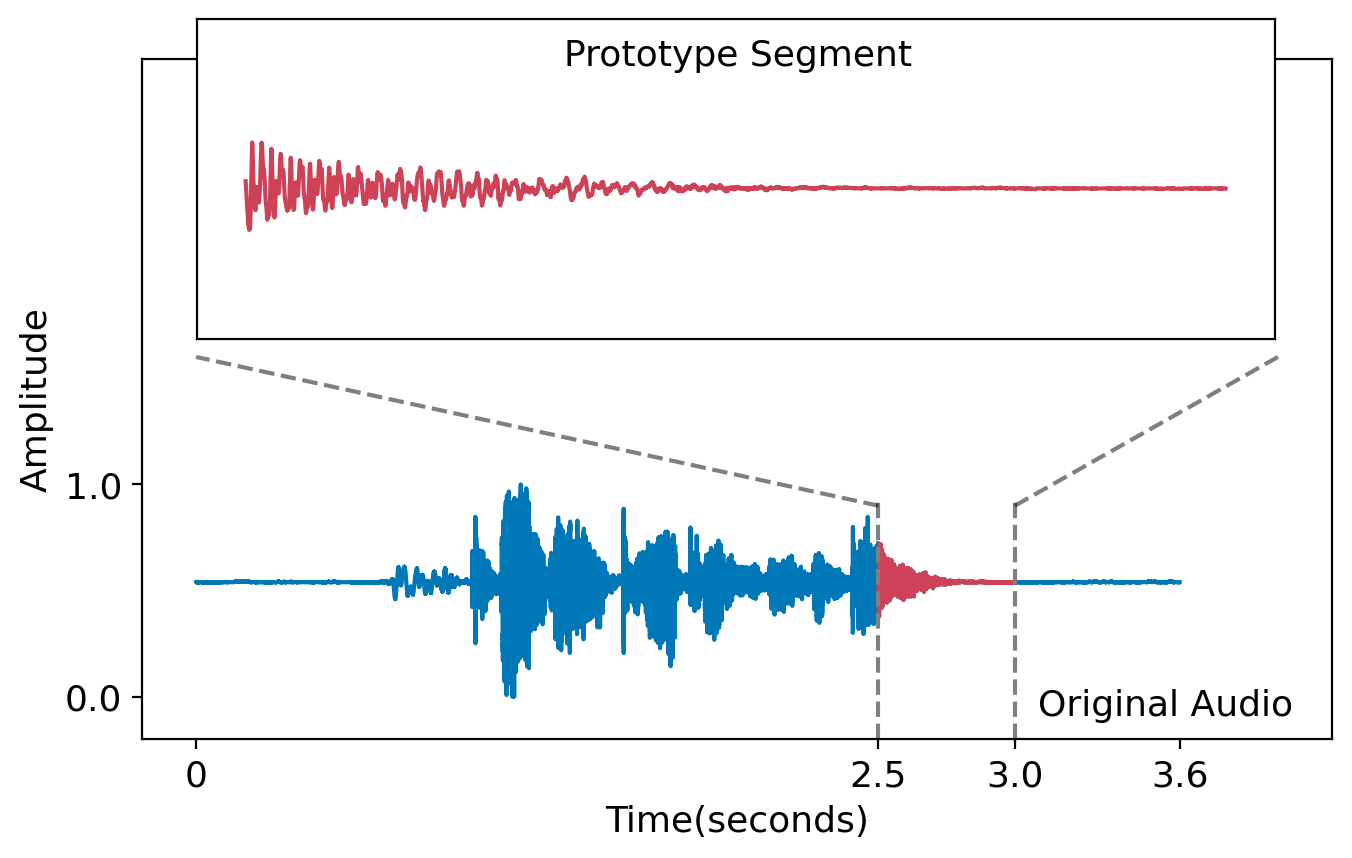}
        \caption{Human audio prototype 3 from $\mD$.}
        \label{fig:humanaudioproto3}
    \end{subfigure}
    \begin{subfigure}[b]{0.5\textwidth}
        \centering
        \includegraphics[width=0.85\textwidth]{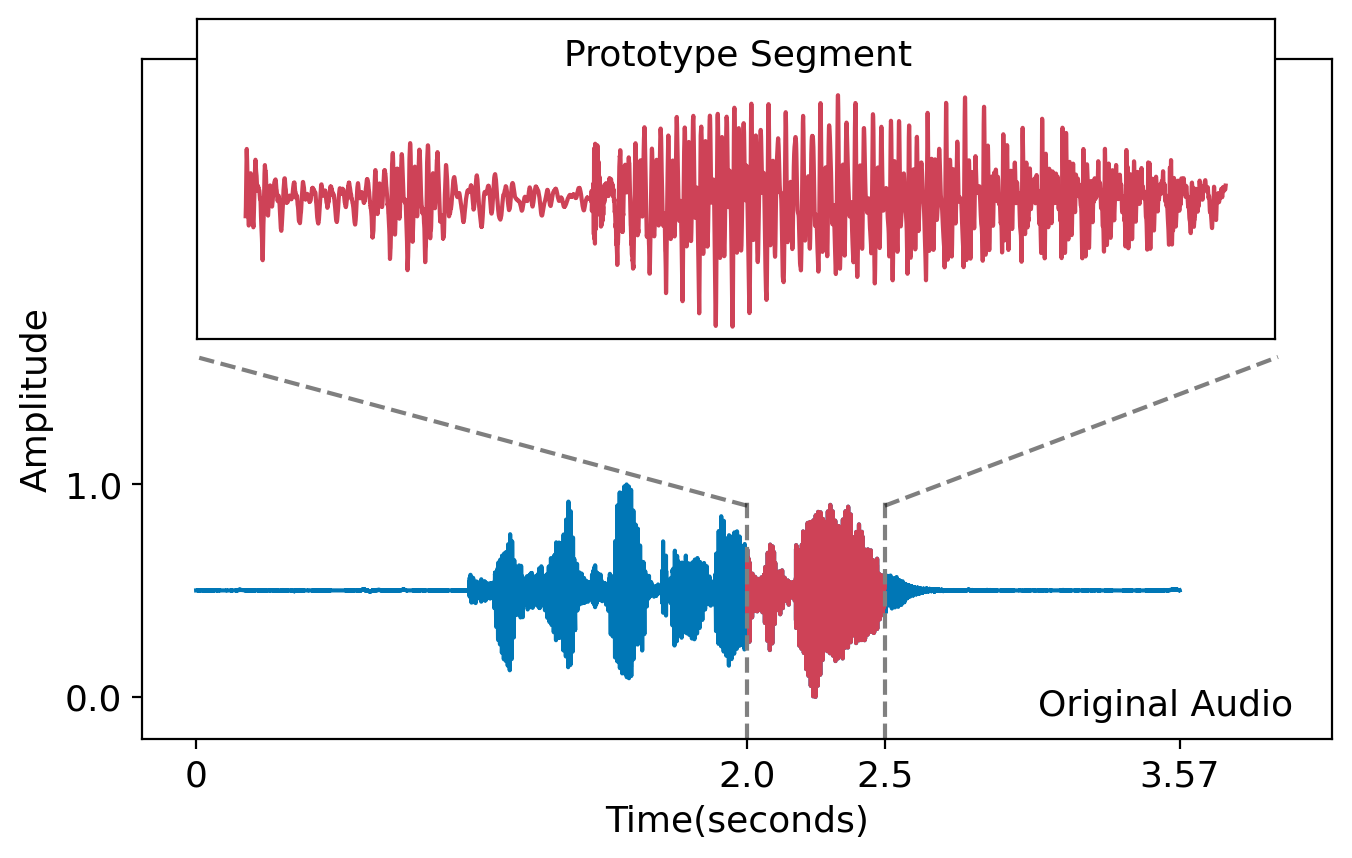}
        \caption{Human audio prototype 4 from $\mD$.}
        \label{fig:humanaudioproto4}
    \end{subfigure}

    \begin{subfigure}[b]{0.5\textwidth}
        \centering
        \includegraphics[width=0.85\textwidth]{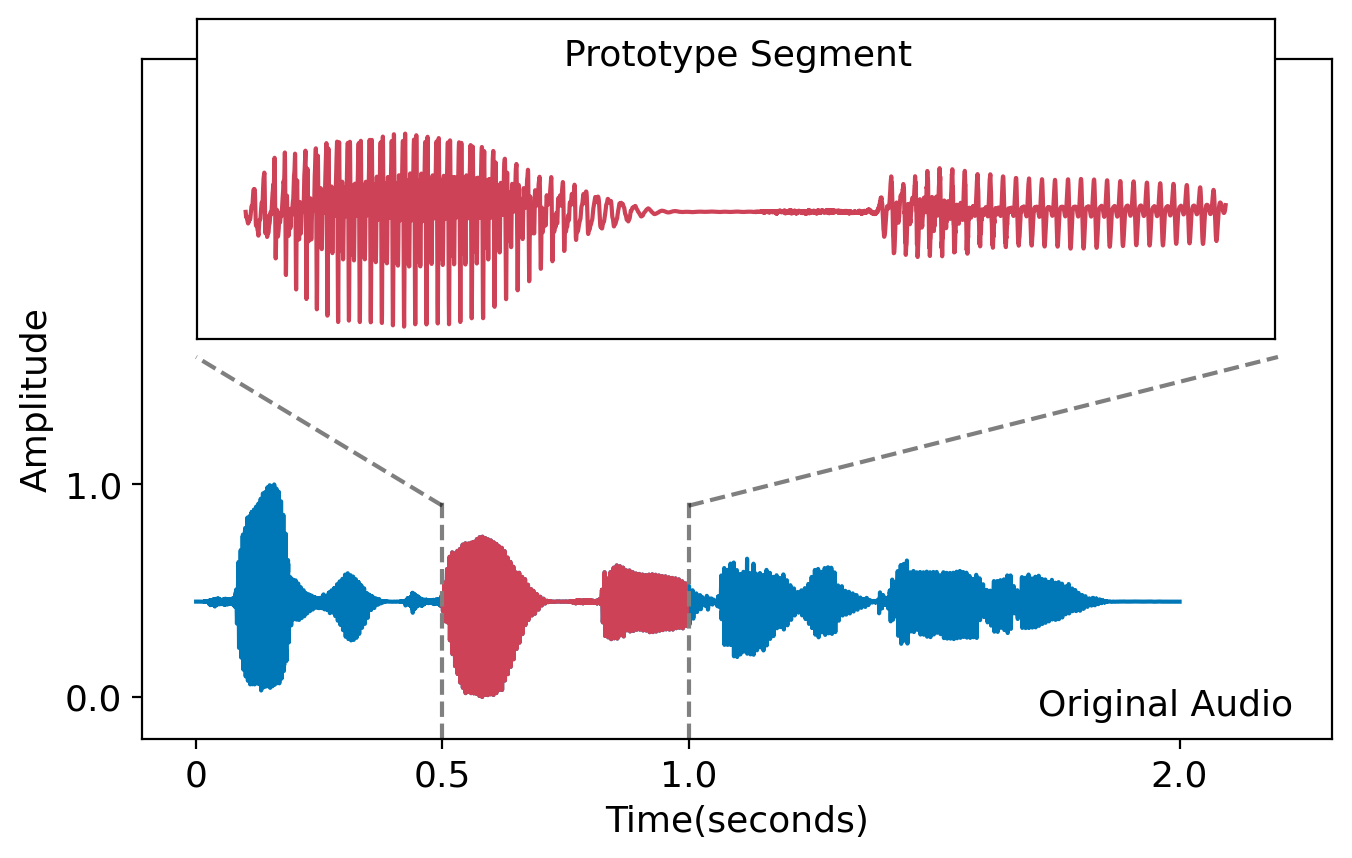}
        \caption{Machine audio prototype 1 from $\mD^\prime$.}
        \label{fig:machineaudioproto1}
    \end{subfigure}
    \begin{subfigure}[b]{0.5\textwidth}
        \centering
        \includegraphics[width=0.85\textwidth]{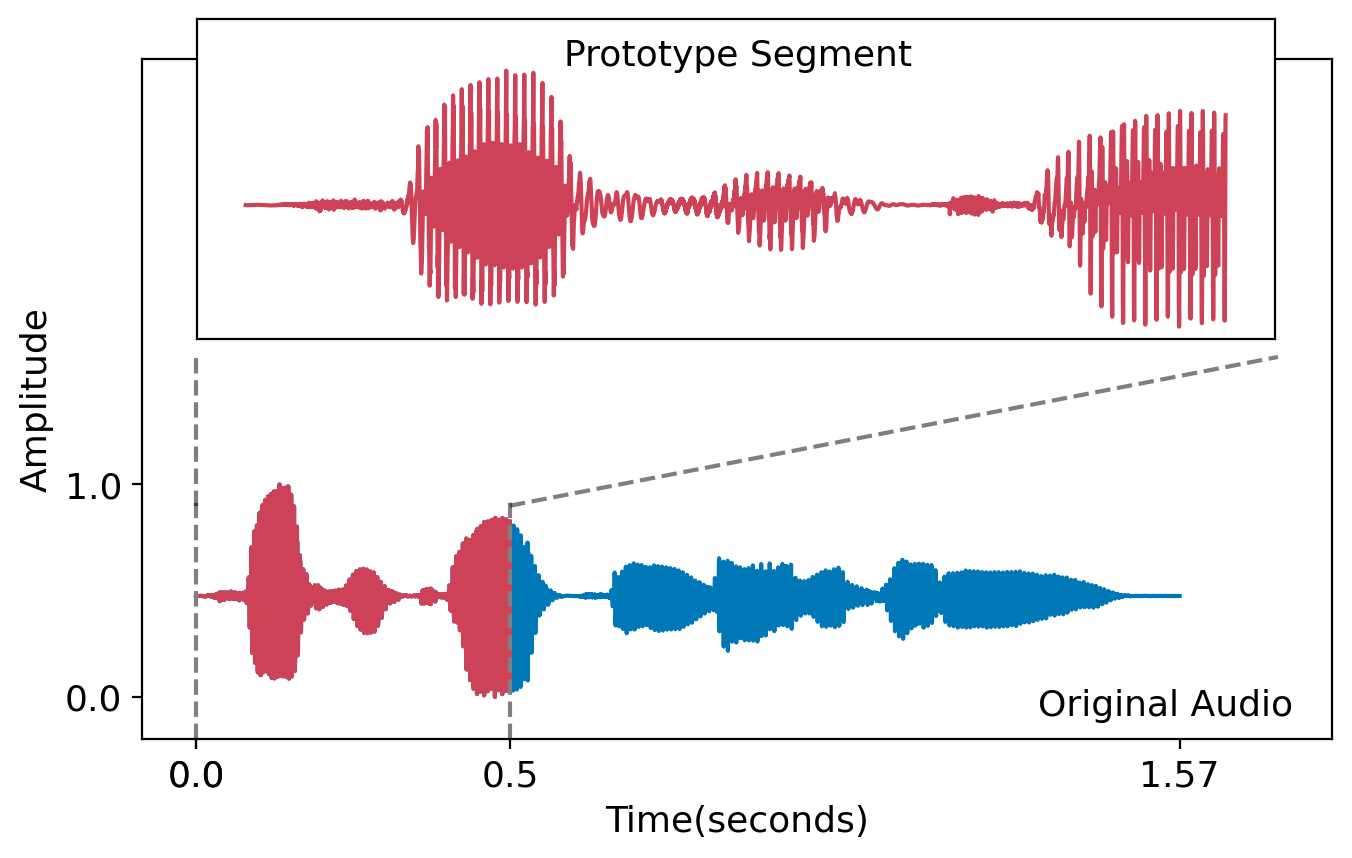}
        \caption{Machine audio prototype 2 from $\mD^\prime$.}
        \label{fig:machineaudioproto2}
    \end{subfigure}
    \newline
    \begin{subfigure}[b]{0.5\textwidth}
        \centering
        \includegraphics[width=0.85\textwidth]{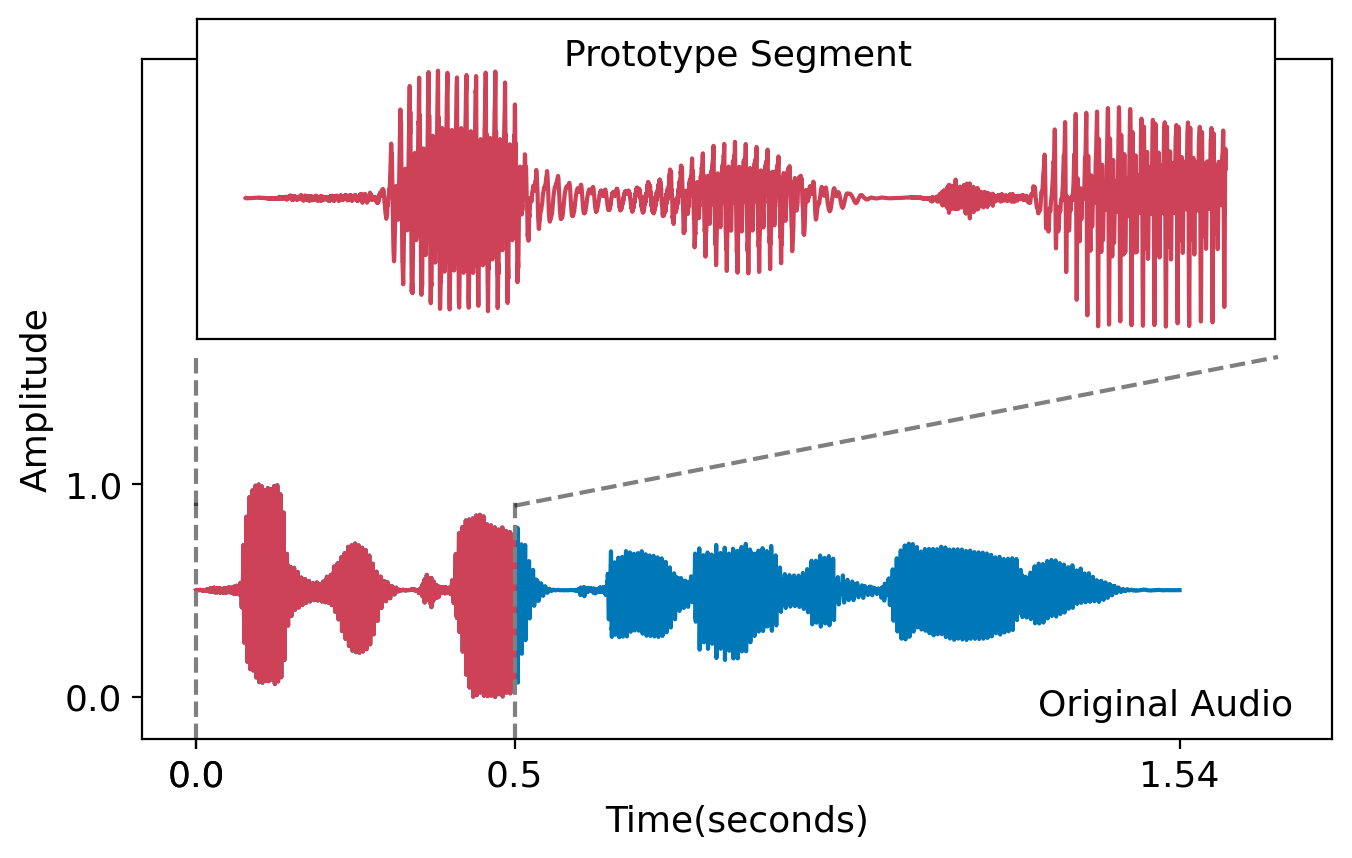}
        \caption{Machine audio prototype 3 from $\mD^\prime$.}
        \label{fig:machineaudioproto3}
    \end{subfigure}
    
    \caption{Learned audio segment prototypes from $\mD$ and $\mD^\prime$. The learned prototypes grasp the most obvious human audio characteristic, human tends to wait before speaking, whereas AI starts speaking right away. Some of the learned human prototypes represent the silent waiting period in human audio. The humans also tend to speak with varying speed, loudness, and pitch as opposed to the machine's paced and monotonic speech, reflected by the constant peak-to-peak interval in machine audio and the very gradual changes in machine audio amplitudes.}
    \label{fig:audioprotos}
\end{figure}

% \subfile{subfiles/exp_audio}
In this section, we compare the following two datasets: 
\begin{itemize}
    \item \textbf{Dataset $\mD$} We use the human emotional speech audio dataset RAVDESS \citep{RAVDESS}, which contains 928 audios of 24 different human speakers speaking two statements with a range of emotions. Statements contain ``Kids are talking by the door,'' 02 = ``Dogs are sitting by the door.'' Audios labeled neutral, happy, sad, and angry were included in this experiment. Figure~\ref{fig:audioprotos} shows a human audio example.
    \item \textbf{Dataset $\mD^\prime$} For this study, we leveraged the Coqui TTS \citep{coquitts} to generate AI audio. The AI audio is generated using 58 AI speakers, and includes the same set of emotions as the human in $\mD$, speaking the same two statements. We generated 864 machine-generated audio signals. Figure~\ref{fig:audioprotos} shows a machine-generated audio example.
\end{itemize}

\paragraph{Forming the explanation} We compute the ``Prototype-summarization-based explanations'' described in Section~\ref{sec:summarizationprotoexp}. To do this, we trained a binary human vs$.$ machine-generated audio prototype-based classifier by fine-tuning the pretrained HUBERT audio classification model \citep{hubertaudio}. 1434 audios were used for training and 358 audios were used for evaluations. During training, each audio was first sliced into 4 equal-length continuous segments and fed into the network, and each prototype is a 0.5-second segment of the audio signal. For each input audio, we use its most similar segment to each prototype to represent the affinity of the audio to the prototype. This means we can pinpoint the specific differences between human audio and machine audio, rather than just comparing entire audio signals, which is less informative.

\begin{figure}[hbt]
\centering
\captionsetup{justification=raggedright}
\begin{minipage}[b]{.39\textwidth}
  \centering
  \includegraphics[width=1\linewidth]{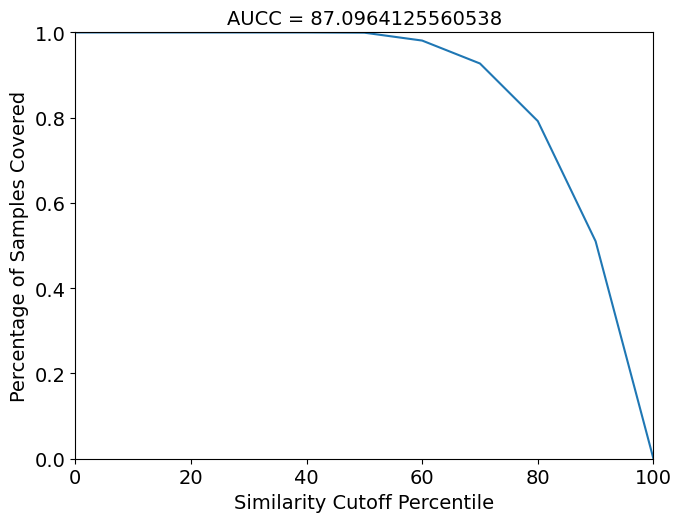}
  \caption{Coverage curve and AUCC score for the human-vs-machine audio case study.}
  \label{fig:aucc_audio}
\end{minipage}\;\;
\begin{minipage}[b]{.55\textwidth}
  \centering
  \includegraphics[width=1\linewidth]{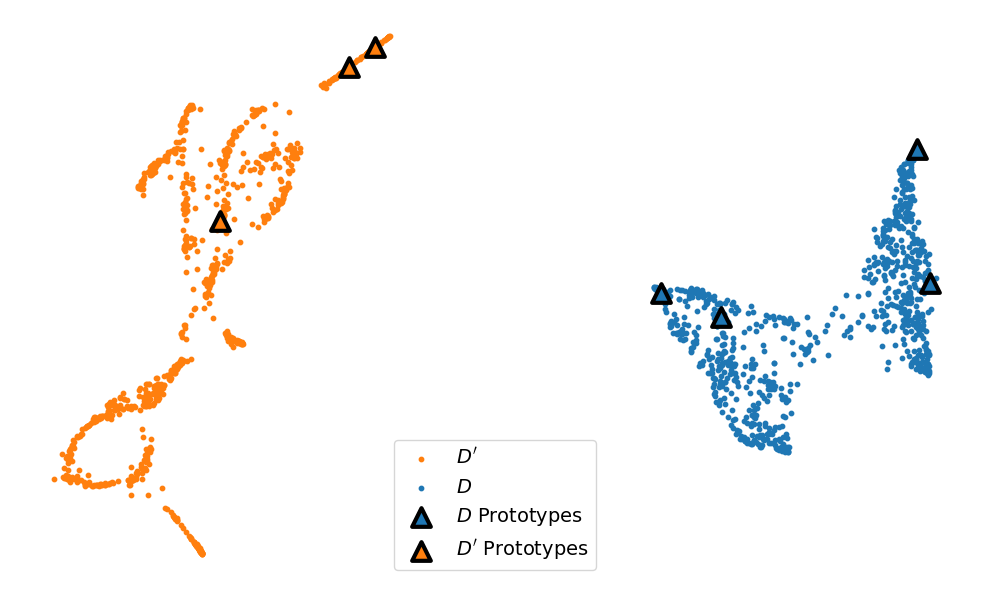}
  \caption{Visualization of the learned latent space for $\mD$ and $\mD^\prime$.}
  \label{fig:audio_pacmap}
\end{minipage}
\end{figure}

\paragraph{Comparing human and machine audio prototypes} It is difficult for a human to tell the difference between the generated and real audio examples, thus we did not know in advance whether there were any differences between them. We show the learned prototypes in Figure~\ref{fig:audioprotos} - we learned four and three unique prototypes respectively for $\mD$ and $\mD^\prime$. These comparisons immediately provide insight into the difference between the human and machine-generated datasets. Specifically, our results indicate that \textit{humans tend to wait before starting to speak, whereas the machine audio starts right away}.  A second observation we can make is the \textit{machine audio waveform has highly periodic patterns where peak-to-peak intervals remain almost constant} throughout the audio piece; we can also see the \textit{machine audio signal amplitude always changes gradually as opposed to human audio}, where there may appear more sudden amplitude changes (e.g., jagged contours of human prototype waveforms). We attribute the the second observation to human nature; human tends to speak with varying speed, loudness, and pitch, whereas synthetic audio always maintains the same pace throughout the whole speech in a more monotonic tone.
The model's insights lead immediately to ways to improve the machine-generated audio to make it more akin to human voice: (1) add in a random wait period before the machine speaks, (2) add frequency and amplitude distortion to the machine audio.
\FloatBarrier
\paragraph{Coverage evaluation} We evaluate the coverage quality of the learned set of prototypes using the AUCC score introduced in Section~\ref{sec:summarizationprotoexp}. The summarization prototype has AUCC of 87.1. The coverage curve is shown in Figure~\ref{fig:aucc_audio}. The learned latent space for $\mD$ and $\mD^\prime$ is shown in Figure~\ref{fig:audio_pacmap} on the right.

\FloatBarrier
\subsubsection{Case Study 2: Medical Image Data - Mammography Patient Population Dataset}
\label{sec:exp_mammo}
% \subfile{subfiles/exp_mammo}
In this case study, we assess the effectiveness of our proposed method in a realistic application, focusing on identifying differences between mammograms from two distinct patient populations. Specifically, we simulate a real-world scenario where users deploy models to analyze mammograms of women with varying tissue density distributions — a challenge commonly encountered when comparing premenopausal and postmenopausal patients or younger and older individuals. Premenopausal or younger patients often exhibit denser breast tissue, whereas postmenopausal or older patients often present with less dense tissue \citep{breatdensityvsage}. This dataset evaluation step is crucial before deploying a clinical breast cancer risk detection model across different patient populations.

\paragraph{Dataset $\mD$ and $\mD^\prime$} We use the publicly available EMBED dataset \citep{embed}. To simulate premenopausal and postmenopausal patient populations, we construct two datasets, $\mD$ and $\mD^\prime$, by randomly sub-sampling from EMBED. Dataset $\mD$ comprises 27,224 mammograms from 8,456 patients with dense breast tissue (density category \textit{three} in EMBED) and 21,675 mammograms from 7,841 patients in density category \textit{two}. Dataset $\mD^\prime$ includes 27,224 mammograms from 2,715 patients with less dense tissue (density category \textit{one}) and 21,675 mammograms from 7,797 patients in density category \textit{two} (medium density). All mammograms were preprocessed to remove clinical markers and aligned such that the breast tissue faces left.

\paragraph{Forming the explanation} For this task, we implemented the ``Prototype-summarization-based explanations'' described in Section~\ref{sec:summarizationprotoexp}. We trained a binary $\mD$ vs $\mD^\prime$ classifier using the VGG19 feature extractor as backbone and learn four prototypes for each dataset. 97798 mammograms were used for training, and 24450 mammograms were used for testing. 

\begin{figure}[H]
    \centering
    \begin{subfigure}{0.49\textwidth}
         \includegraphics[width=\textwidth]{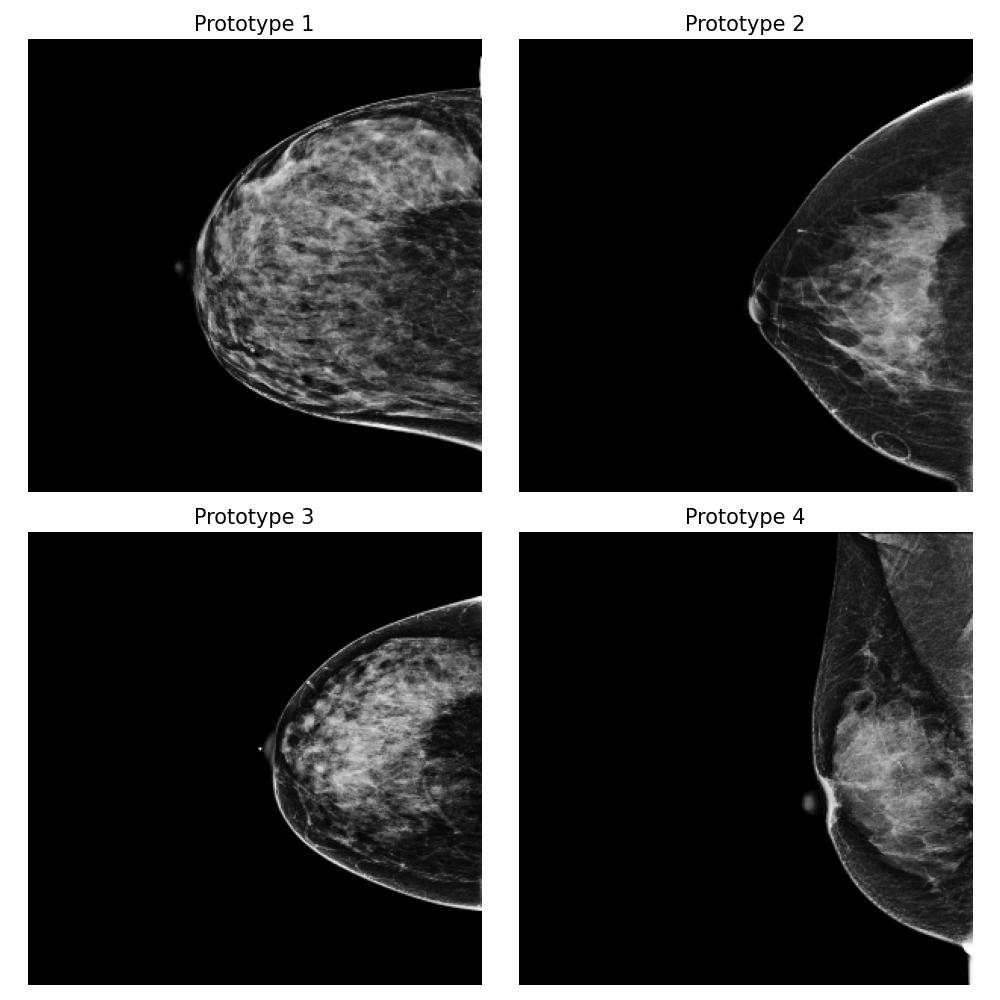}
         \caption{The learned prototypes for dataset $\mD$ with denser tissue.}
         \label{fig:officehome_home_office}
     \end{subfigure}
     \begin{subfigure}{0.49\textwidth}
         \includegraphics[width=\textwidth]{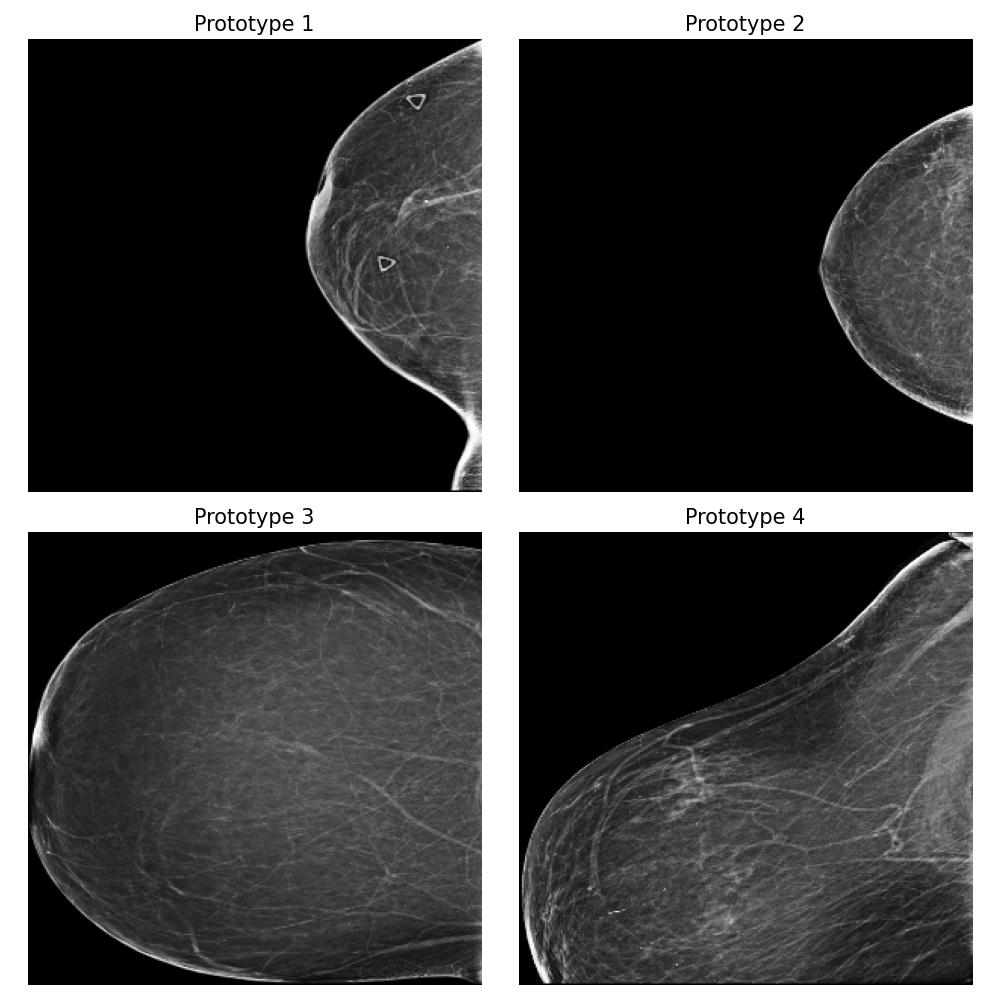}
         \caption{The learned prototypes for dataset $\mD^\prime$ with less dense tissue.}
         \label{fig:officehome_home_protos}
     \end{subfigure}
     \caption{An inspection of these prototypes in $\mD$ and $\mD^\prime$ suggests that our method is able to successfully discover tissue density and size differences between the datasets.}
     \label{fig:mammo_protos}
\end{figure}

\paragraph{Result} By examining the summarization prototypes shown in Figure~\ref{fig:mammo_protos}, we identified tissue density as the primary difference between $\mD$ and $\mD^\prime$. In mammograms, brighter areas correspond to denser tissue. Additionally, we observed that less dense tissue is often associated with larger tissue size. Without our proposed method, human users would need to manually analyze the dataset, which is a labor-intensive and time-consuming task, to reach the same conclusions. 

\begin{figure}
\centering
\captionsetup{justification=raggedright}
\begin{minipage}[b]{.4\textwidth}
  \centering
  \includegraphics[width=1\linewidth]{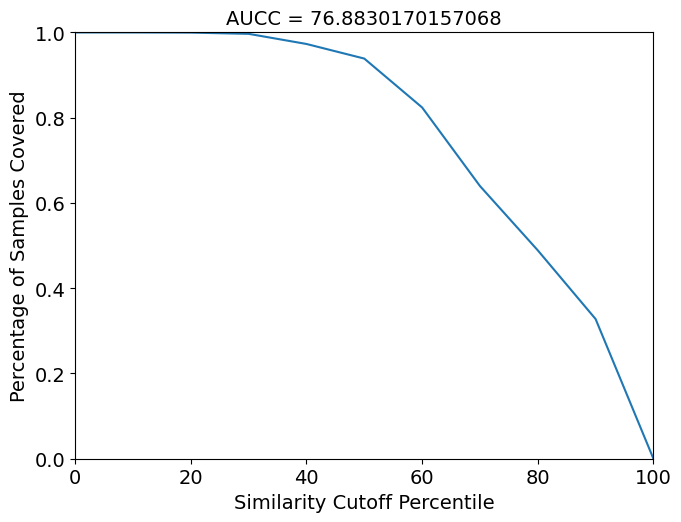}
  \caption{The coverage curve and area under the coverage curve for mammography case study.}
  \label{fig:aucc_mammo}
\end{minipage}%
\begin{minipage}[b]{.6\textwidth}
  \centering
  \includegraphics[width=1\linewidth]{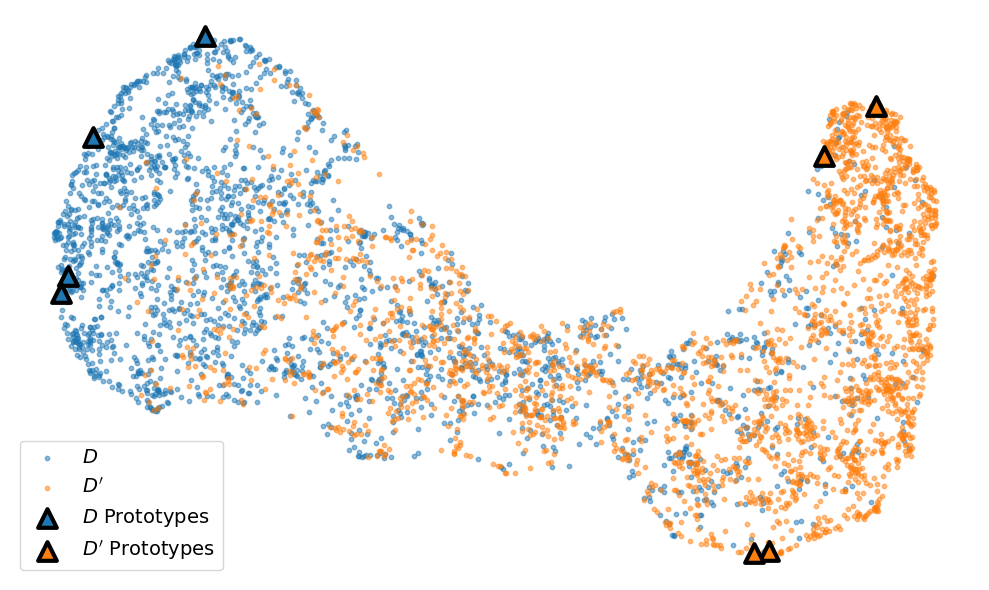}
  \caption{Visualization of the learned latent space for $\mD$ and $\mD^\prime$.}
  \label{fig:mammo_pacmap}
\end{minipage}
\end{figure}

\paragraph{Robustness of the explanation} To examine the robustness of our explanation result, we repeat the explanation algorithm approach on bootstrapped versions of $\mD$ and $\mD^\prime$. Five bootstrapped datasets were constructed by resampling by patients with replacement. As shown in Figure~\ref{fig:mammo_bootstrap_fig} in Appendix Section~\ref{appdx:mammo_bootstrap}, we reach the same conclusion for all the bootstrapped datasets.
\FloatBarrier
\paragraph{Coverage evaluation} We again evaluate the coverage quality of the learned set of prototypes using the AUCC score. The coverage curve is shown in Figure~\ref{fig:aucc_mammo}. We also display the learned latent space for $\mD$ and $\mD^\prime$ in Figure~\ref{fig:mammo_pacmap} and the two datasets and the prototypes are well separated even though they contain overlapping mammograms with density category \textit{two} (i.e. medium density breasts).

\FloatBarrier
\subsubsection{Case Study 3: Image Data - Office-Home Dataset}
\label{sec:exp_officehome}
\paragraph{Dataset $\mD$ and $\mD^\prime$} For this case study, we use an established domain adaptation dataset -- the Office-Home dataset \cite{officehome}. The dataset contains different styles of images from different objects that belong in office or home scenes. In this experiment, dataset $\mD$ contains office objects including ``Table, Shelf, File\_Cabinet, Printer, Calculator, Postit\_Notes, Calendar, Fan, Monitor, Eraser, Folder, Pencil, Mouse,  Push\_Pin, Telephone, Trash\_Can, Paper\_Clip, Ruler, Computer,  Desk\_Lamp, Speaker, Pen, Scissors, Chair, Keyboard, Marker,  Notebook, Clipboards, Laptop, Webcam,'' whereas the $\mD^\prime$ contains home objects including ``ToothBrush, Bed, Oven, Lamp\_Shade, Bottle, Flowers, Radio, Bike, Glasses, Flipflops, Alarm\_Clock, Sneakers, Couch, Mop, Pan, Helmet, Kettle, Mug, Toys, TV, Drill, Spoon, Fork, Refrigerator, Batteries, Candles, Soda, Backpack, Exit\_Sign, Curtains, Hammer, Sink, Bucket, Screwdriver, Knives.'' $\mD$ contains 7101 images, and $\mD^\prime$ contains 8487 images.

\paragraph{Forming the explanation} In this showcase, we again use the ``Prototype-summarization-based explanations'' described in Section~\ref{sec:summarizationprotoexp}. Here, we train a binary $\mD$ vs $\mD^\prime$ classifier with VGG-19 as the feature extractor. 12470 images were used for training, and 3118 images were used for testing. Since we have some rough prior knowledge of the content of the datasets (i.e., they contain several types of objects, but we do not know what objects), we opt for the alternative clustering term defined in Equation~\ref{eq:clst_v2}. 
% We learn 100 prototypes for each dataset, around 3-4 prototypes for each object. Doing so, the algorithm was able to discover 19 out of the 30 office-setting objects and 24 out of 35 home-setting objects. 
We learn 200 prototypes for each dataset, around 6-8 prototypes for each object. We expected the algorithm to be able to capture the diverse art styles and account for the possibility of a large number of concepts/objects in each dataset. 

\begin{figure}[hbt]
    \centering
    \begin{subfigure}{0.49\textwidth}
         \includegraphics[width=\textwidth]{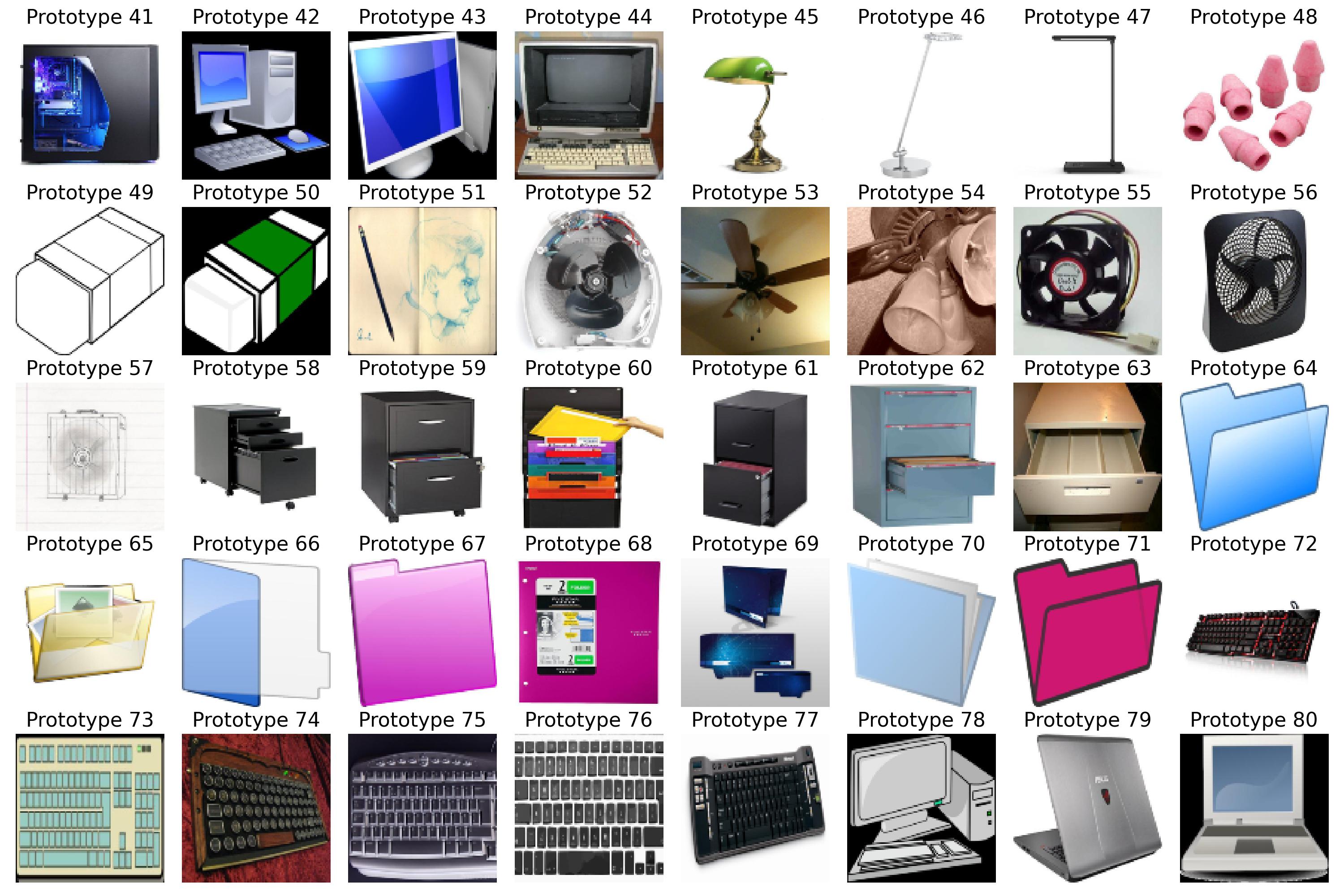}
         \caption{Some of the learned prototypes for dataset $\mD$ with office objects.}
     \end{subfigure}
     \begin{subfigure}{0.49\textwidth}
         \includegraphics[width=\textwidth]{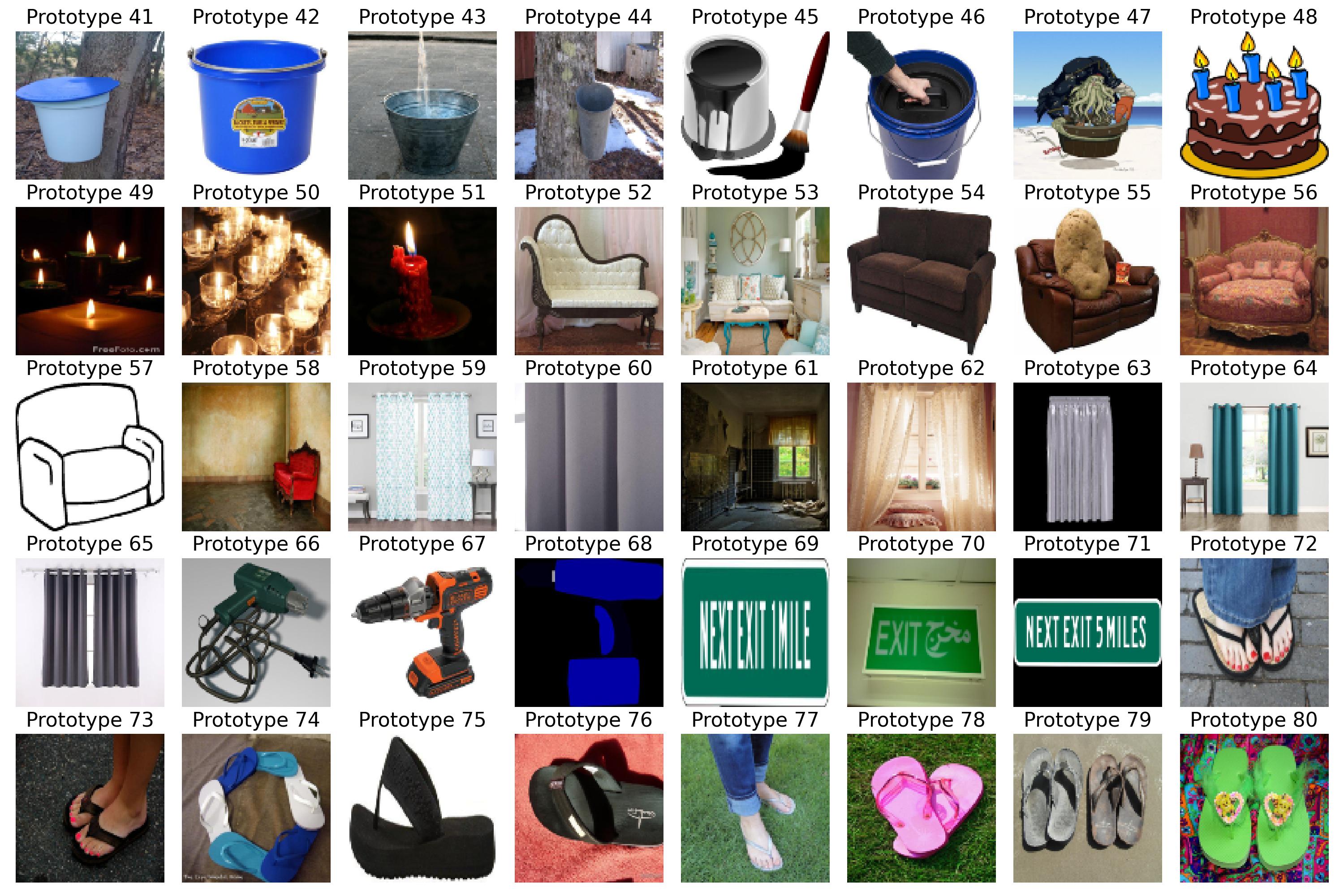}
         \caption{Some of the learned prototypes for dataset $\mD^\prime$ with home objects.}
     \end{subfigure}
     \caption{An inspection of these prototypes in $\mD$ and $\mD^\prime$ suggests that our method is able to successfully discover underlying differences between the datasets. }
     \label{fig:officehome_protos}
\end{figure}

\paragraph{Results} In the retrospective evaluation, the proposed algorithm was able to discover all 30 office-scene objects and 34 out of the total 35 home-scene objects ($\approx{98\%}$ of total objects). The learned prototypes also successfully captured a diverse set of the art styles of the objects in each dataset. We can observe that the datasets $\mD$ and $\mD^\prime$ mainly differ in their object composition; from Figure~\ref{fig:officehome_protos}, we see that $\mD$ contains office objects while $\mD^\prime$ contains home scene objects (note that the full set of learned summarization prototypes are shown in Figure~\ref{fig:officehomeproto_home_office} and Figure~\ref{fig:officehomeproto_home_home} in Appendix Section~\ref{appdx:all_officehome_protos} and all prototype neighbourhoods shown in Figure~\ref{fig:10neigh_officehome}). In addition, we can observe that there are several sets of art styles of the images; these art styles exist in both datasets, thus are not a differentiating factor between the two datasets. Based on the examination of the learned prototypes, our proposed method is able to discover the underlying differences between datasets without introducing incorrect explanations, even when there exists a large number of concepts in the two datasets.

\begin{figure}[hbt]
\centering
\captionsetup{justification=raggedright}
\begin{minipage}[b]{.4\textwidth}
  \centering
  \includegraphics[width=1\linewidth]{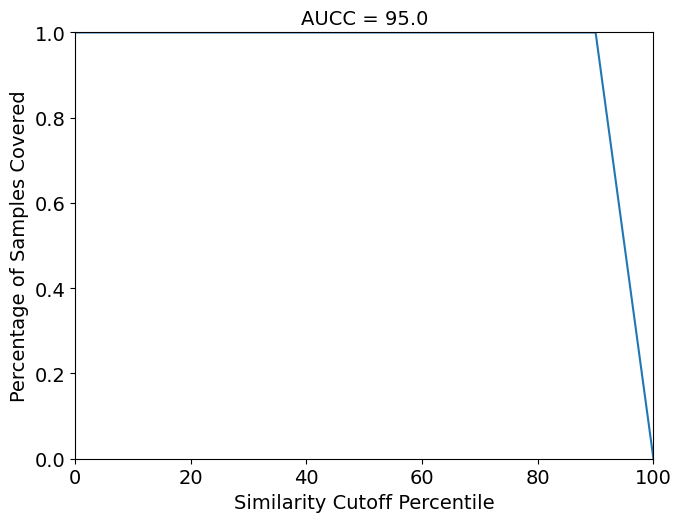}
  \caption{The coverage curve and area under the coverage curve for the office-home dataset case study.}
  \label{fig:aucc_officehome}
\end{minipage}%
\begin{minipage}[b]{.6\textwidth}
  \centering
  \includegraphics[width=1\linewidth]{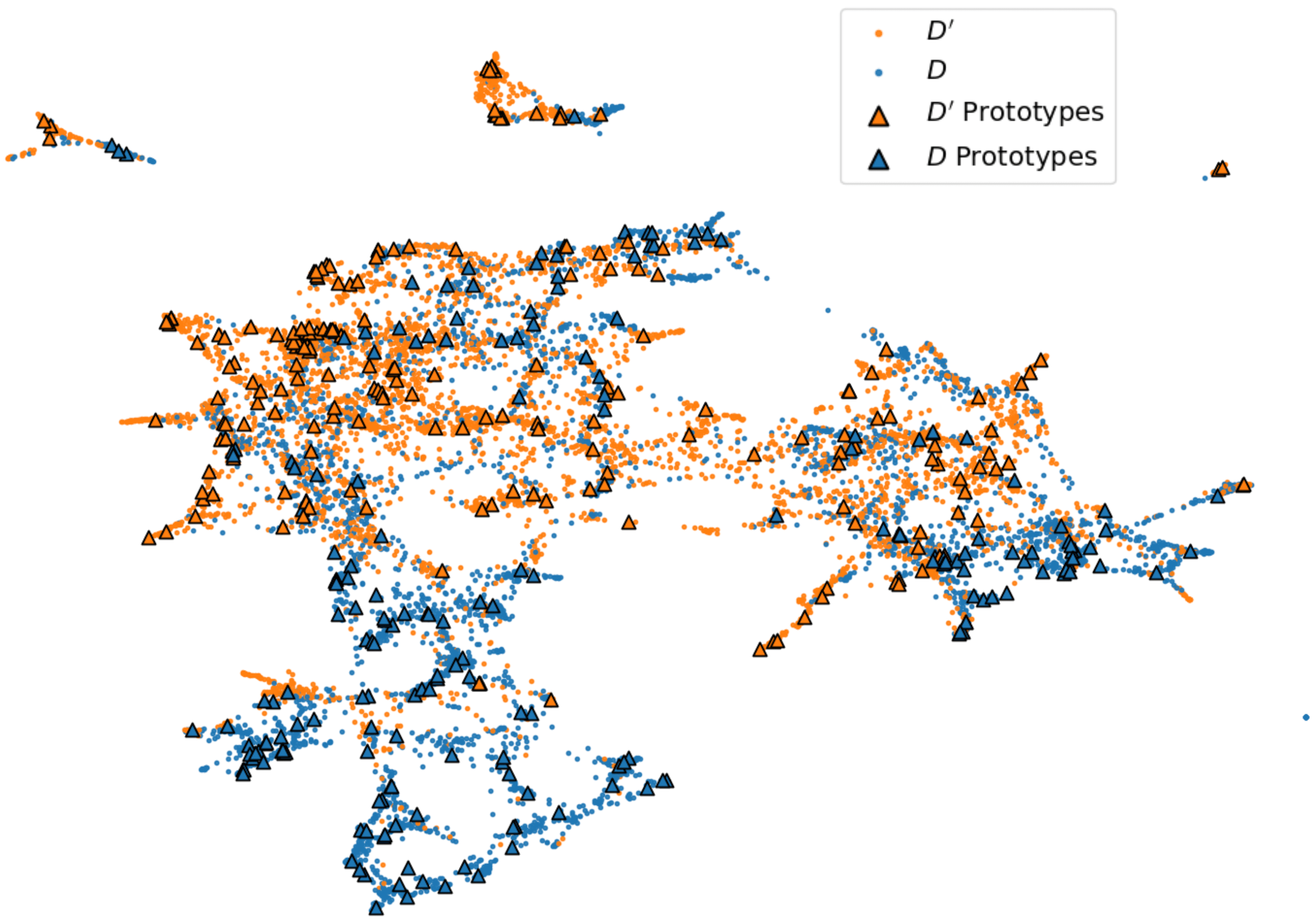}
  \caption{Visualization of the learned latent space for $\mD$ and $\mD^\prime$.}
  \label{fig:officehome_pacmap}
\end{minipage}
\end{figure}

\paragraph{Coverage evaluation} The AUCC score and the coverage curve are shown in Figure~\ref{fig:aucc_officehome}. The model achieved AUCC score of 95, implying that the learned prototypes are highly representative of the latent space. We visualize this learned latent space for $\mD$ and $\mD^\prime$ in Figure~\ref{fig:officehome_pacmap}.

% \subfile{subfiles/exp_officehome}

\FloatBarrier
\subsection{A Brief Note on Comparing Natural Language Datasets}
\label{sec:nlpattributemethod}
Text data are central to modern machine learning, driven by advances in large language models. Unlike other modalities, text data are inherently rich in information, multi-semantic, and context-dependent, producing an intractable number of dimensions for interpretation and explanation. For example, sentiment analysis relies on understanding emotional tone, while machine translation requires preserving contextual meaning across languages, and tasks like named entity recognition demand precise extraction of information.

Our framework proposes extraction and identification of relevant attributes that serve as different dimensions of comparison between two datasets. This formalizes existing work of \citet{elazar2023whats}, where large text corpora were analyzed by extracting summary statistics of different attributes from the data. 
We show an example of how we can compare datasets under this framework in Appendix Section~\ref{sec:llmexp}.

\section{Discussion}
\label{sec:discussion}
In this work, we developed practical approaches to understand shifts in data distributions in an interpretable manner. Our approaches can be seen as constituents of a dataset explanation framework, whose taxonomy is outlined in Figure~\ref{fig:dataset_explanation_tree}. Dataset differences, however, can take on many forms, and what classifies as an interpretable explanation for a shift can be highly context-dependent. We view our work as one of the early efforts towards creating a formal framework for interpretable explanations of dataset differences. To this end, we first provide practical guidance, highlight some caveats associated with our explanations below and propose directions for future work. 
\subsection{Practical Guidance for Using Our Framework}
\label{sec:practical_guide}
\paragraph{Which explanation method should I use?}
Here is our recommendation for how the framework should be used: 
\begin{itemize}
    \item The first thing we generally aim for is exploratory data analysis: visualizing the datasets using a state of the art dimensionality reduction (DR) technique. This will help the user understand high-level structural differences between the datasets. It is important to use a good DR technique so it faithfully captures both global and local structure of the data -- we use PaCMAP currently for this purpose because its performance on global structure preservation tends to be more reliable than other methods \citep{huang2022towards}.
    \item If the datasets are supervised and tabular, the next step would involve understanding which features are important for the underlying predictive task and how this differs between $\mD$ and $\mD^\prime$. Pinpointing specific groups of influential examples that are most responsible for this difference can help inform the user about potential biases present in a dataset (e.g., the presence of highly educated women who were still low income in Section~\ref{sec:fi_exp_adult}).\footnote{For image / signal data, it may be possible to use influential examples if the dataset can be mapped to a fully interpretable latent space (i.e., each dimension of the latent space is interpretable). This is because a crucial aspect of these explanations is the computation of feature importances -- if each element of the feature importance vector is interpretable (which is not the case if the data is composed of pixels), then the explanations provided can become more actionable and useful for stakeholders.}
    \item The next step in the pipeline, which is applicable for tabular, image, and signal datasets (both supervised and unsupervised), is to compare local neighborhoods in the datasets. To this end, using prototypical explanations or prototype summarization explanations will provide valuable insights for image and signal data. For high dimensional tabular data, using partial prototypical explanations is a better option since it provides greater interpretability; we need only explain datasets in terms of features most important in the prototype neighborhood.
\end{itemize}

\paragraph{How many prototypes should I choose?}
Each prototype should represent a neighborhood of samples with a homogeneous concept or feature. For tabular data, one might estimate the number of prototypes in a dataset by first eyeballing the data using its PaCMAP projection and understanding the cluster structure. For unstructured data such as images, audio, or signals, we recommend starting with a higher prototype count and then gradually reducing it by visualizing the latent space with PaCMAP (or other dimension reduction techniques), as well as observing the visualized prototypes to check if there are duplicates. A rough idea of the content of the datasets would be beneficial, but not required, in this process. The prototype explanation method supports any arbitrary number of prototypes for each of $\mD$ and $\mD^\prime$. 

\paragraph{What are some hyperparameter choices for influential example explanations?}
An important design choice for influential example explanations is the size of the set of near-optimal models (i.e., the Rashomon set) used to compute RID \citep{donnelly2023rashomon}. We demonstrate in the appendix (e.g., Figure~\ref{fig:heloc_influential_examples_rset_size}) that once the set of near-optimal models across all bootstrapped Rashomon sets is large enough (i.e., $~\mathcal{O}(10^2)$), the chosen group of influential examples become relatively stable. 

Once the RID importance metric is computed, we now need to decide how many influential examples to consider to explain how $\mD$ differs from $\mD^\prime$. The first step in this endeavour is to consider the distribution of influences across all examples in dataset $\mD$. Technically, all examples with influence $> 0$ could be assessed, as removing them would align feature importances of $\mD$ and $\mD^\prime$ (recall that we train a discriminator to classify whether a feature importance vector is from $\mD$ or $\mD^\prime$). However, \citet{influence_theory} shows that while computed influences are highly correlated with the true difference in discriminator loss (which is corroborated in this work in Figure~\ref{fig:influence_vs_test_loss}, examples with small positive influences may actually reduce discriminator loss in practice, which is the opposite of the intended explanation. Hence, it is advisable to choose a small group of examples from either dataset that exhibit the highest positive influences. In this work, for example, we chose the top $1\%$ of examples for both Adult female and HELOC low-risk datasets.

\paragraph{What is the computational complexity / runtime of our methods?}

The computational time of a prototype learning model scales with the size of the dataset. Additionally, the choice of backbone is a critical factor: while more sophisticated backbones improve feature extraction capabilities, they also lead to longer training and inference times. In contrast, increasing the number of prototypes has a negligible impact on computational time, as the additional parameters are minimal compared to the overall model.

For influential example explanations, the primary computational challenges lie in computing the Rashomon set and deriving the associated feature importances. Fortunately, practical solutions exist to address these bottlenecks. Recent work has demonstrated efficient techniques to approximate or sample from the Rashomon set \citep{treefarms,ciaperoni2024efficientexplorationrashomonset}. Moreover, standard decision tree methods—such as CART, when configured with different initializations (e.g., varying depth budgets and regularization parameters)—can produce a well-performing ensemble of trees. Although this approach does not offer optimality guarantees, it can be effectively used to generate variable importance estimates. A detailed comparison between the method proposed in \citet{donnelly2023rashomon} and this decision tree-based approach remains an interesting avenue for future research.

\subsection{Potential Failure Modes and Avenues for Future Research}
\paragraph{It is unclear how to evaluate dataset-level explanations:} Compared to instance-level explanations, for which there exist several evaluation criteria, there is not yet a well-defined metric to assess the quality of dataset-level explanations. For instance-level explanations, \citet{openxai} has focused on the evaluation of explanations such as SHAP \citep{shap}, LIME \citep{lime}, and IntegratedGrad \citep{integratedgrad} along several axes such as performance, faithfulness, and stability. Other works such as that of \citet{clue} have evaluated counterfactual explanations through human subject experiments \citep{mark_keane}. Analogous tasks for dataset-level explanations are unclear at this point, as there exists no standard evaluative framework. We have some tools to quantify the efficacy of our explanations and their associated tradeoffs (e.g., AUCC score and evaluation methods in Section~\ref{sec:explanation_quality_degradation} in the appendix) -- we consider these to be starting points for future work. %Quantitative measures of comparisons for dataset difference explanations may involve human subject experiments in addition to methodological innovations. 

One potential evaluation would show that taking action based on our explanations would make two datasets more similar, as we showed in Sections~\ref{sec:fi_exp_adult} and \ref{sec:fi_exp_heloc}. \citet{towards_explain} performed such an evaluation for their distribution shift recourse method by showing that their explanation maps Dataset A to a data distribution that more closely resembles that of Dataset B. For our framework, we show a recourse-based evaluation for influential example explanations in the appendix of this paper, however, analogous recourse is not yet clear for other forms of explanations, e.g., prototype-based explanations. We could envision generating/collecting data around prototypes that are not represented in one of the two datasets, for example. Other evaluations could include the creation of benchmark datasets with known ground truth differences that future dataset explanations need to uncover.
It is also not yet clear which explanation is most suitable for a given dataset or whether this is subjective and depends on the human. Figure~\ref{fig:dataset_explanation_tree} in the paper provides a starting point for choosing an explanation method, but this is based more on the modality of the dataset rather than any inherent property of the data itself (e.g., latent structure, dimensionality).

\paragraph{It is hard to determine dimensions along which to compare complex natural language datasets without knowing the task:}  Recent work \citep{elazar2023whats} devised a tool called WIMBD which can analyze text corpora to understand the content of large-scale datasets. This provides information on summary statistics of the dataset, including token distributions, personally identifiable information, and potential biases present within the text. An explanation framework focused on comparing natural language datasets could utilize this tool and compare high level summaries of the datasets along these axes. Our work focuses on a brief formalism of some ideas in their work, suggesting that mining attributes of language corpora and generating summary statistics along those attributes (e.g., sentiment, topic) can yield valuable comparative insights between corpora. However, it is not yet clear how (or if) we can design a generic framework for more nuanced comparison of text corpora, e.g., comparing writing styles, tone, and topic compositions, or mine appropriate attributes along which to compare corpora. This is an open direction for future work.

\paragraph{There is no guarantee of actionability:} The actionability of insights and explanations generated from our analysis may vary based on the task. A generated insight can be very actionable if a difference between datasets can be directly fixed by tuning a parameter in the generative algorithm or changing the data sampling and collection strategy. While our explanation is most useful for understanding the differences, it might be harder to design algorithms to actually mitigate those differences for the datasets in question.

\paragraph{For high dimensional data with non-interpretable features, the quality of the explanation  depends on the quality of the latent space:} For prototype-based explanations derived from training discriminators (e.g., in Sections~\ref{sec:exp_officehome} and \ref{sec:exp_mammo}), the prototype distances are computed in the latent space. The underlying assumption is that distances in the latent space are meaningful. That is, if two examples are close in latent space, it is because these examples share some similar characteristics that are semantically meaningful and interpretable to humans. In our experiments, the learned summarization prototypes worked well to aid in explaining the differences between two datasets. However, they are not guaranteed to be absolutely faithful. Users should examine the detailed visualization of the neighbourhood before taking further action. For example, there are inconsistent and ambiguous groupings shown in Figure~\ref{fig:10neigh_officehome}, where the neighboring samples contain different objects but the same art styles, leaving the actual difference up to interpretation. We also would like to note that even though PaCMAP is optimized to maintain both global and local structure, there is no absolute guarantee of the dimension reduction algorithm's ability to maintain faithful structure in extremely high-dimensional cases. See \citet{huang2022towards} for a review on trustworthiness of dimension reduction methods.

\paragraph{Guarantees of completeness:} 
Prototype methods capture a \textit{sufficient} set of differences between the two datasets but do not necessarily capture \textit{all} differences between the datasets. In other words, there may be other ways the datasets differ that are not captured by a single prototype model. This may not be problematic in some cases, since we may only need to see the main differences. In our current framework, users could iteratively run our algorithm in an explain-then-mitigate loop to find out more differences. Although fully delineating the differences between datasets may not be immediately necessary for practical purposes, it presents a promising avenue for future research.\\
\paragraph{Influential example explanations can only be used if local feature importance can be computed:} Some methods such as permutation importance and Gini-impurity provide global feature importances (GiFIMs) directly, with no information given on the importance of a feature for the prediction of a particular instance (LiFIM). Because our explanation technique relies on a) computing GiFIMs as an aggregate of LiFIMs, and b) computing influences from LiFIMs, methods that skip this step will not be compatible with our technique. Future work will seek to implement this explanation in scenarios where only a global feature importance metric is provided. 

\paragraph{Influential example explanations may require access to a Rashomon set:}
In our version of influential example explanations, we were considering features that were important across the entire set of near optimal models (i.e., the Rashomon set). We also evaluated decision trees as the model class, however, other the method can easily be adaptable to other model classes. The only caveat is that it can be expensive to compute this set, especially for datasets with many features and with more complex model classes. In these situations, a useful proxy metric can be to examine an easily computable subset of representative models rather than the entire Rashomon set (e.g., collecting all empirical risk minimizers trained on bootstrapped versions of the data).

\paragraph{AUCC score:}
The AUCC score is a useful metric to evaluate the coverage of the prototypes. It is worth noting that the score could be trivially maximized when the model learns a trivial solution that aligns every single sample to prototypes regardless of content (even though we offer optional learning targets that regularize against this behavior) or when the user chooses an unreasonably large number of prototypes as the hyperparameter. The AUCC score should be interpreted along with the visualization of the prototype neighborhood and, optionally, the visualization of the latent space.

\section{Conclusion}
In conclusion, we developed an explainable AI paradigm for explaining the differences between any two datasets in an interpretable manner. The suite of approaches proposed in this work provides end users with insights and actionable clues to understand and mitigate the differences.
With case studies and experiments that cover a variety of data modalities and common machine learning tasks, we demonstrate the comprehensiveness and adaptability of our methods. Our framework is most useful for detecting biases in synthetic data, understanding erroneous examples in specific regions of the input space, and exploring the impact of discrepancies on model performance. Our study could potentially improve machine learning algorithm robustness in the real world by allowing researchers to examine changing factors, enabling future studies to improve generative algorithms, and other data science / exploratory data analysis applications. We envision future work focusing more on evaluation and the HCI aspect of this research direction, aiming to understand how practitioners can benefit most from our framework.

\section{Ethics note}
This work could be used to help improve deepfake techniques and adverse / malicious machine signal, image, and tabular data generation in various domains.

\section{Acknowledgement}
We thank Jon Donnelly and Srikar Katta for helpful discussions and guidance on using Rashomon variable importance. We acknowledge support from NIH 1R01HL166233-01, NIH/NIBIB 5P41-EB028744, and NSF IIS-2130250.

\FloatBarrier
\bibliography{main}
\newpage
\appendix
\section{Cardiac Signal Simulation Parameters}
\label{sec:PPG_simulation_parameters}

The synthetic PPG (Cardiac) signals were generated using the following parameters using neurokit2 \cite{neurokit}. Parameter values were chosen either as a fixed value or randomly chosen from the listed value range. 

\begin{table}[hbt!]
\centering
\begin{tabular}{|c|c|} \hline 

\textbf{Parameters}  & \textbf{Value Range} \\ \hline  
 sampling\_rate& 80\\ \hline  
heart\_rate& 81 - 100             \\ \hline  
frequency\_modulation& 5 - 21\\ \hline  
ibi\_randomness& 5 - 21\\ \hline  
drift& 0 - 1\\ \hline  
powerline\_amplitude& 0 - 1\\ \hline  
burst\_amplitude& 0 - 1\\ \hline 
 burst\_number&0 - 9\\ \hline 
 noise\_shape&laplace\\ \hline 
 artifacts\_amplitude&1\\ \hline 
 artifacts\_frequency&5 - 9\\ \hline 
 artifacts\_number&15 - 31\\ \hline 
 linear\_drift&True / False\\ \hline
\end{tabular}
\end{table}

\section{Evaluation of Prototypical Explanations}
\label{sec:explanation_quality_degradation}
We perform several evaluations of our explanations in order to characterise their quality under different scenarios. As is typical with function approximation, approximation faithfulness and completeness is sacrificed if we reduce the complexity of the explanations. We define explanation quality in terms of its \textit{faithfulness} to the underlying difference between two datasets:
\begin{definition}[Faithfulness]
    The faithfulness of a dataset difference explanation is the extent to which the explanation captures the actual difference between two datasets. The exact measure of faithfulness depends on the type of dataset explanation being generated.
\end{definition}
\subsection{Prototype-Based Explanations for NSPD and NSDD}
\label{sec:prototypical_explanations_appendix_analysis}
The main assumption made for prototype-based explanations of this type is that the prototype is representative of the neighbourhood. This is generally true if the neighbourhood is small, but the quality of the explanation will degrade as the neighbourhood grows, because it will contain more varied data.
On the other hand, the explanations are more general if they cover a larger neighbourhood. 
This is analogous to an argument made in selecting the number of clusters for $k$-means, except that prototypes are a generalisation of cluster-centres and can be chosen to prioritise certain neighbourhoods. Based on the above analysis, we have two conflicting desiderata for an ideal prototypical explanation:
\begin{itemize}
    \item Each prototype must faithfully represent its neighbourhood, which means the neighbourhood should be sufficiently small. 
    \item The explanations must be general, which means the neighbourhood should be sufficiently large. (This leads to a smaller overall number of prototypes.)
\end{itemize}
We illustrate the tradeoffs associated with these desiderata for explaining the HELOC and Adult datasets in the figures below. We assume a similar setup as in Sections~\ref{sec:tab_heloc_prototype} and \ref{sec:tab_adult_prototype}. 
\begin{figure}[H]
    \centering
    \includegraphics[width=0.5\textwidth]{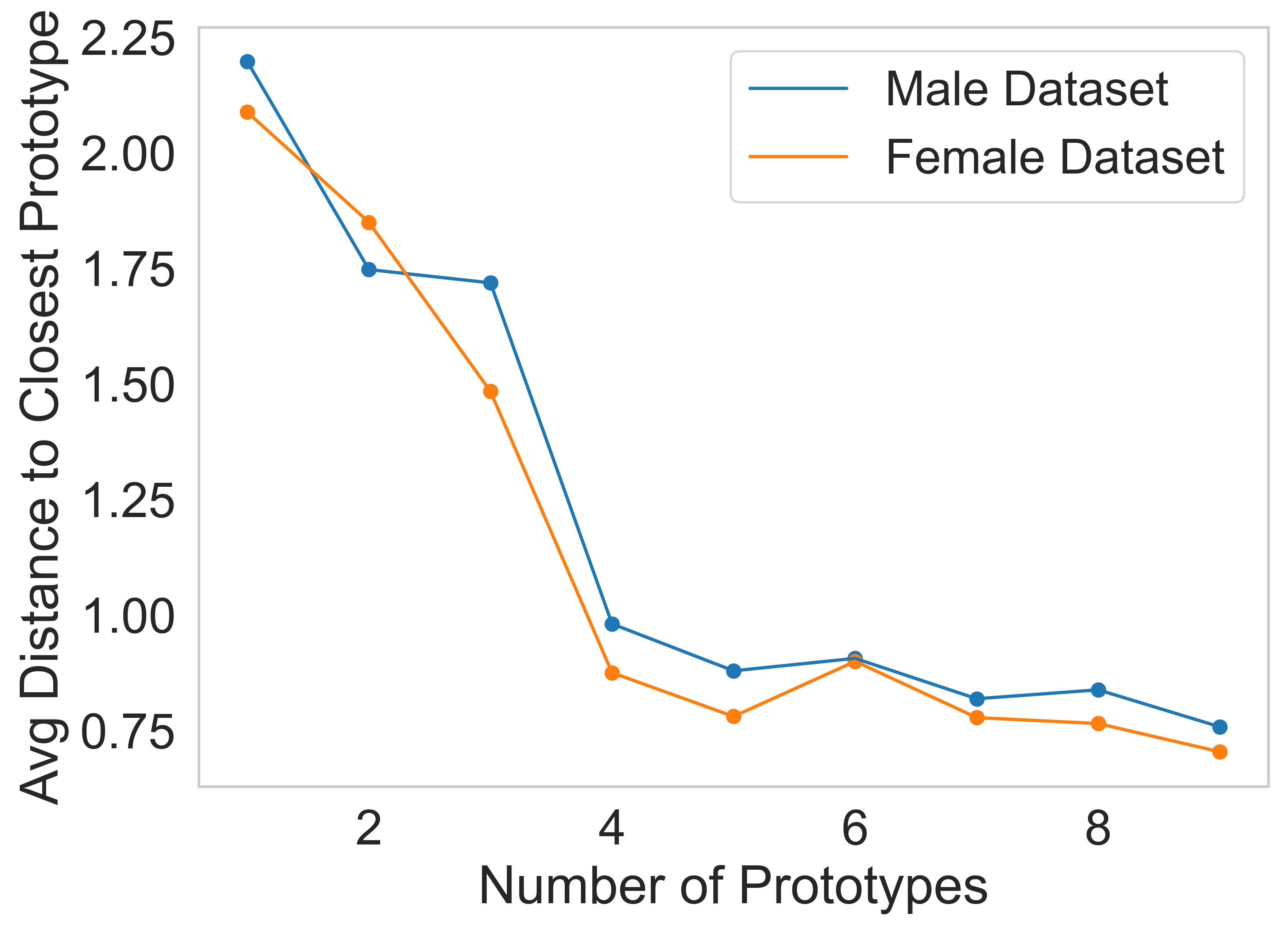}
    \caption{Illustrating the faithfulness-interpretability tradeoff for prototype-based explanations on the Adult male and female datasets. Here, the complexity -- and therefore interpretability -- of the explanation is determined by the number of prototypes. The representativeness of a prototype in its neighbourhood -- as measured by the average distance of points to the prototype -- determines faithfulness.}
\end{figure}
\subsection{Choosing Partial Prototypes}
\label{sec:partial_prototypes_analysis}
For tabular datasets such as HELOC, we want to choose a subset of features for each prototype that best represent the neighbourhood. In Section~\ref{sec:fi_exp_heloc}, we selected features according to the following desiderata: 
\begin{itemize}
    \item The chosen features should not vary much in the neighbourhood of the prototype.
    \item The chosen features should be important for both the prototype and its neighbourhood. Here, importance is measured relative to the underlying task at hand.
\end{itemize}
Because the distance metric used in computing the NSPD and NSDD involved all of the features, when we then restrict ourselves to using a subset of the features, the observations within the neighbourhood may differ from each other on this subset.

To capture this degradation in the distance metric, we use two measures of faithfulness: 
\begin{definition}[\textbf{Random Triplet Accuracy }(from \citet{pacmap})]
    Choose any $3$ points from a dataset over $N = 1000$ trials. The random triplet accuracy then measures the proportion of trials where the triplets maintain their relative order in both low and high-dimensional space feature spaces.
\end{definition}
\begin{definition}[\textbf{Global Permutation Accuracy}]
    The global permutation accuracy captures the distance between permutations. Given two arrays containing separate distance measurements, first argsort both arrays. The global permutation accuracy is the proportion of positions where the rankings of elements in both arrays match (e.g., if element $1$ has the $36^{th}$ largest distance in both arrays, then this is considered one match).
\end{definition}
Let $X_p \in \mathcal{X}$ be a prototype of interest in dataset $\mD$. Let $V$ be all the points in its neighbourhood. Let $\{\alpha_{(1)},...\alpha_{(K)}\}$ represent the chosen subset of $K$ features, with corresponding partial prototype $X_p[\alpha_{(1)},...\alpha_{(K)}]$. We now illustrate this degradation in explanation quality as a function of $K$.
\begin{enumerate}
    \item We first order the points in $V$ according to distance from the prototype $X_p$. Let $\sigma(V)$ represent this ordering. 
    \item We then select $K$ random features from the feature set and compute the partial prototype $X_p[\alpha_{(1)},...\alpha_{(K)}]$. 
    \item We now order points in $V$ according to distance to the partial prototype. Let $\sigma_K(V)$ represent this new ordering of points.
    \item We then compute the Random Triplet and Global Permutation Accuracies of the chosen $K$ features. 
\end{enumerate}
\begin{figure}[H]
    \centering
    \includegraphics[width=0.85\textwidth]{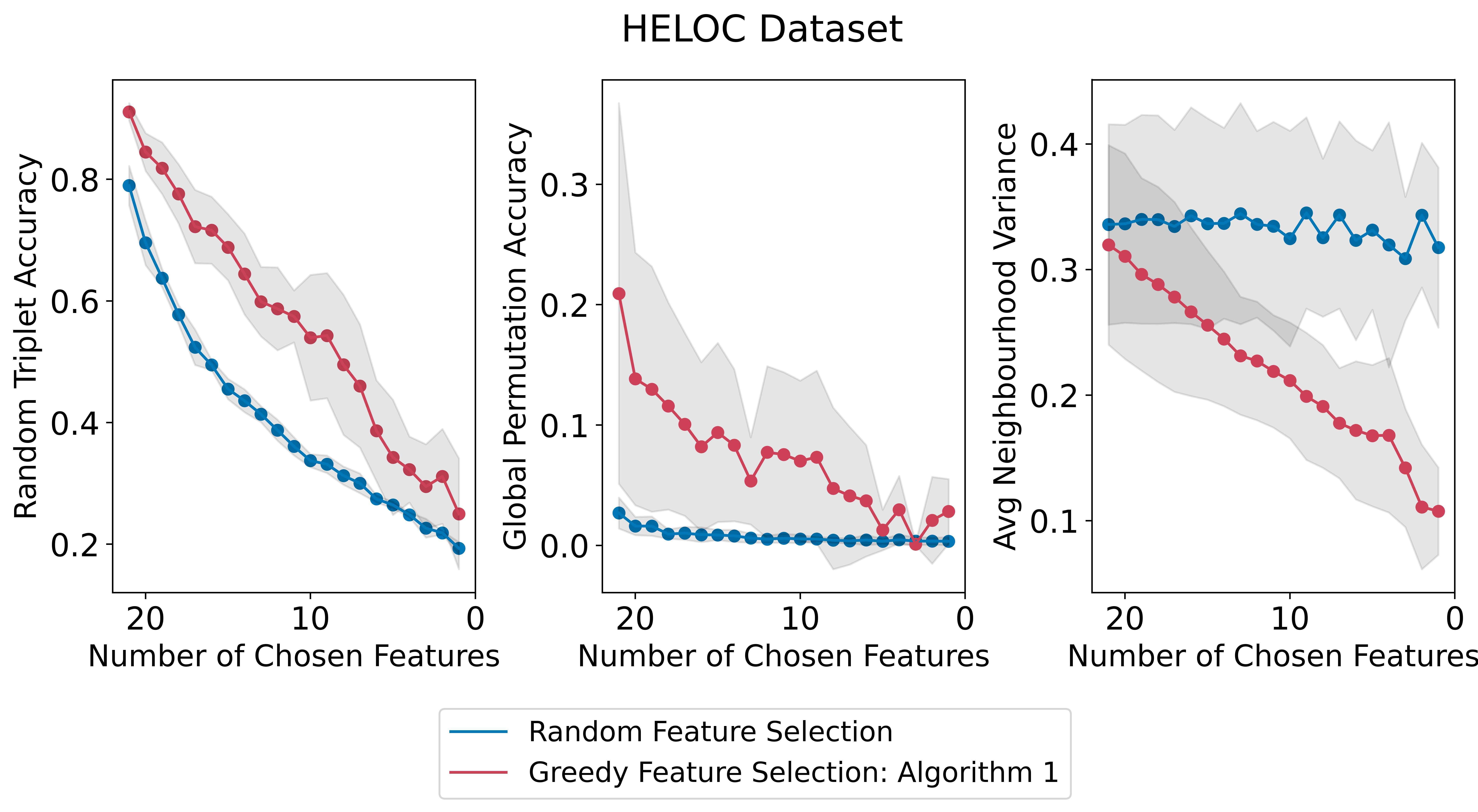}
    \caption{Illustrating the faithfulness-interpretability tradeoff (averaged across all prototypes) for partial prototype-based explanations on the HELOC risky and non-risky datasets in Section~\ref{sec:tab_heloc_prototype}. The greedy feature selection procedure in Algorithm~\ref{alg:partial_prototypes} is choosing features which are value-stable in the prototype neighbourhood (i.e., they vary the least) by setting $c_1$ and $c_2$ to $0$. As the number of features chosen for a partial prototype reduces, there is an increasing degradation in local and global structure preservation. This is measured using two related interpretations of faithfulness -- Random Triplet Accuracy (left) and Global Permutation Accuracy (middle). We also note that the variance around the neighbourhood (right figure) is lower than with random feature selection, implying that the partial prototype generated using our method is more faithful to the neighbourhood. }
    \label{fig:faithfulness_vs_K}
\end{figure}
\subsection{Partial Prototype Feature Selection: Sensitivity Analysis}

\begin{figure}[H]
    \centering
    \includegraphics[width=0.8\textwidth]{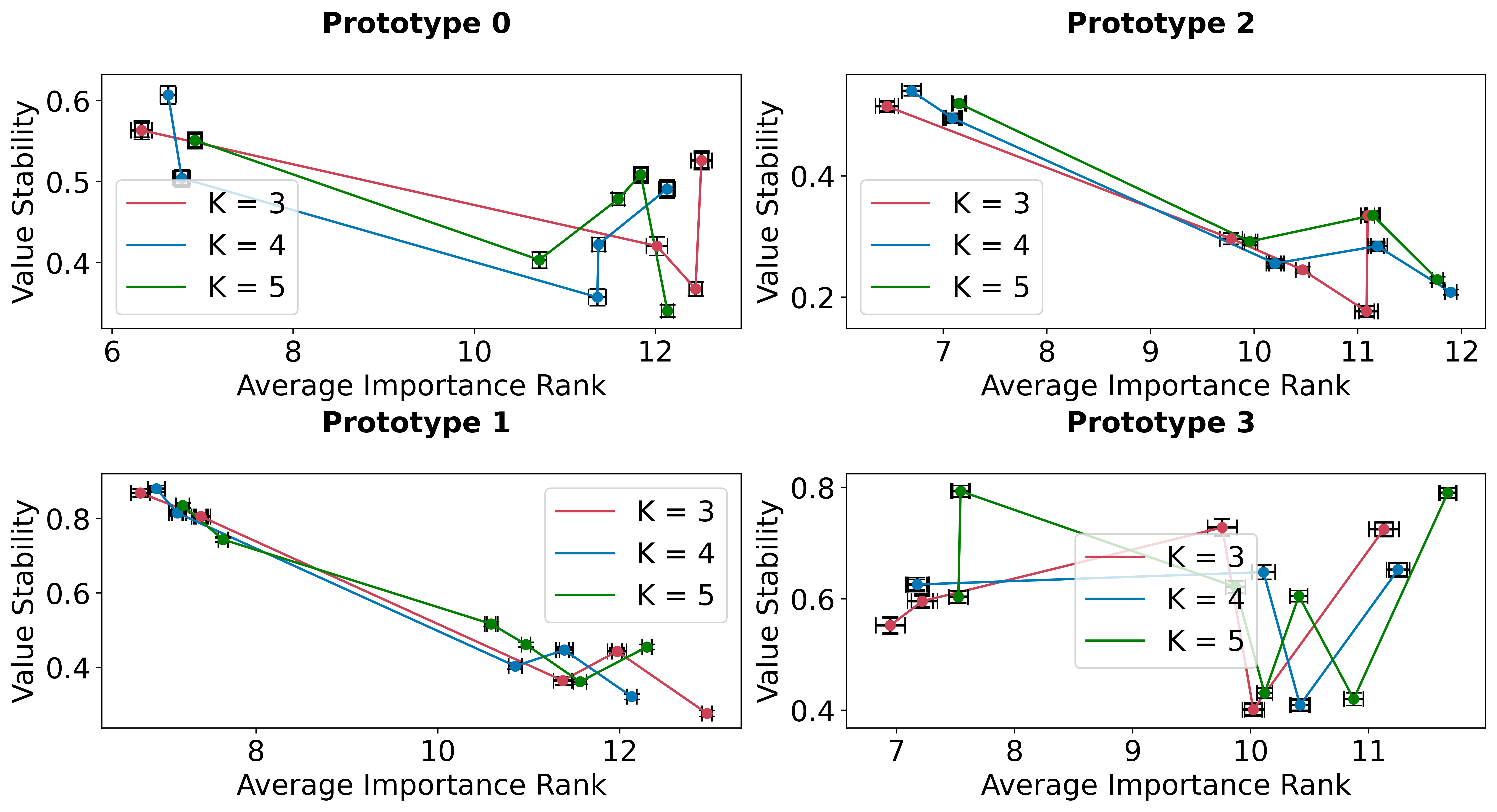}
    \includegraphics[width=0.8\textwidth]{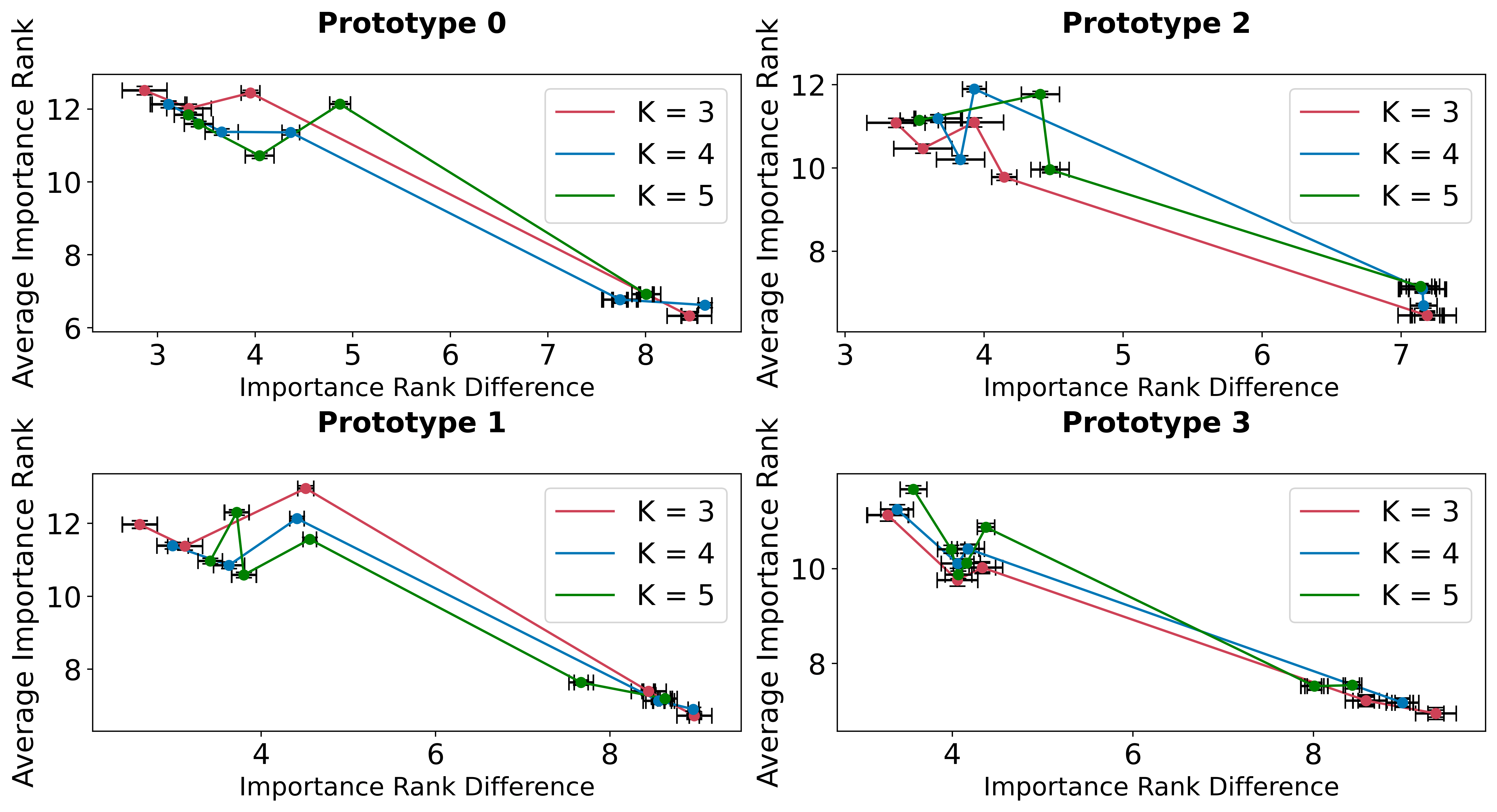}
    \caption{Relationship Between Rank Difference and Value Stability (top) and Rank Difference and Absolute Rank (bottom) of chosen partial prototype features (HELOC dataset). These curves were obtained by sampling random values of $c_1$, $c_2$, and $c_3$ in a logarithmically spaced interval $[10^{-2}, 10]$. For a given $(c_1,c_2,c_3)$ tuple (which represents a single point on both curves), we then found the $K \in \{3,4,5\}$ most relevant features of the prototype and compute the rank difference, value stability, and average rank of these features in the $\delta$ neighbourhood. We chose $\delta$ as the $10^{th}$ percentile distance of all points from the prototype.}
    \label{fig:heloc_fsr_tradeoff_study}
\end{figure}
From Figure~\ref{fig:heloc_fsr_tradeoff_study}, we can see the following: 
\begin{itemize}
    \item There exists a tradeoff between importance rank difference and average importance rank -- this is analogous to a bias-variance tradeoff. In particular, one can choose a feature that is on average more important in a prototype neighbourhood, but this feature will have higher variation in importance rank. That is, the absolute rank difference between the feature's importance for the prototype and the importance of its neighbourhood point will be higher. 
    \item There is also a tradeoff between average importance rank and value stability. This means that one can choose a feature that is on average more important to the underlying task in a prototype neighbourhood, but this feature is likely to take on a larger spread of values.
\end{itemize}
Navigating these tradeoffs according to user requirements is an essential aspect of choosing the correct partial prototype features that are truly representative of the neighbourhood.

\subsection{All learned summarization prototypes for the Office-Home Experiment}
\label{appdx:all_officehome_protos}
Here, we show all the learned prototypes for summarizing the differences between dataset $\mD$ and $\mD^\prime$ in Section~\ref{sec:exp_officehome}.

For dataset $\mD$ (which contains office objects):
\begin{figure}[H]
    \centering
    \begin{subfigure}{0.49\textwidth}
         \includegraphics[width=\textwidth]{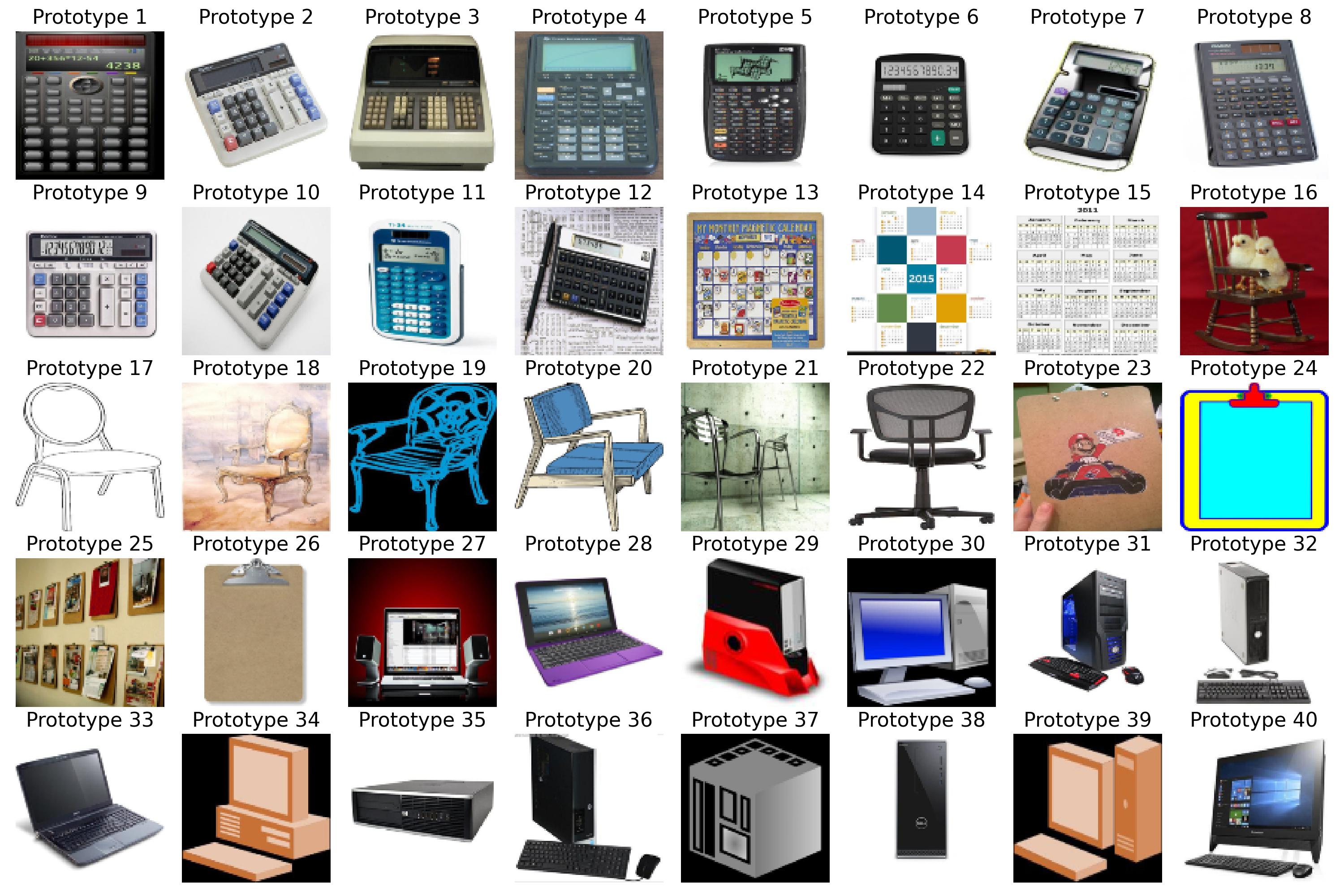}
         
     \end{subfigure}
    \begin{subfigure}{0.49\textwidth}
         \includegraphics[width=\textwidth]{figures/officehome_nppc200/collage_0_1.jpg}
         
     \end{subfigure}
     \begin{subfigure}{0.49\textwidth}
         \includegraphics[width=\textwidth]{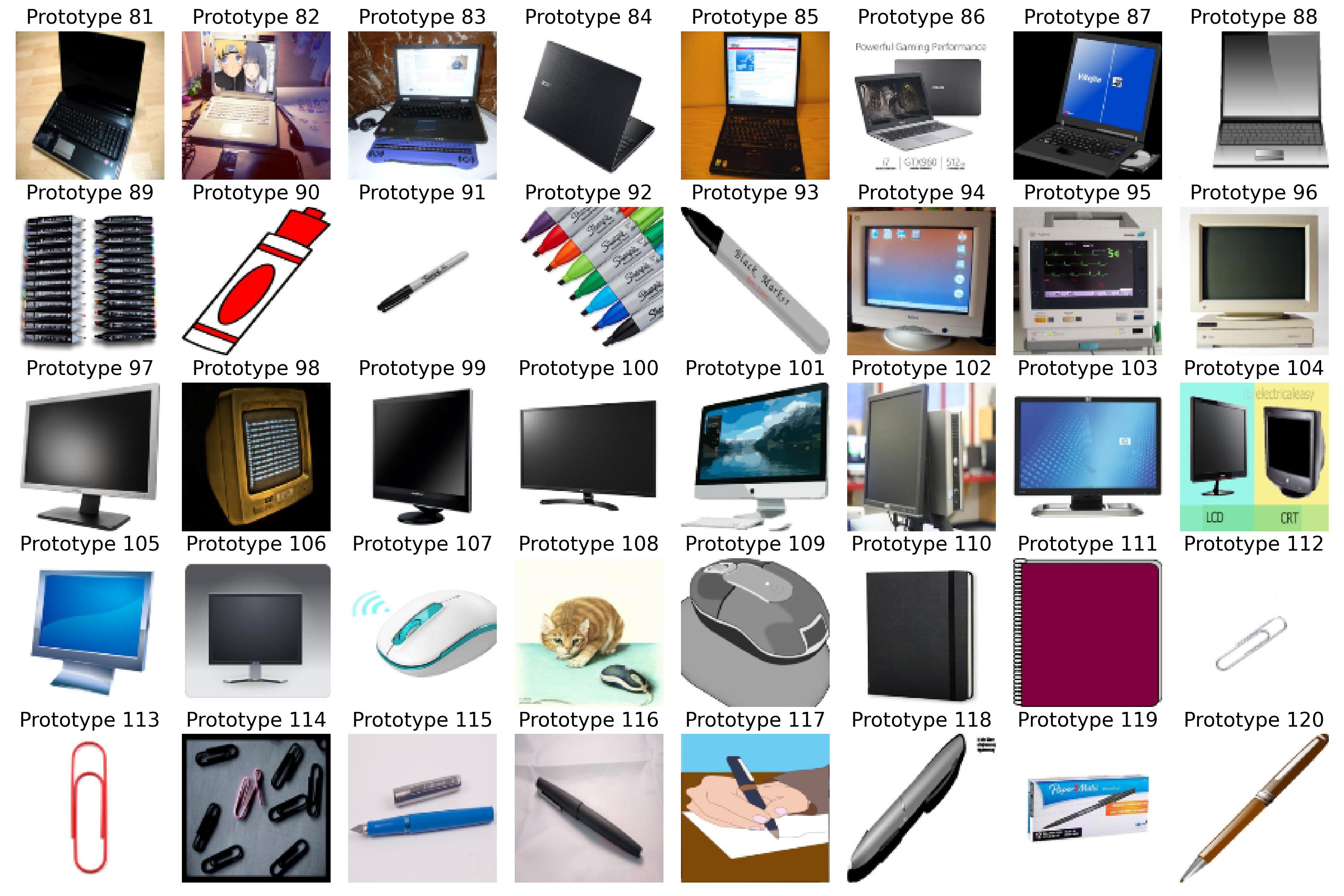}
     \end{subfigure}
     \begin{subfigure}{0.49\textwidth}
         \includegraphics[width=\textwidth]{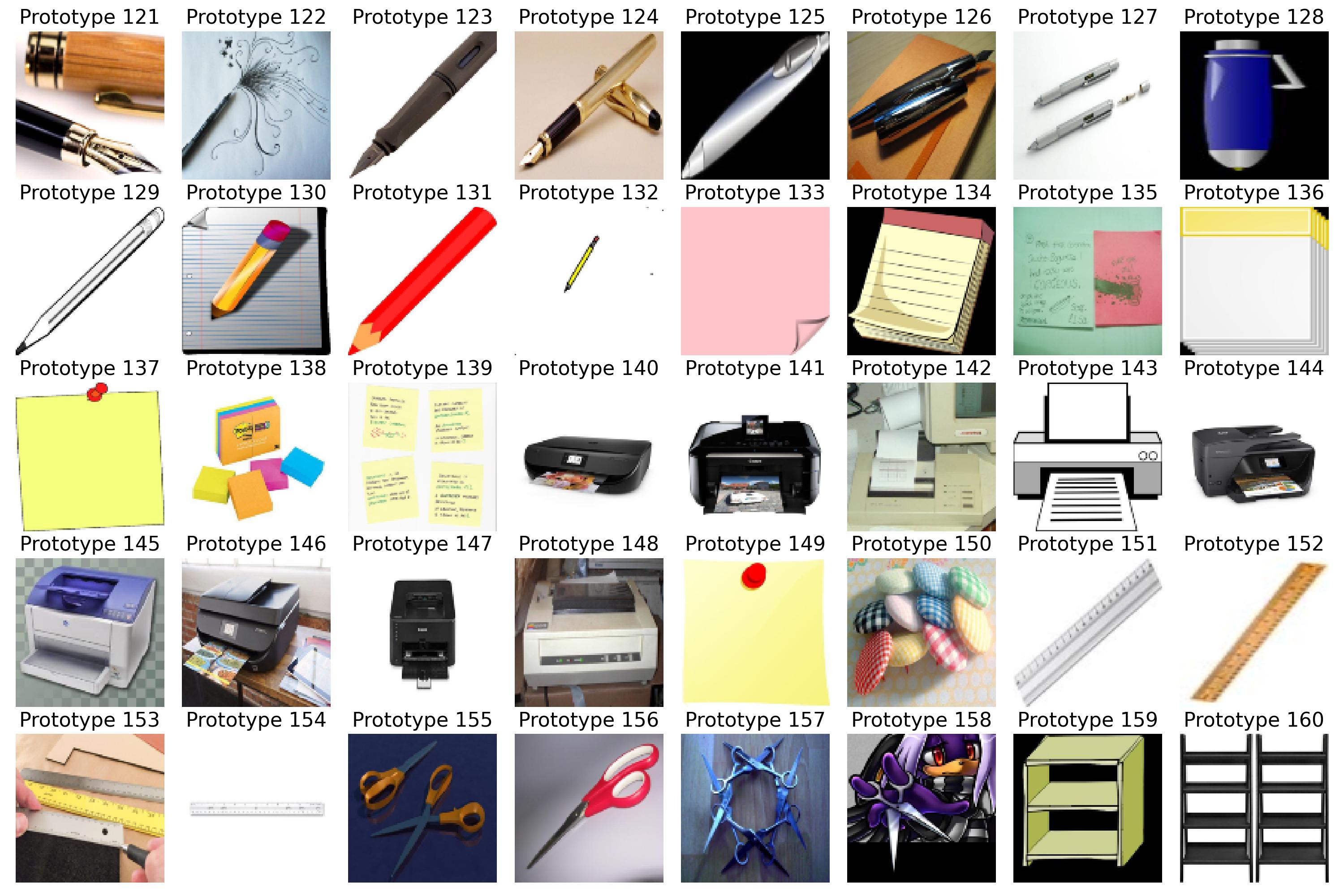}
     \end{subfigure}
     \begin{subfigure}{0.49\textwidth}
         \includegraphics[width=\textwidth]{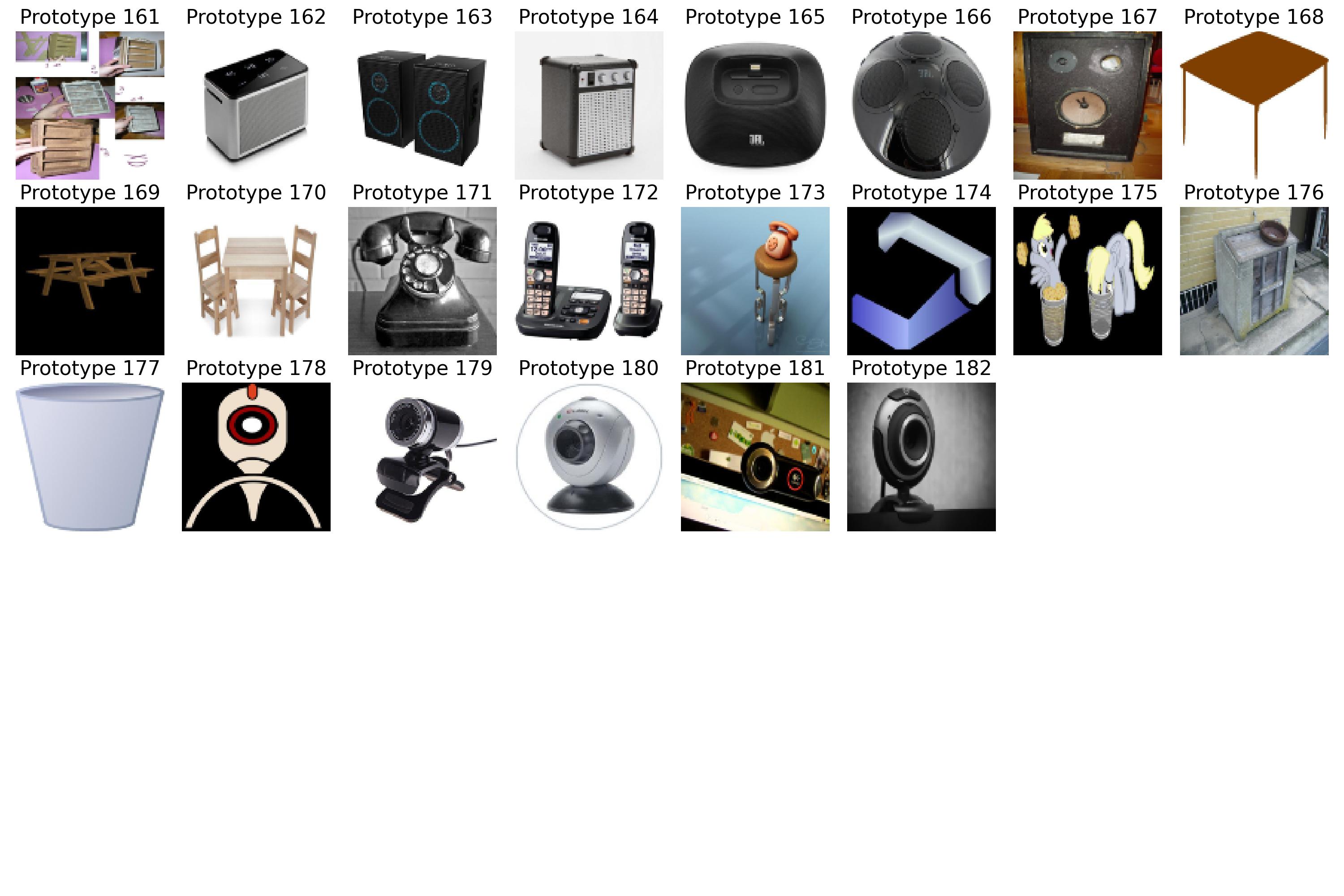}
     \end{subfigure}
     \caption{All learned prototypes for dataset $\mD$ with office objects.}
    \label{fig:officehomeproto_home_office}
\end{figure}

For dataset $\mD^\prime$ (which contains home objects):
\begin{figure}[H]
    \centering
    \begin{subfigure}{0.49\textwidth}
         \includegraphics[width=\textwidth]{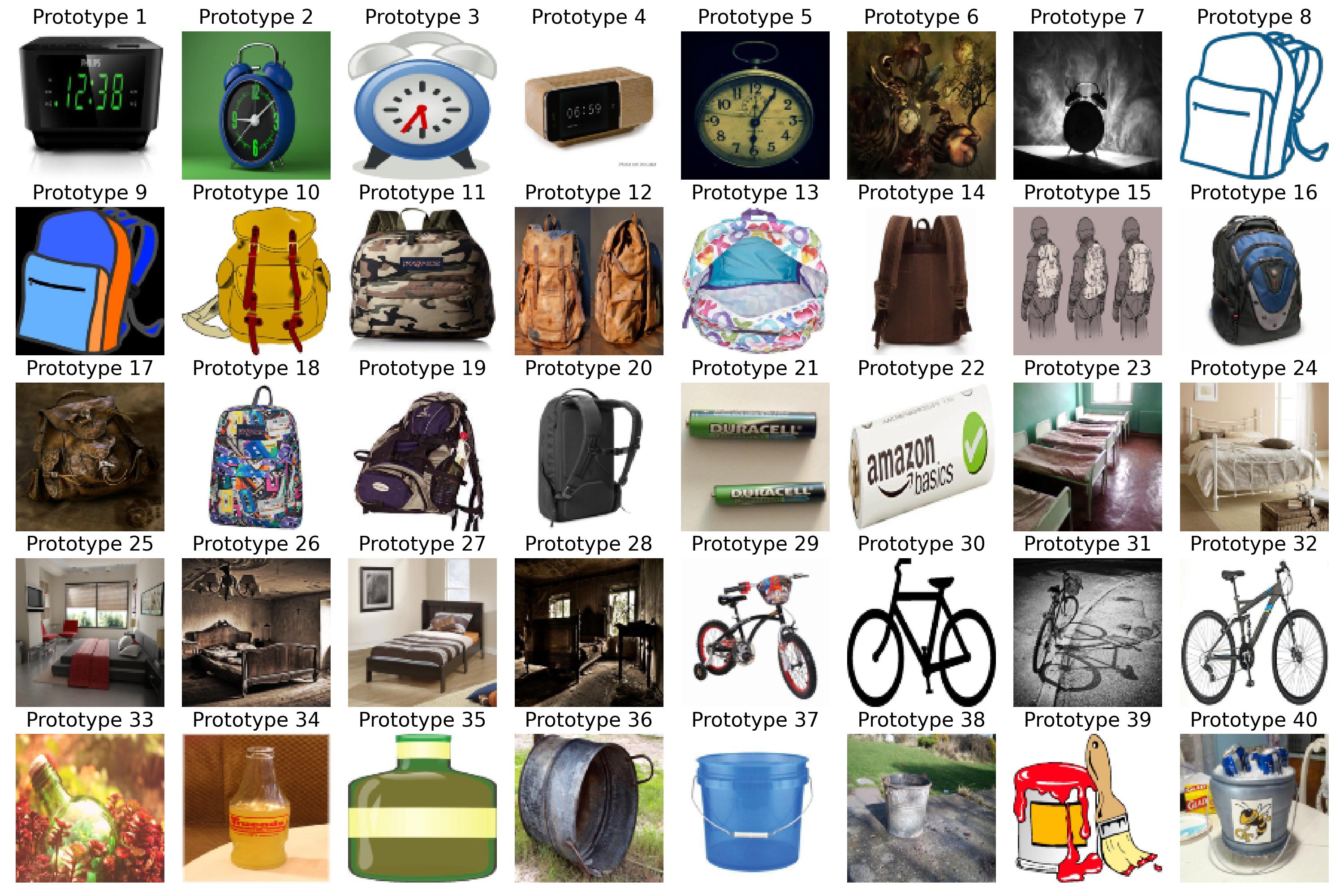}
     \end{subfigure}
    \begin{subfigure}{0.49\textwidth}
         \includegraphics[width=\textwidth]{figures/officehome_nppc200/collage_1_1.jpg}
     \end{subfigure}
     \begin{subfigure}{0.49\textwidth}
         \includegraphics[width=\textwidth]{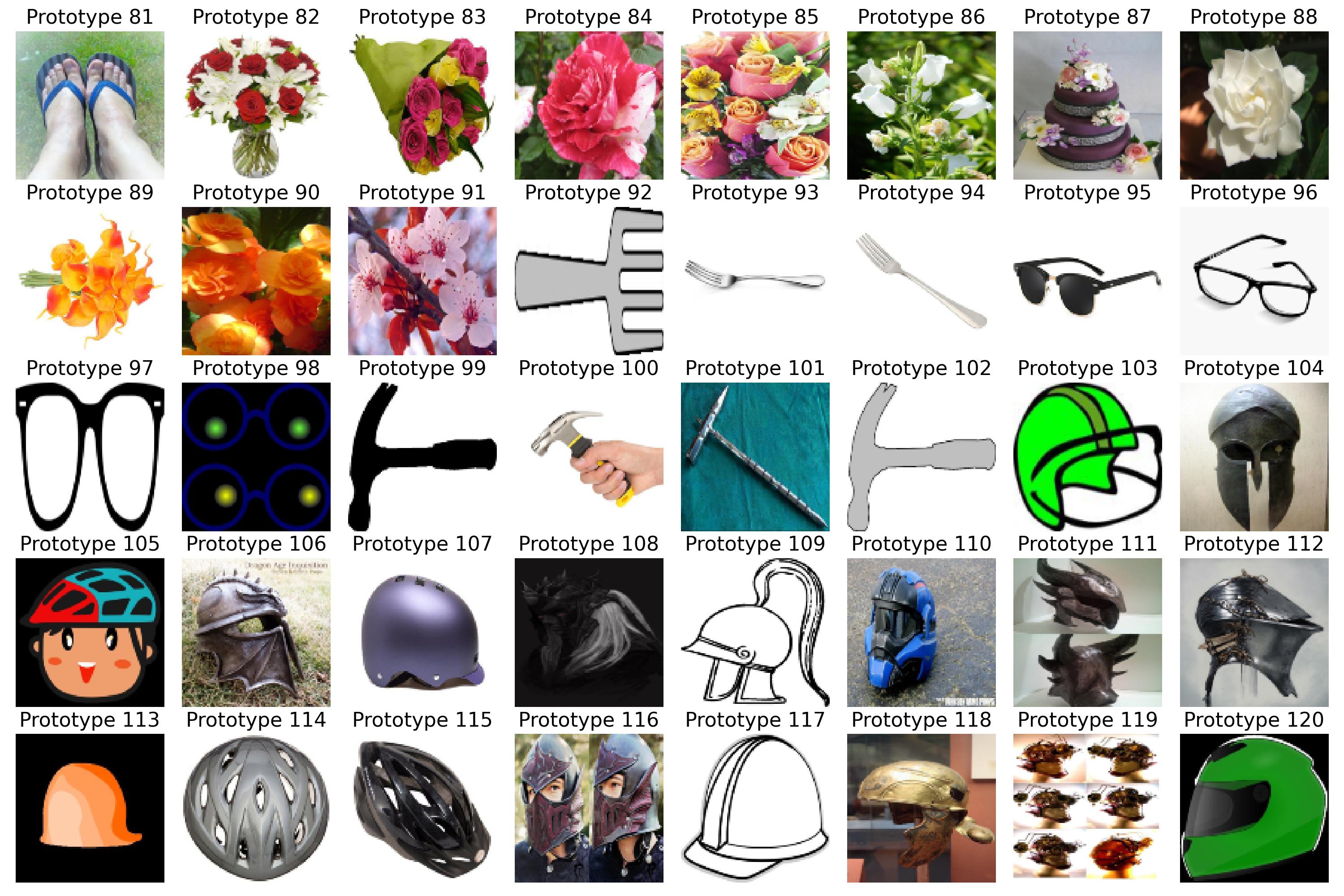}
     \end{subfigure}
     \begin{subfigure}{0.49\textwidth}
         \includegraphics[width=\textwidth]{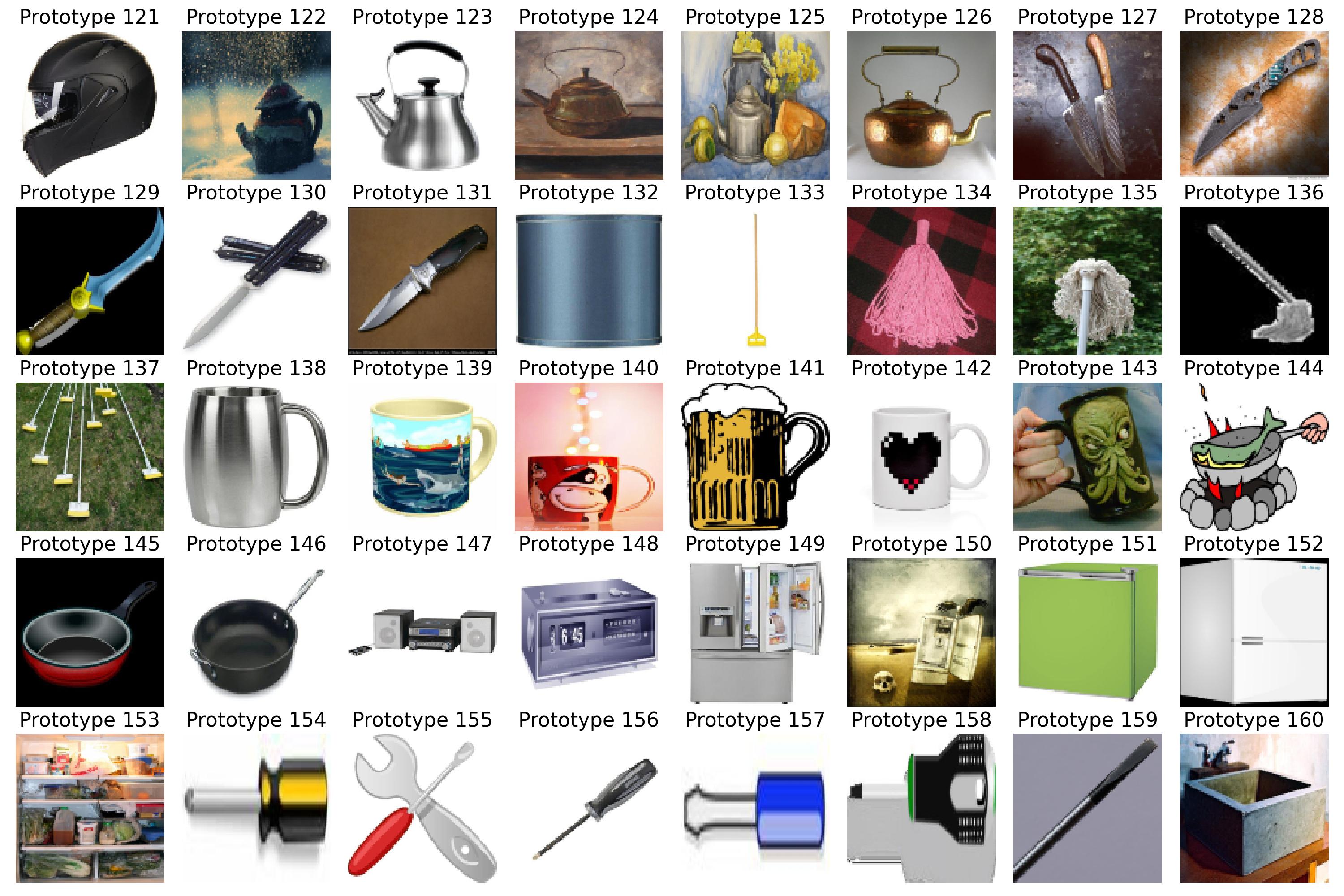}
     \end{subfigure}
     \begin{subfigure}{0.49\textwidth}
         \includegraphics[width=\textwidth]{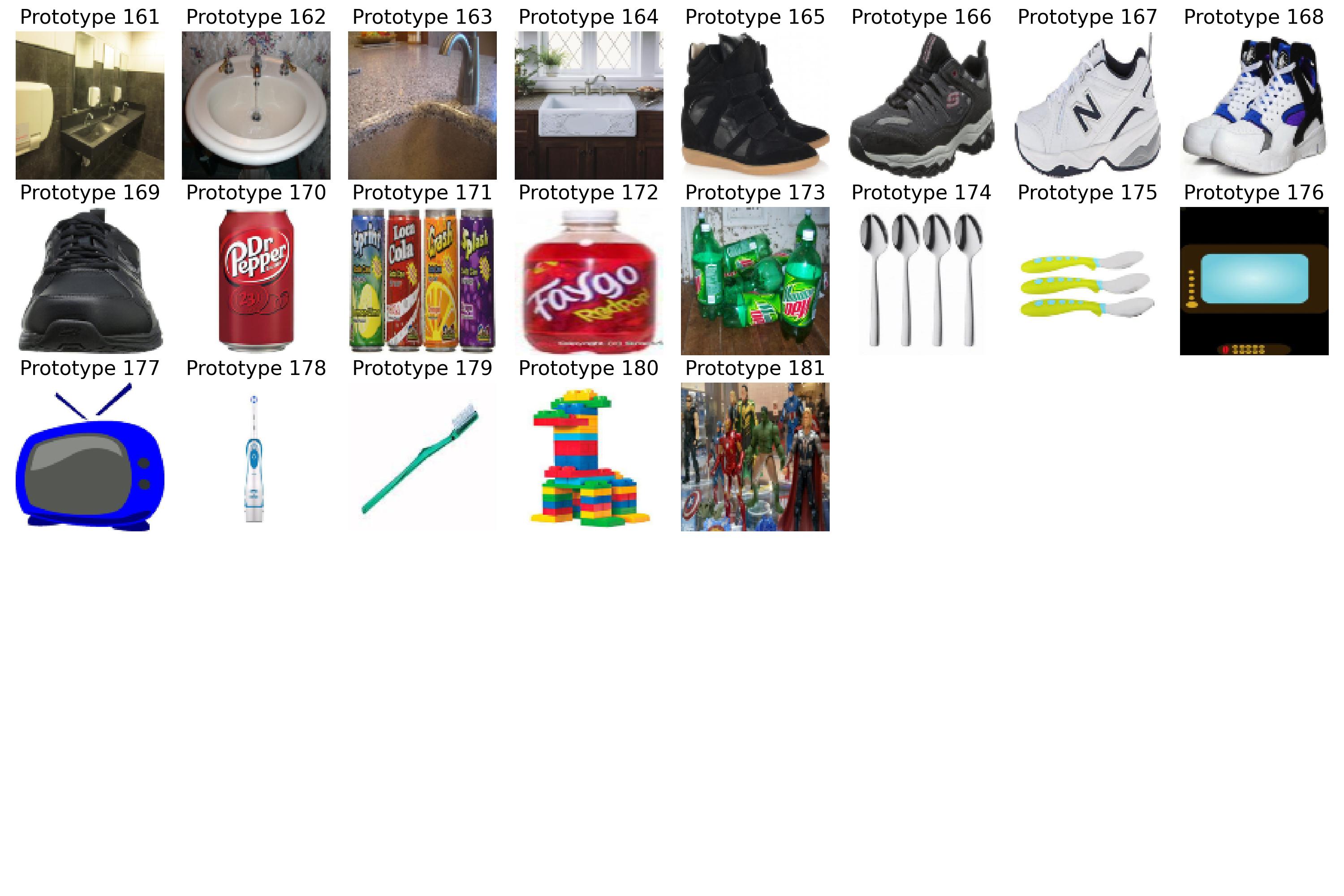}
     \end{subfigure}
     \caption{All learned prototypes for dataset $\mD^\prime$ with home objects.}
      \label{fig:officehomeproto_home_home}
\end{figure}

\subsection{Robustness of Prototype-summarization-based explanations}
\label{appdx:mammo_bootstrap}
In this section, we present the five sets of summarization prototypes learned from the bootstrap versions of the mammography dataset $\mD$ and $\mD^\prime$ described in Section~\ref{sec:exp_mammo}. We are able to reach the same conclusion from all different bootstraps, thus demonstrating the robustness of our approach.
\begin{figure}[H]
    \centering
    \begin{subfigure}{0.49\textwidth}
         \includegraphics[width=\textwidth]{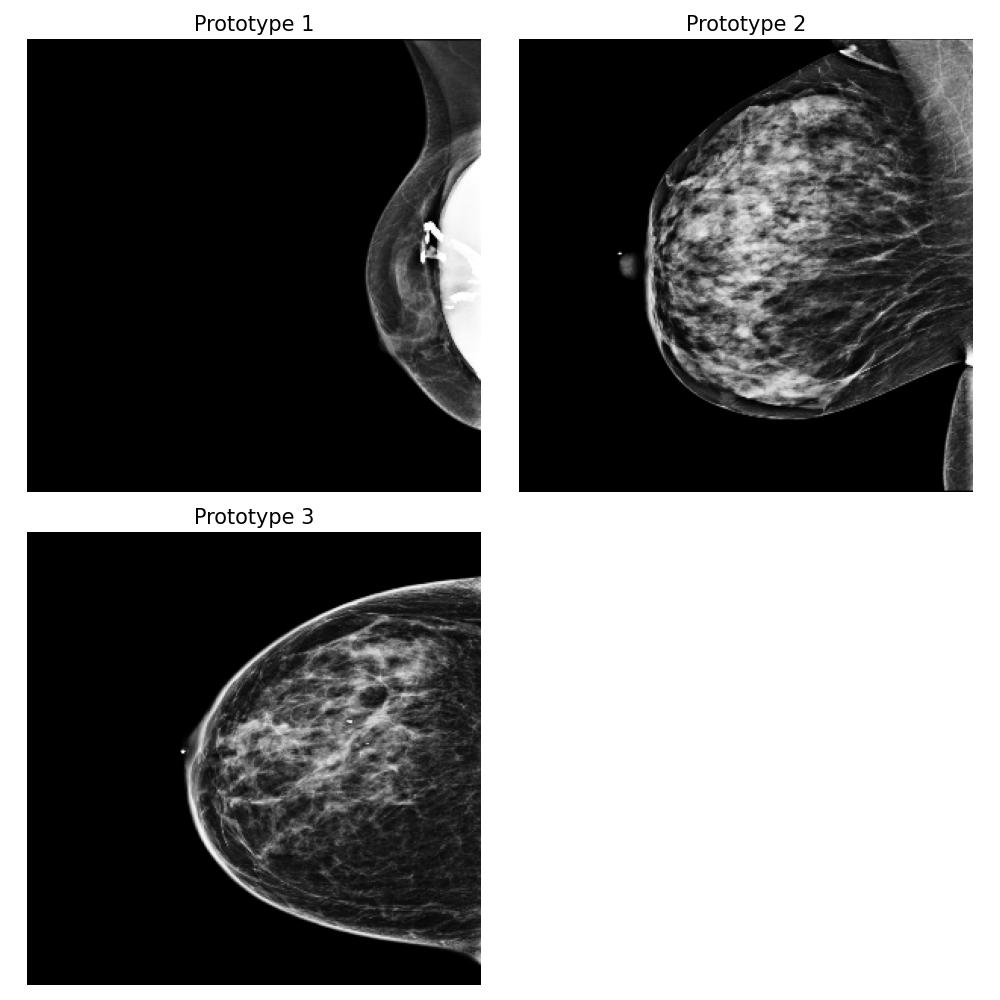}
         \caption{$\mD$ prototypes from bootstrap 1.}
     \end{subfigure}
    \begin{subfigure}{0.49\textwidth}
         \includegraphics[width=\textwidth]{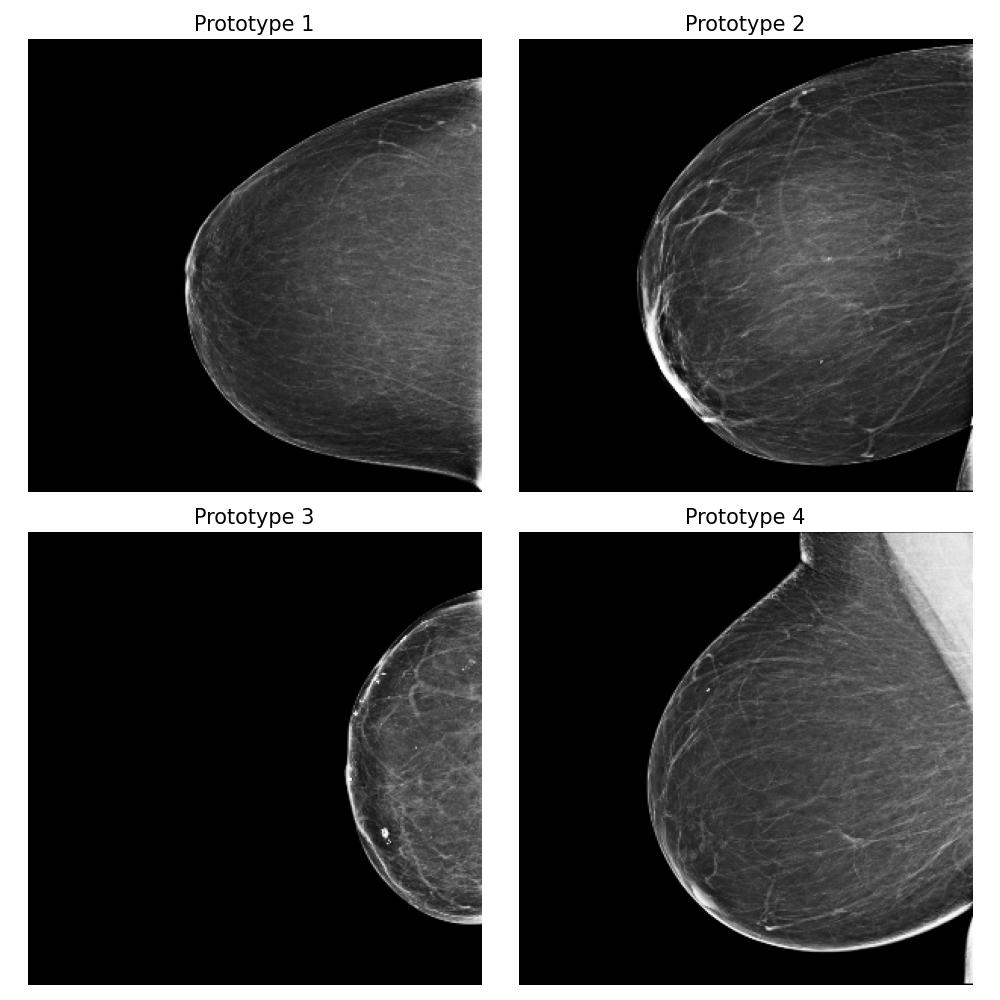}
         \caption{$\mD^\prime$ prototypes from bootstrap 1.}
     \end{subfigure} \\
     \begin{subfigure}{0.49\textwidth}
         \includegraphics[width=\textwidth]{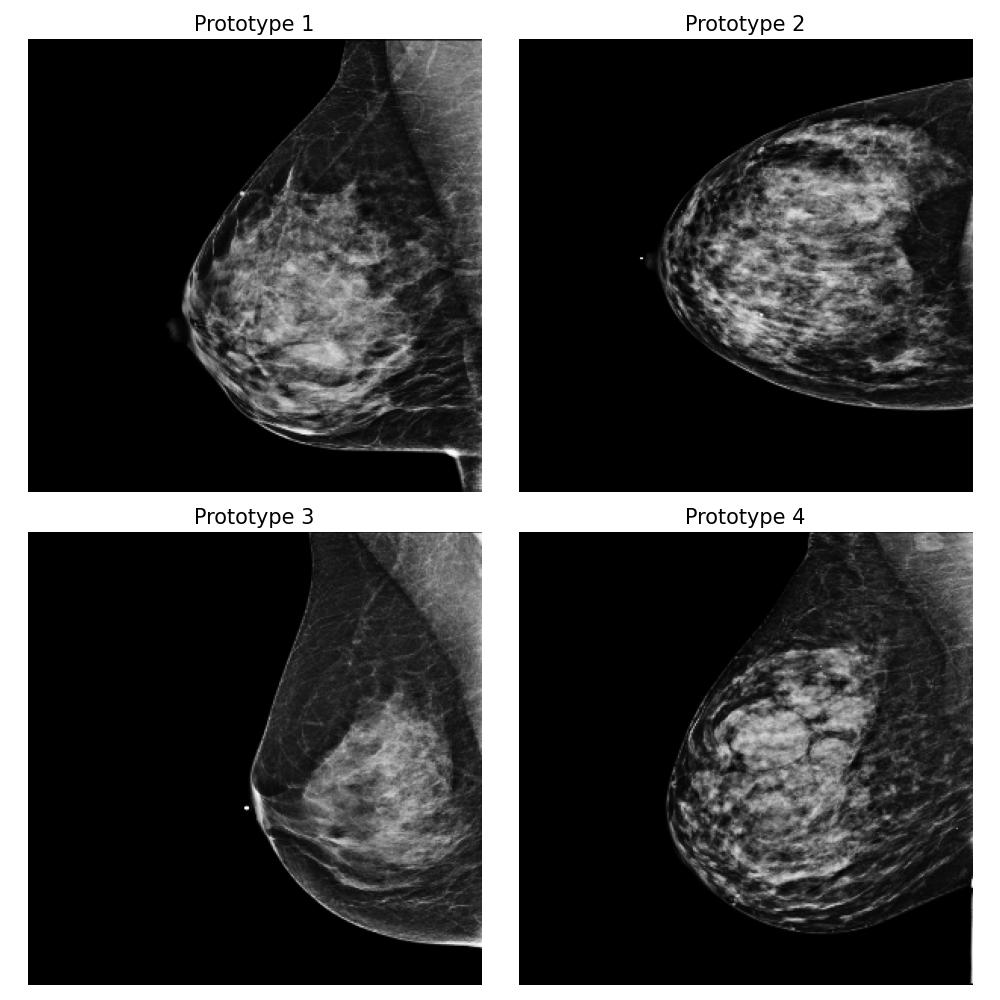}
         \caption{$\mD$ prototypes from bootstrap 2.}
     \end{subfigure}
    \begin{subfigure}{0.49\textwidth}
         \includegraphics[width=\textwidth]{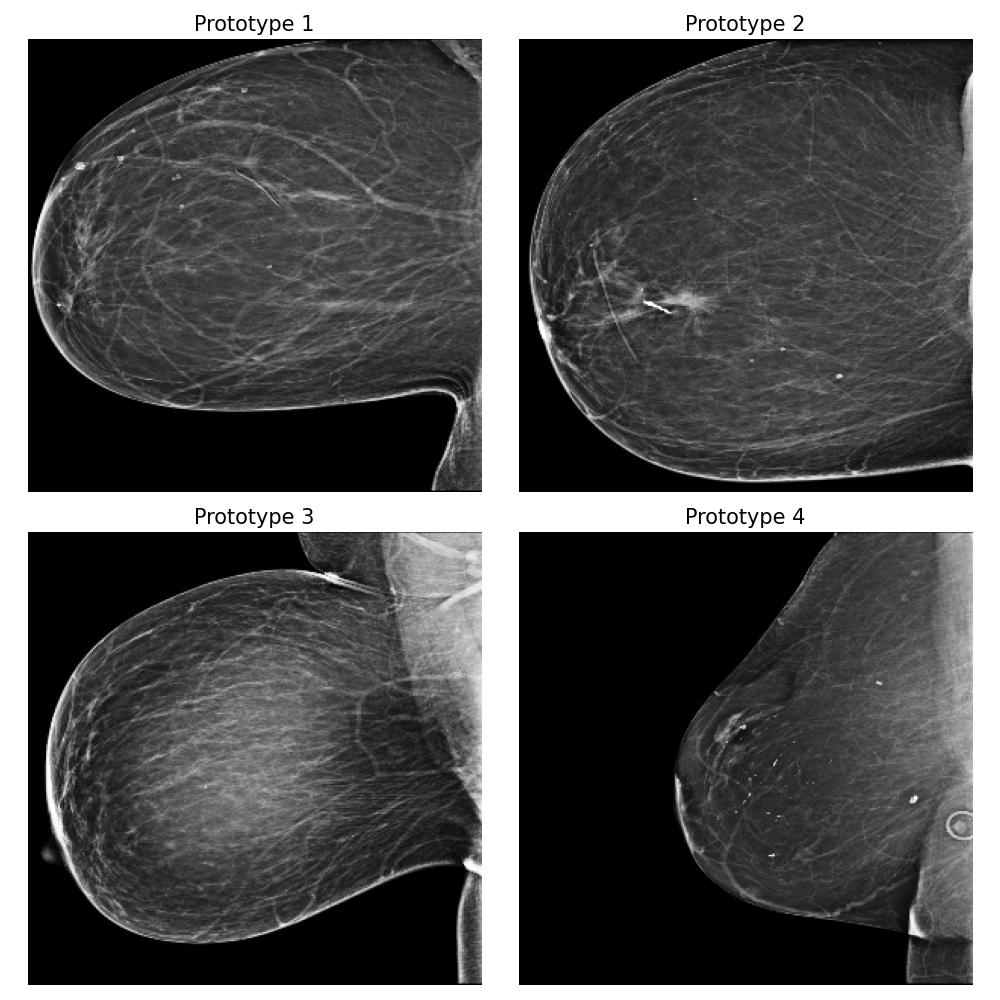}
         \caption{$\mD^\prime$ prototypes from bootstrap 2.}
     \end{subfigure} \\
\end{figure}
\begin{figure}[H]\ContinuedFloat
     \begin{subfigure}{0.49\textwidth}
         \includegraphics[width=\textwidth]{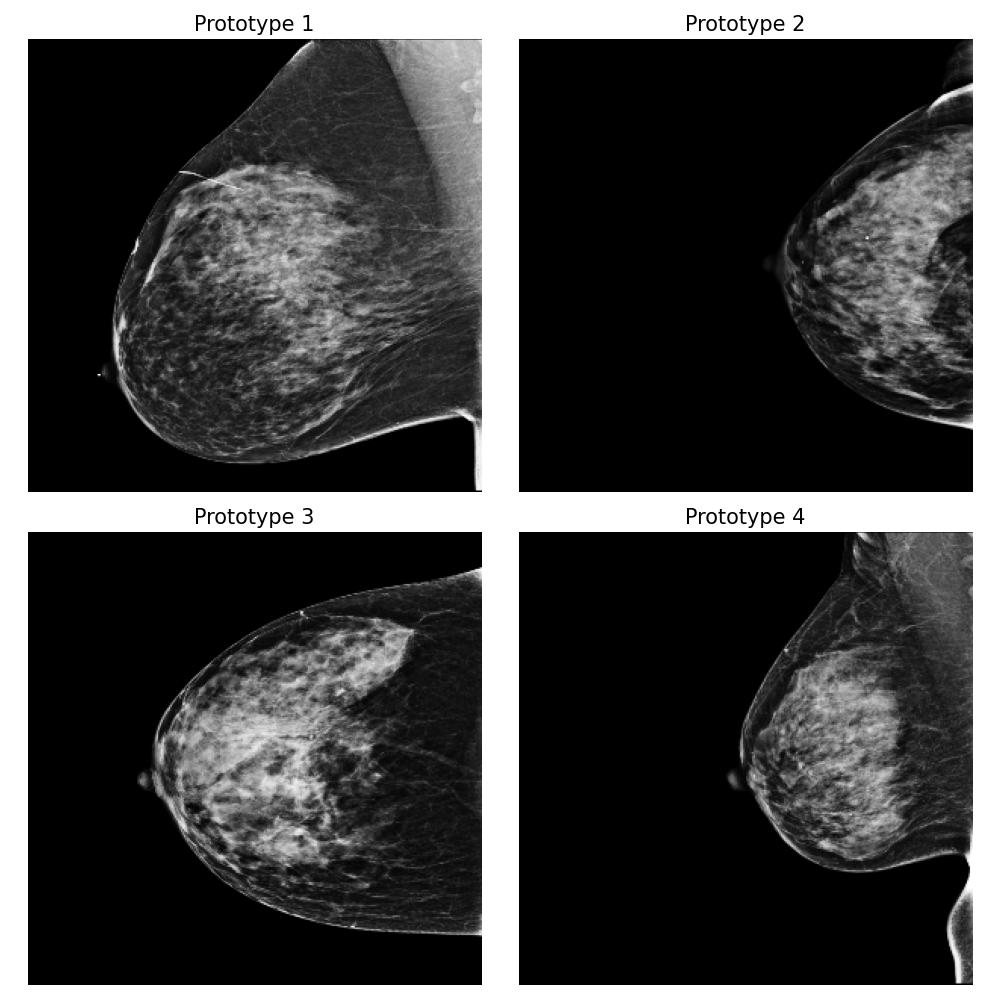}
         \caption{$\mD$ prototypes from bootstrap 3.}
     \end{subfigure}
    \begin{subfigure}{0.49\textwidth}
         \includegraphics[width=\textwidth]{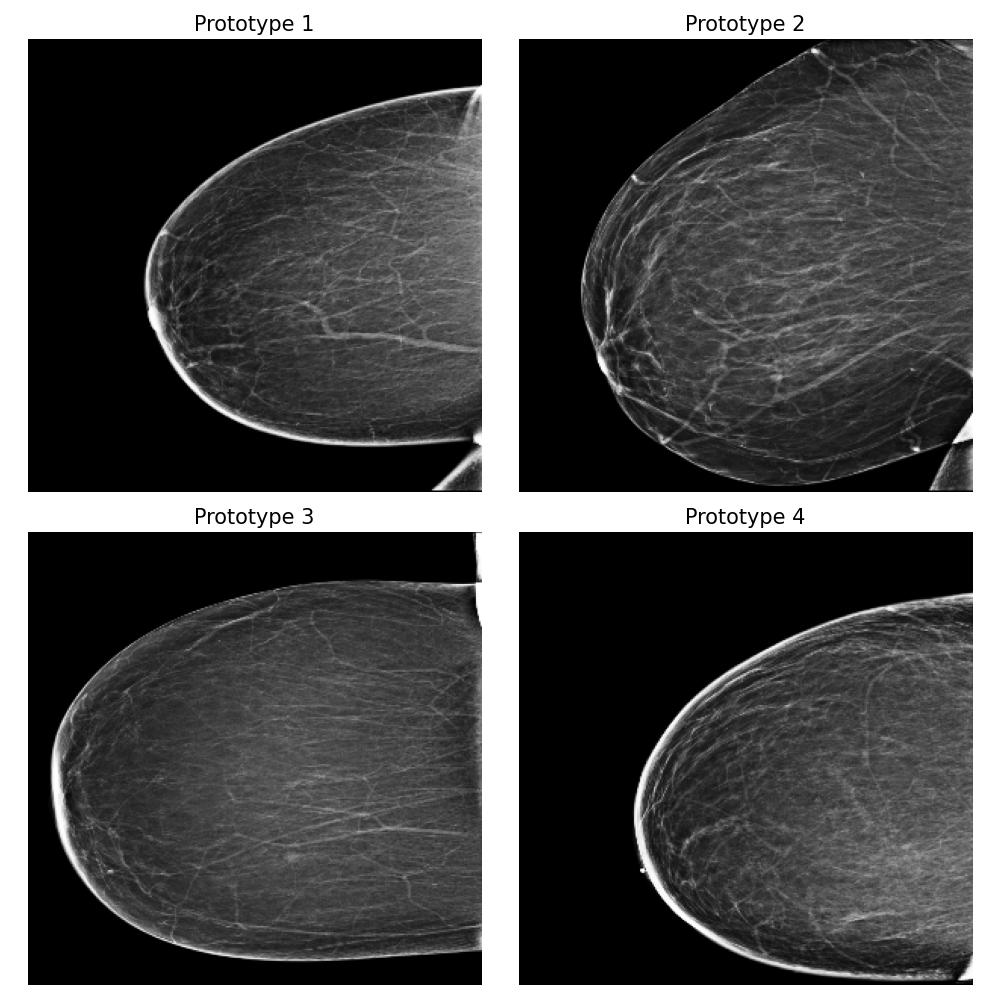}
         \caption{$\mD^\prime$ prototypes from bootstrap 3.}
     \end{subfigure} \\
     \begin{subfigure}{0.49\textwidth}
         \includegraphics[width=\textwidth]{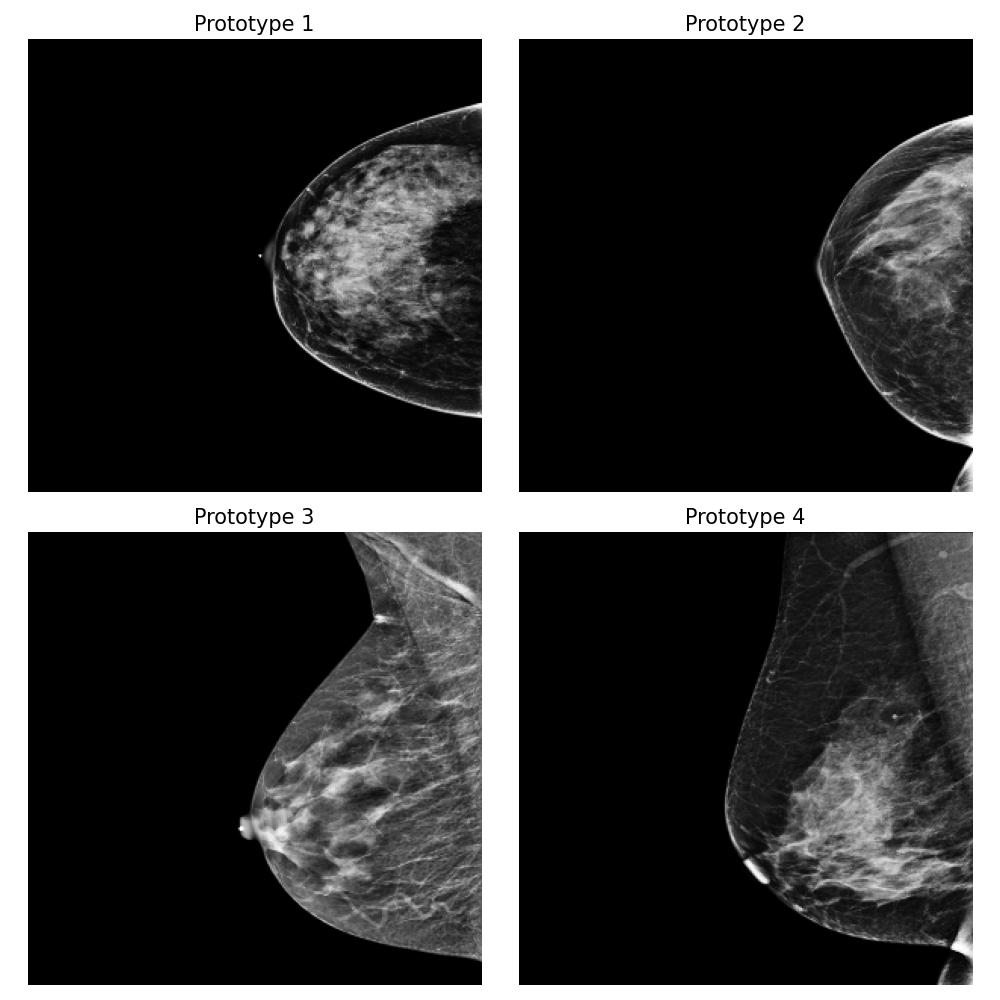}
         \caption{$\mD$ prototypes from bootstrap 4.}
     \end{subfigure}
    \begin{subfigure}{0.49\textwidth}
         \includegraphics[width=\textwidth]{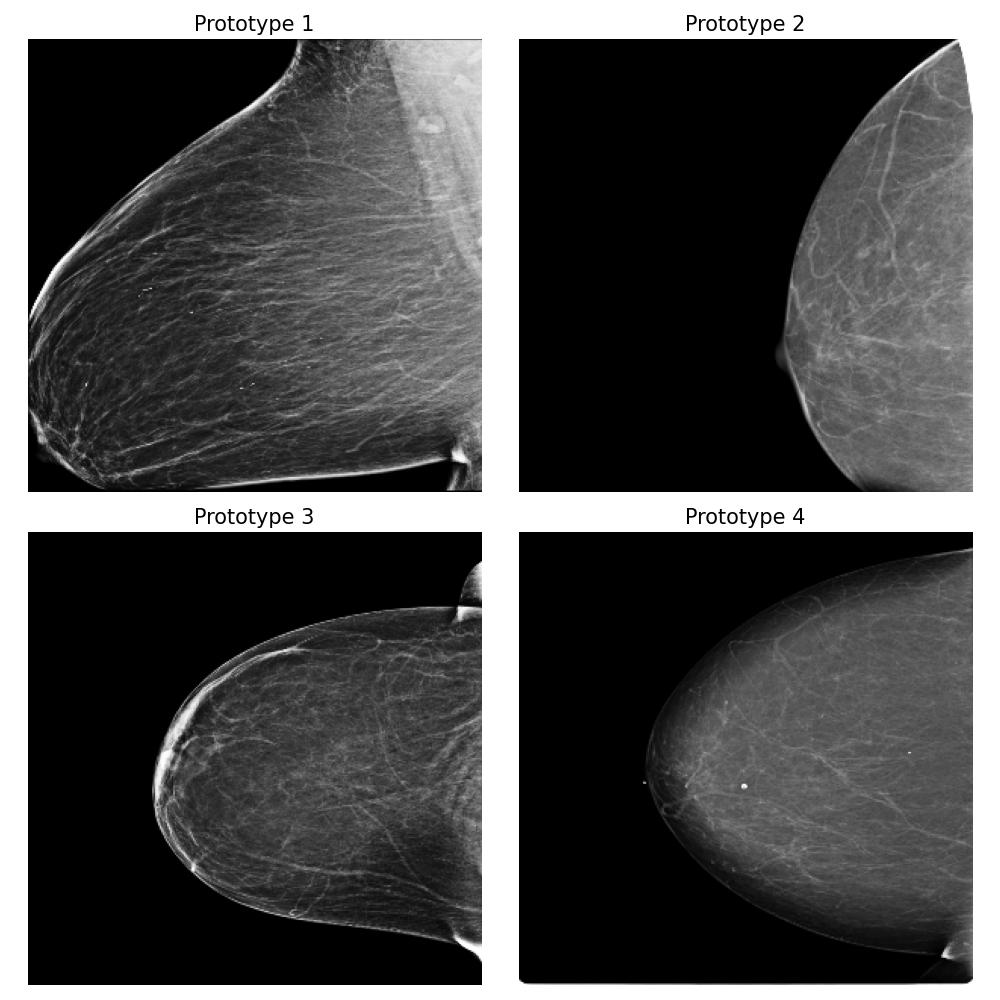}
         \caption{$\mD^\prime$ prototypes from bootstrap 4.}
     \end{subfigure} \\
\end{figure}
\begin{figure}[H]

     \begin{subfigure}{0.49\textwidth}
         \includegraphics[width=\textwidth]{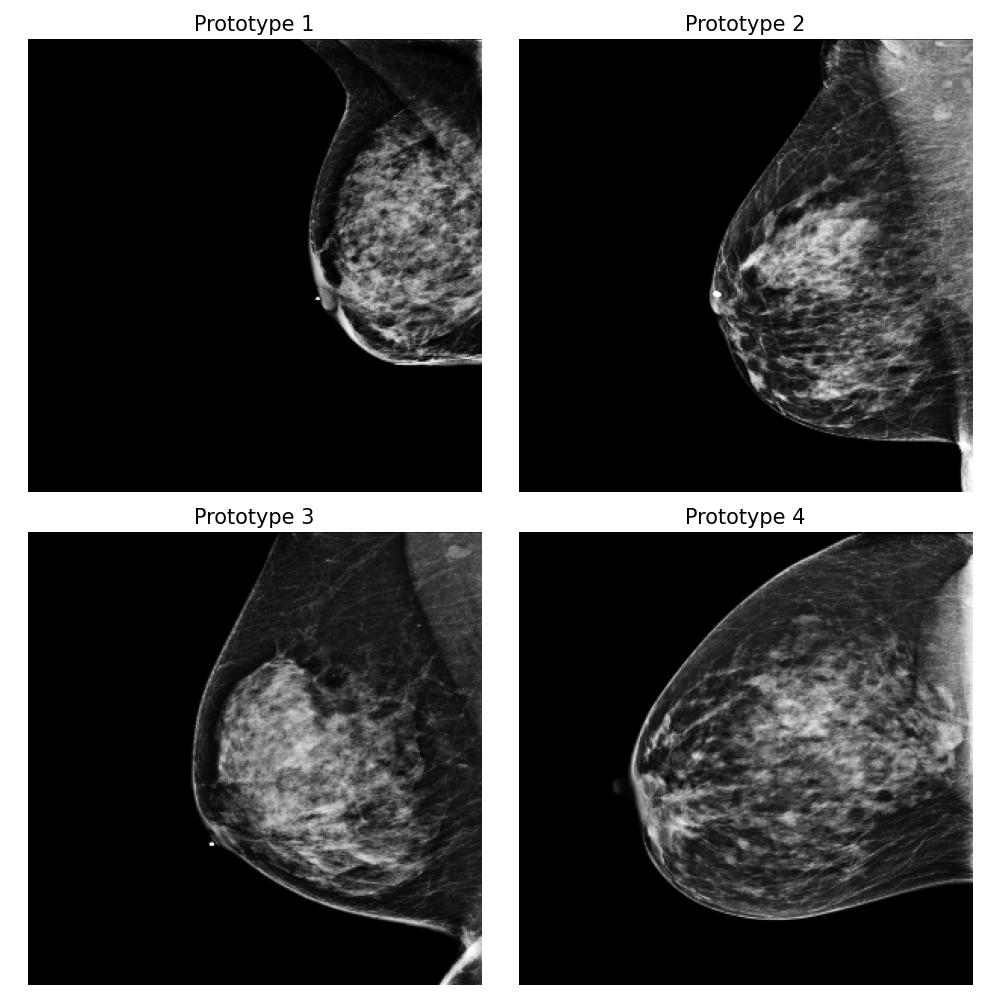}
         \caption{$\mD$ prototypes from bootstrap 5.}
     \end{subfigure}
    \begin{subfigure}{0.49\textwidth}
         \includegraphics[width=\textwidth]{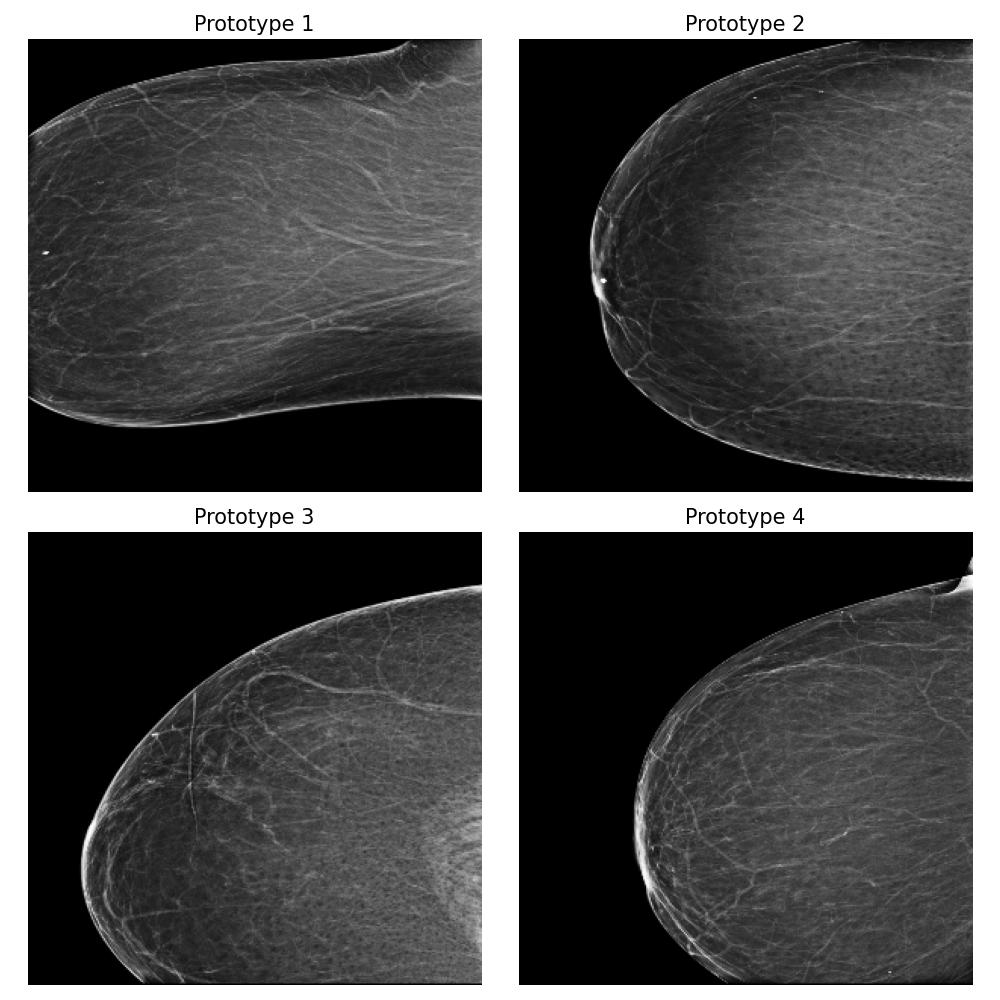}
         \caption{$\mD^\prime$ prototypes from bootstrap 5.}
     \end{subfigure} \\
     \caption{All learned prototypes for dataset $\mD^\prime$ with home objects. The model only learned $3$ unique prototypes for Bootstrap $1$ of $D$.}
     \label{fig:mammo_bootstrap_fig}
\end{figure}
\subsection{Are Prototypical Neighborhoods Faithful for Vision and Signal Data?}
\label{sec:10neigh}
We examine the robustness and faithfulness of the learned prototype neighbourhoods from our summarization approach in each experiment. To do so, we visualize a selection of prototypes along with 10 nearest samples for each prototype. We can see that all the models have learned a high quality latent space, as neighbouring samples appear to be very similar to the prototype. Note that while Bootstrap 1 in Figure~\ref{fig:mammo_bootstrap_fig}a only learned 3 unique prototypes for $D$, this does not detract from our observations.

\begin{figure}[hbt]
\centering
    \includegraphics[width=0.98\textwidth]{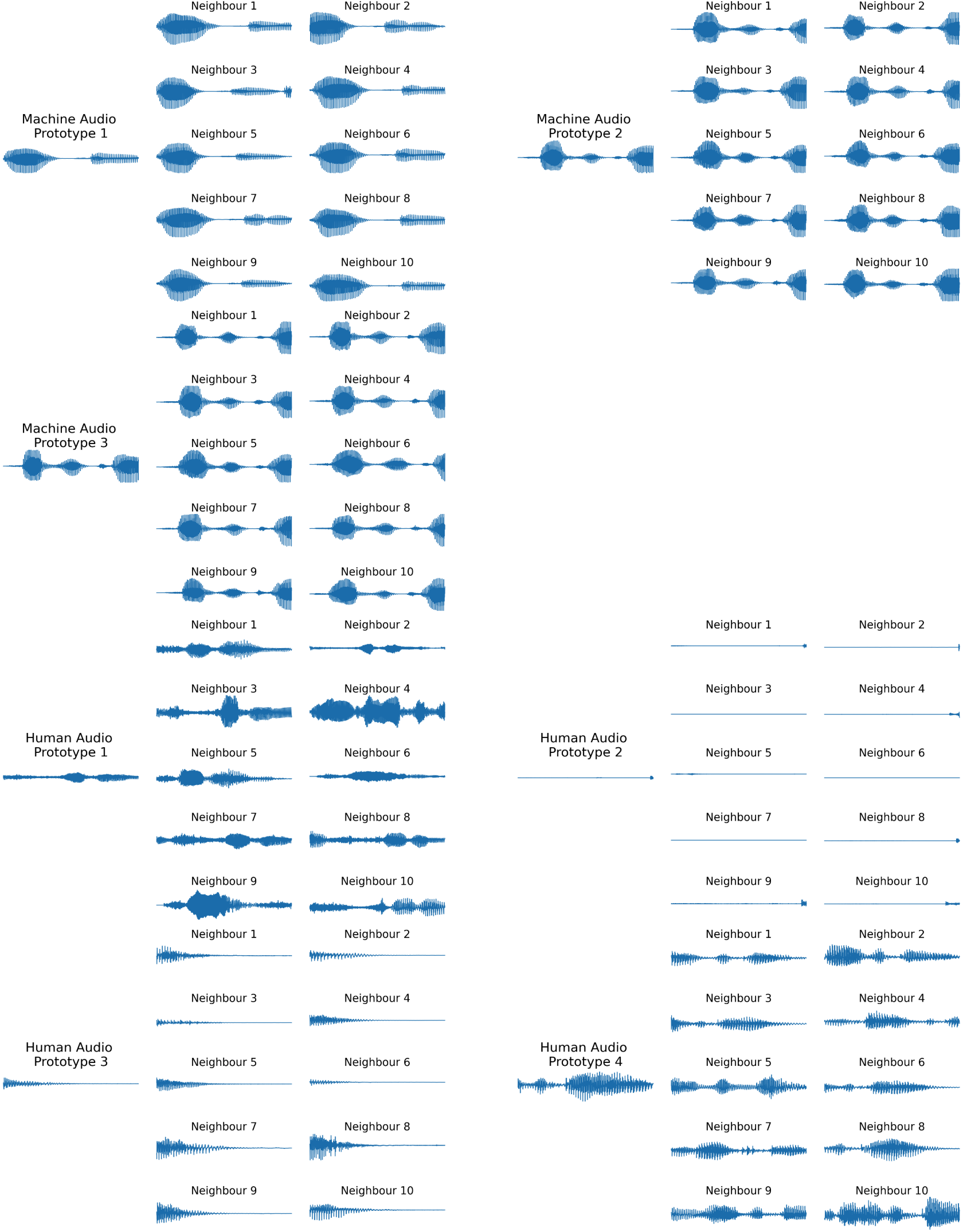}
    \caption{Neighbouring sample visualization for audio prototypes learned in Section~\ref{sec:exp_audio}.}
    \label{fig:10neigh_audio}
\end{figure}

\begin{figure}[hbt]
\centering
    \includegraphics[width=0.98\textwidth]{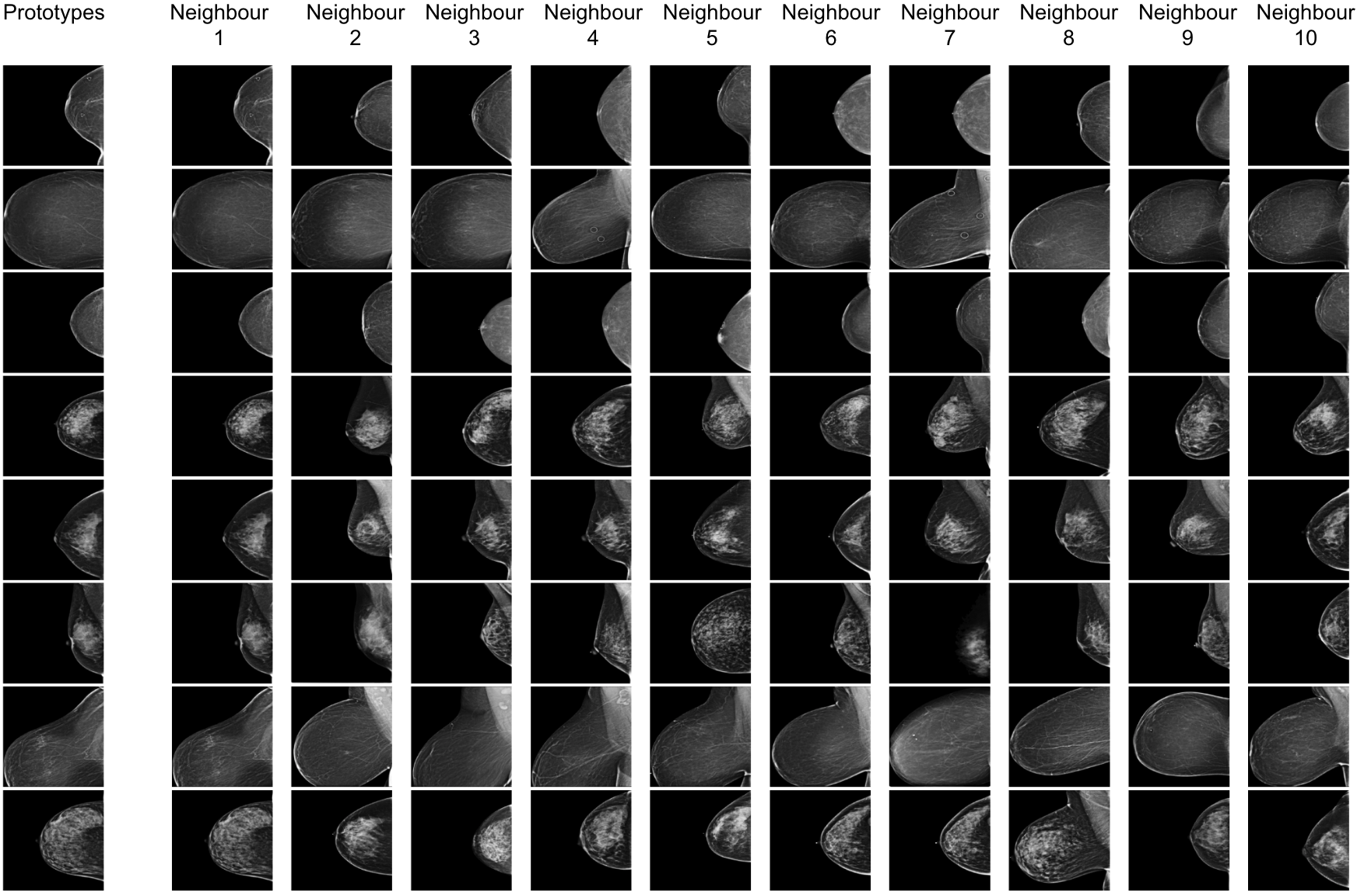}
    \caption{Neighbouring sample visualization for mammography prototypes learned in Section~\ref{sec:exp_mammo}.}
    \label{fig:10neigh_mammo}
\end{figure}

\begin{figure}[hbt!]
\centering
    \includegraphics[width=0.98\textwidth]{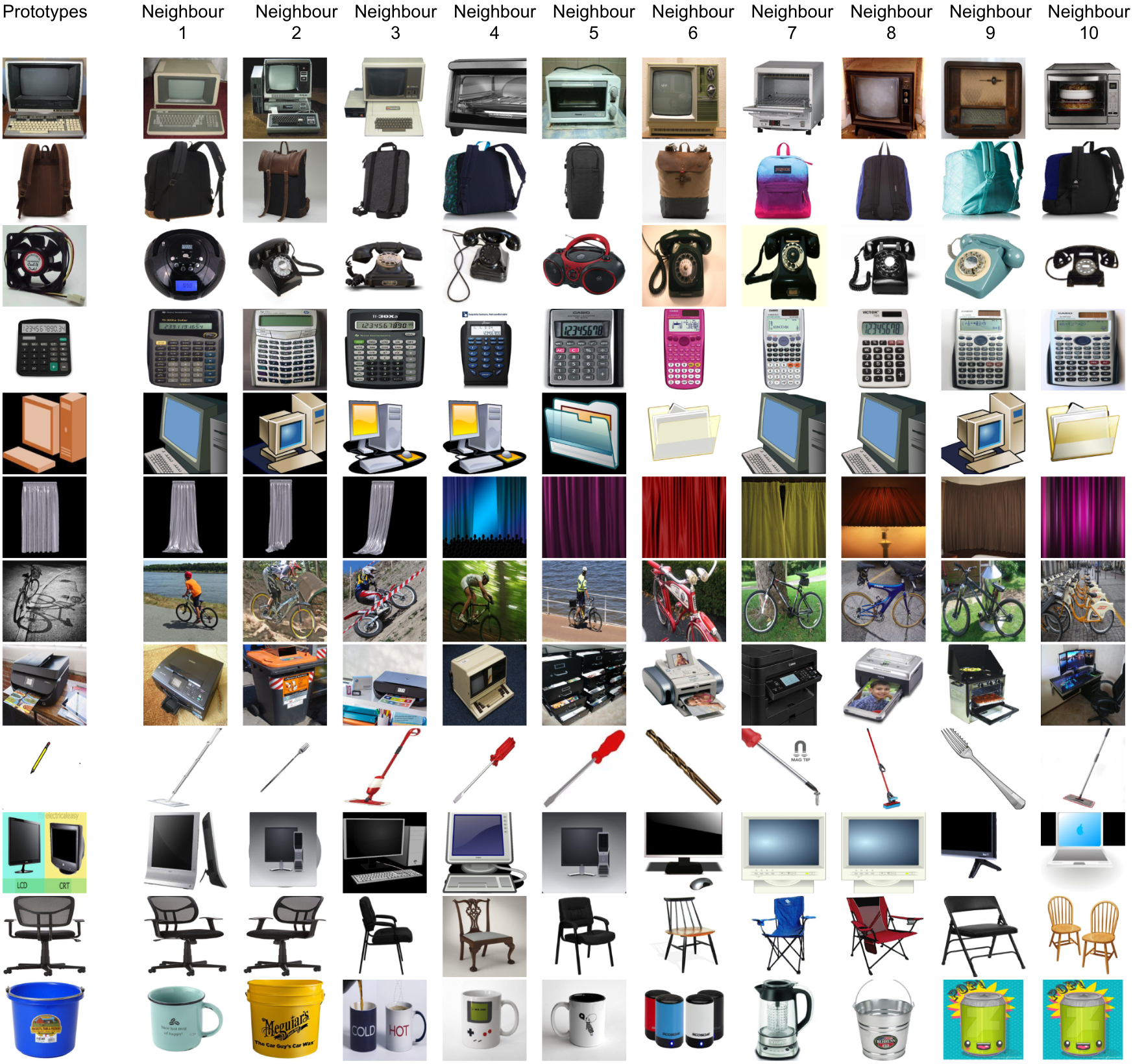}
    \caption{Neighbouring sample visualization for office home object prototypes learned in Section~\ref{sec:exp_officehome}.}
    \label{fig:10neigh_officehome}
\end{figure}

\begin{figure}[hbt!]
\begin{subfigure}[b]{0.48\textwidth}
    \centering
    \includegraphics[width=0.7\textwidth]{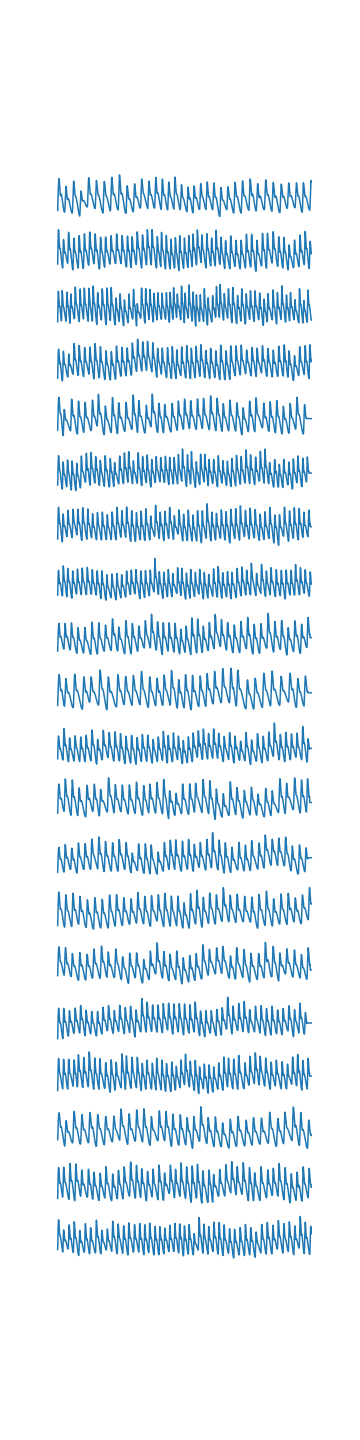}
    \caption{Examples of synthetic PPG signals.}
    \label{fig:synth_ppg}
\end{subfigure}
\begin{subfigure}[b]{0.48\textwidth}
    \centering
    \includegraphics[width=0.7\textwidth]{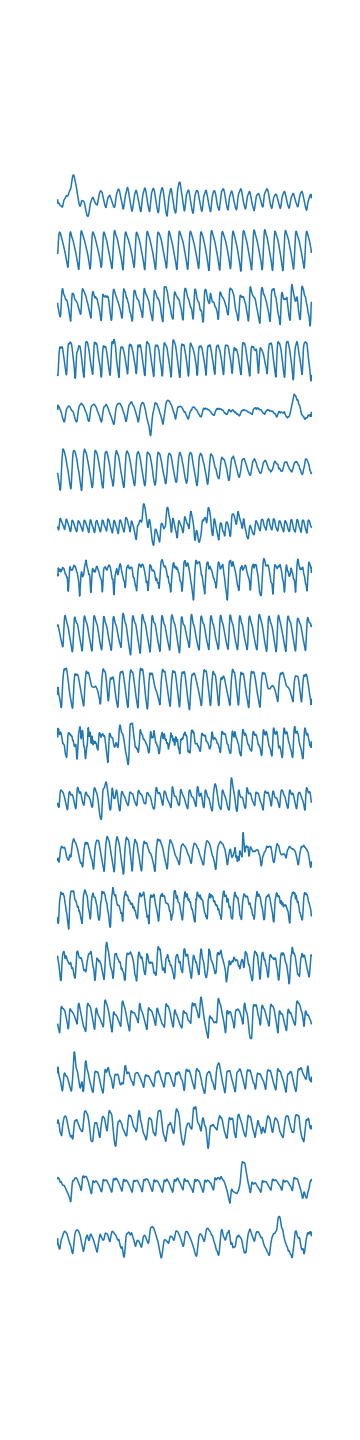}
    \caption{Examples of real PPG signals.}
    \label{fig:real_ppg}
\end{subfigure}
\caption{Here we show 20 samples each of both real and synthetic PPG signals.}
\label{fig:real_synth_ppg}
\end{figure}
\FloatBarrier

\section{Evaluation of Influential Example Explanations}
\subsection{Influential Example Explanations: Alignment}
We now evaluate influential example explanations by asking the question:
\textit{Given datasets $\mD$ and $\mD^\prime$ and their respective models, are we choosing the right examples from either dataset to remove to remediate feature importance differences between the datasets?} \\
The first step in the explanation pipeline is training a discriminator to classify whether the local intrinsic feature importance (LiFIM) for an example originates from $\mD$ or $\mD^\prime$ (see Algorithm~\ref{alg:fi_alignment}). Then, by computing influences for LiFIM, we determine the appropriate examples that have the highest (positive) influence on the discriminator loss (i.e., removing them increases the loss). We first determine the validity of our method by computing the theoretical influences of each example in $\mD$ and $\mD^\prime$ using Equation~\ref{eqn:influence} and then empirically calculating the loss of the discriminator after the example is removed from training. This is shown in Figure~\ref{fig:influence_vs_test_loss} below. Here, the empirical estimates match the theoretically computed values for the influence. We remark here that these are results for logistic regression, whereas the correlation may not be as high for nonlinear functions, though other works have found high correlations \citep{influence}.
 
\begin{figure}[H]
    \centering 
    \includegraphics[width=0.49\textwidth]{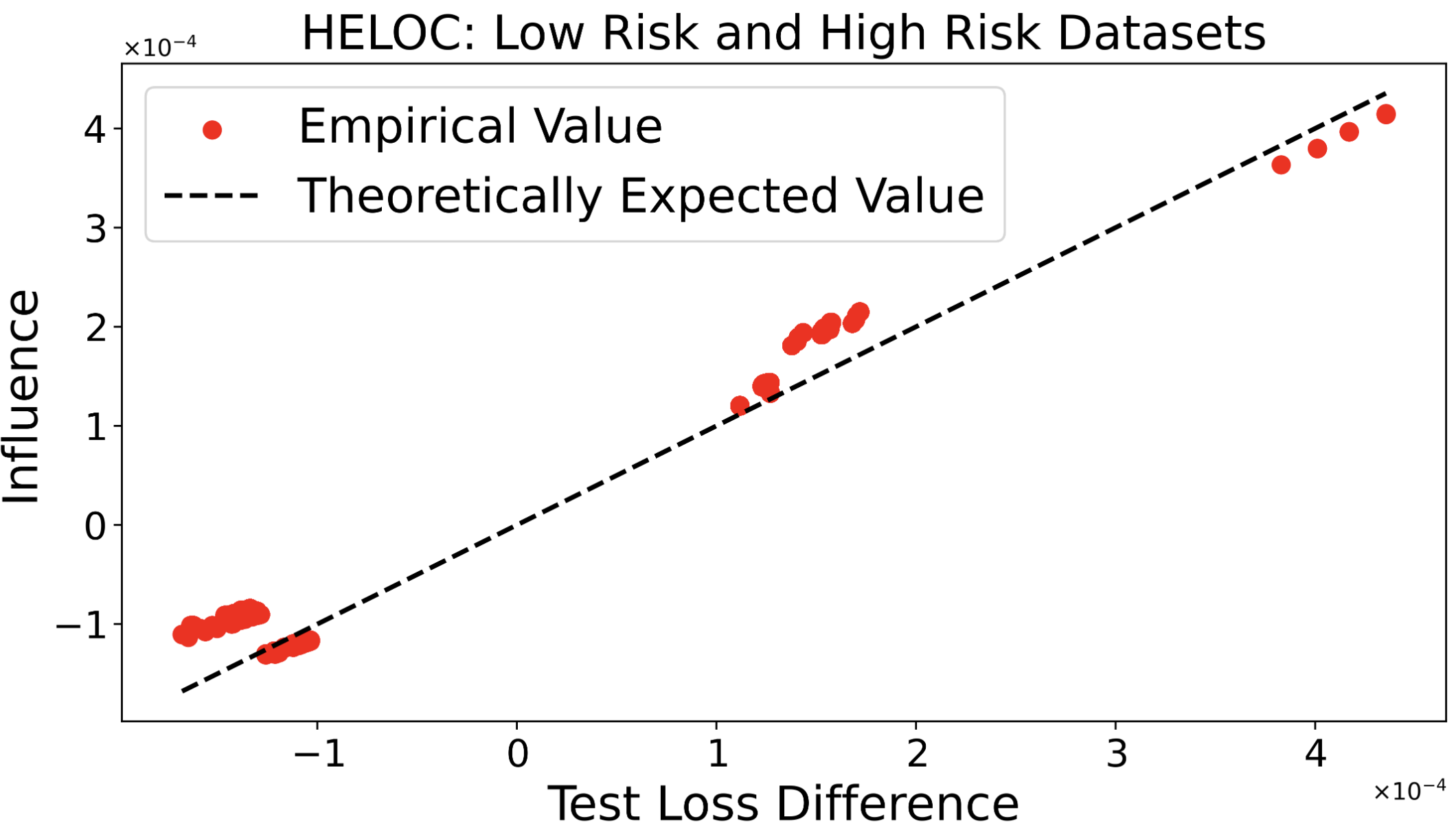}
    \includegraphics[width=0.49\textwidth]{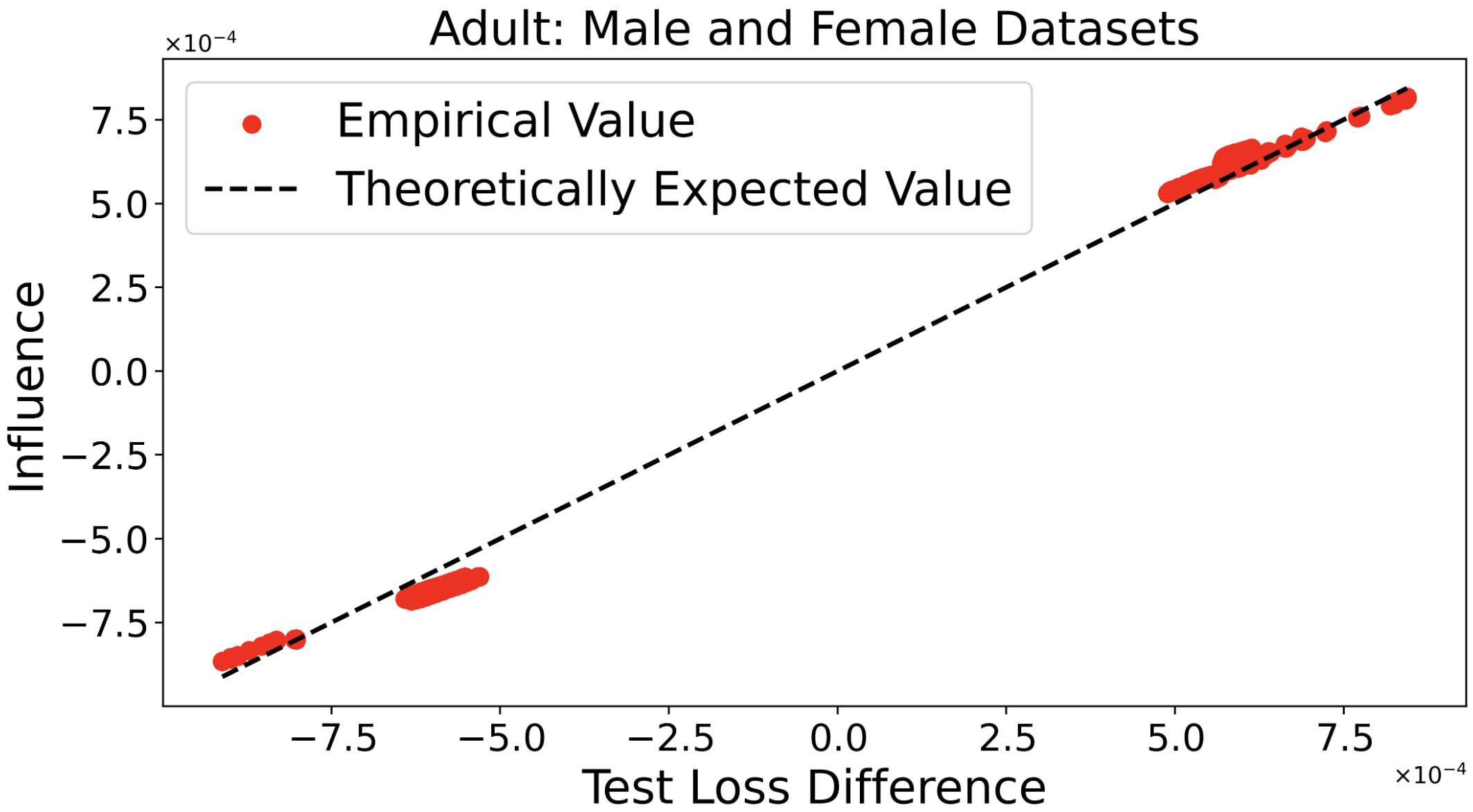}
    \caption{Theoretical Influence vs Empirical Change in discriminator loss for HELOC (left) and Adult (right) datasets. Each point corresponds to an example (represented by its feature importance vector) in the dataset $\mD \cup \mD^\prime$. We compute the theoretical influence of each example and compare it with the empirical test loss obtained after removing the example from the training set.}
    \label{fig:influence_vs_test_loss}
\end{figure}
\begin{figure}[H]
    \centering
    \includegraphics[width=0.49\textwidth]{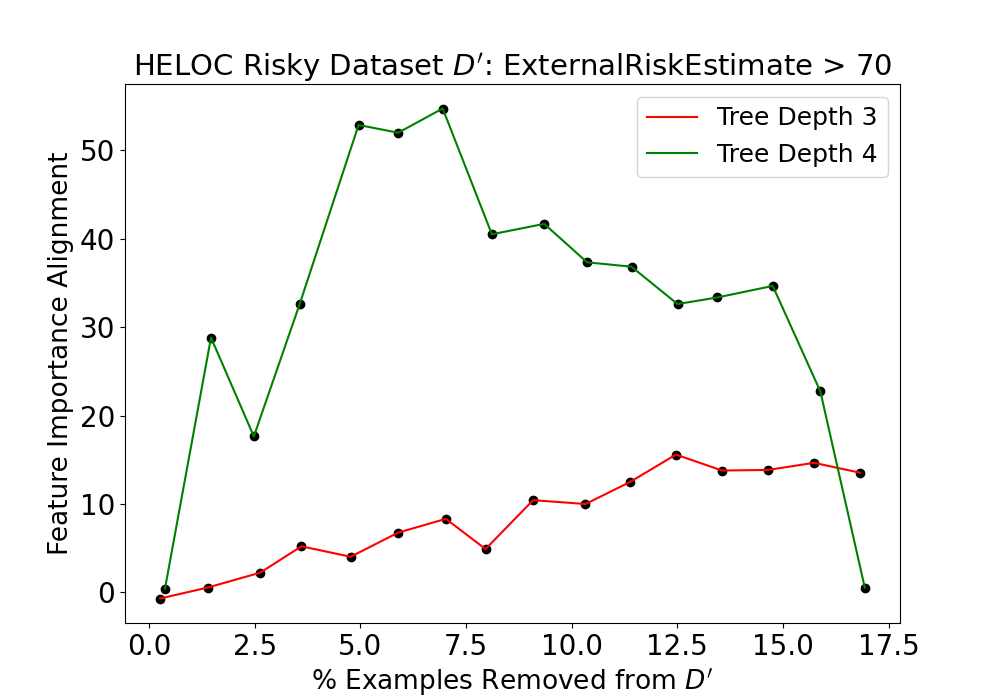}
    \includegraphics[width=0.49\textwidth]{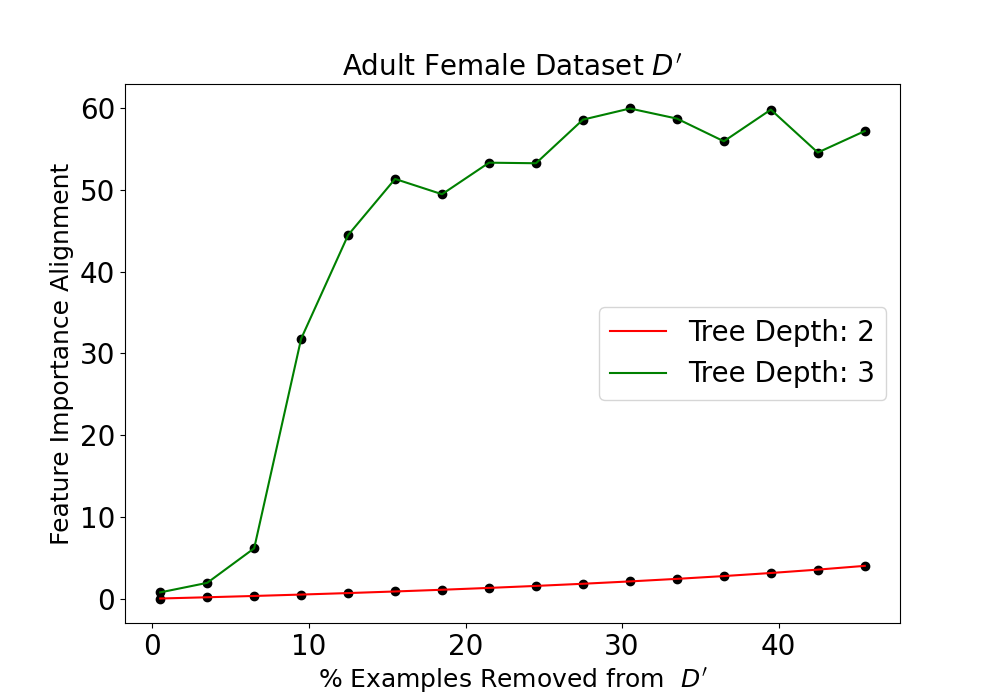}
    \caption{Feature Importance Alignment vs $\%$ Examples Removed from $\mD^\prime$ for HELOC and Adult datasets. There are two distinct regimes here. These regimes correspond to removing all the positively influential points (which make the distributions different), and then when we run out of those points, we start to remove some additional points that are actually keeping the distributions similar.
    In the first regime, the number of examples removed is small relative to the dataset size. In this regime, increasing the number of examples causes an increasing alignment between the GiFIMs of $\mD$ and $\mD^\prime \backslash S$, thereby reducing the error.
    In the second regime, the number of examples removed is no longer insignificant relative to the dataset size. Here, in most cases,  increasing the number of examples causes a plateau or reversal in alignment (i.e., the error increases). We hypothesize that this occurs because there are only a certain number of examples in a dataset with positive influence on the alignment. Once these examples have been removed, the remaining examples will have an increasingly negative influence on alignment. }
    \label{fig:num_removed_fi_alignment_adult}
\end{figure}

In the next step, we want to see if the removed examples actually change the global feature importance measure when the task model is retrained.
\begin{itemize}
    \item Without loss of generality, we choose the dataset we are removing examples from as $\mD^\prime$. For HELOC, this corresponds to the High-Risk dataset with \texttt{ExternalRiskEstimate} $ \leq 70$. For Adult, this is the Female dataset.
    \item We remove a varying $\%$ of the most influential examples from $\mD^\prime$ decided by Algorithm~\ref{alg:fi_alignment}. Let $S$ be the set of examples removed and $f_{\mD^\prime}$ and $f_{\mD^\prime \backslash S}$ the task specific models trained on $\mD^\prime$ and $\mD^\prime\backslash S$ respectively. 
    \item We then compute the average global feature importance alignment between the dataset $\mD$ and the datasets $\mD^\prime$ and $f\mD^\prime \backslash S$ respectively. Given GiFIMs $\phi_g(\mD)$, $\phi_g(\mD^\prime)$, and $\phi_g(\mD^\prime\backslash S)$ for $\mD$, $\mD^\prime$, and $\mD^\prime\backslash S$ respectively, alignment is defined as:
    \begin{equation}
        \textrm{Alignment} = \frac{\|\phi_g(\mD) - \phi_g(\mD^\prime)\| - \|\phi_g(\mD) - \phi_g(\mD^\prime\backslash S)\|}{\|\phi_g(\mD) - \phi_g(\mD^\prime)\|}
    \end{equation}
    or the $\%$ reduction in error between GiFIMs of $\mD$ and $\mD^\prime$ once the influential examples are removed from consideration.
    \item For the task models, we experiment with decision trees of different depths to see if there is any impact on alignment. Figure \ref{fig:num_removed_fi_alignment_adult} shows that removing a small number of examples can improve the alignment in feature importances of $\mD$ and $\mD^\prime$ - however, there is a limit to this.
\end{itemize}

\subsection{Influential Example Explanations: Robustness}
\begin{figure}[H]
    \centering
    \includegraphics[width=0.48\linewidth]{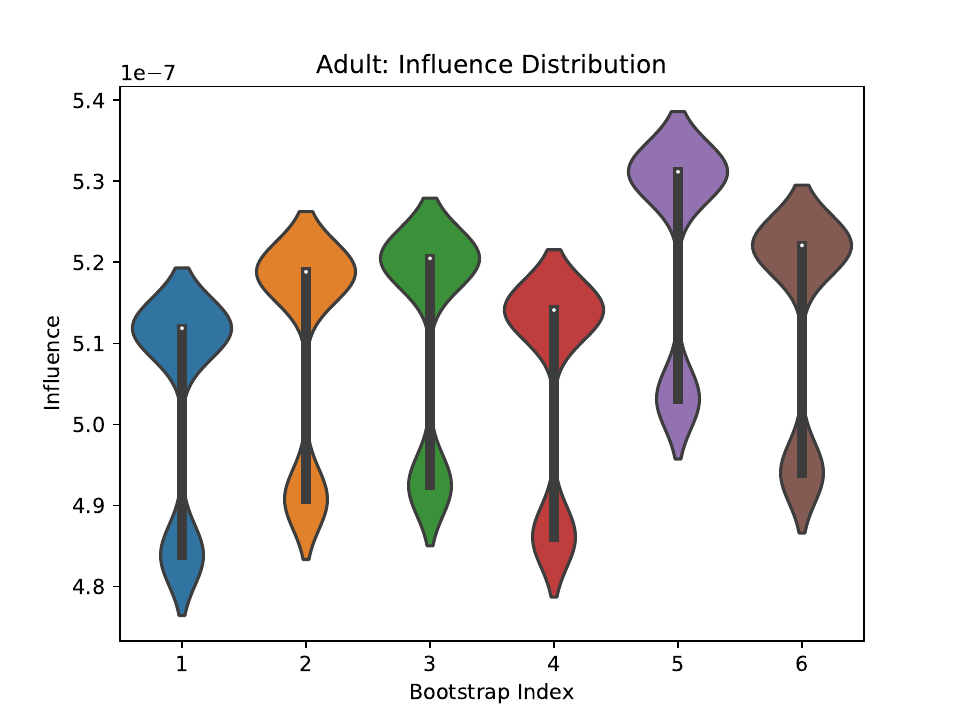}
    \includegraphics[width=0.48\linewidth]{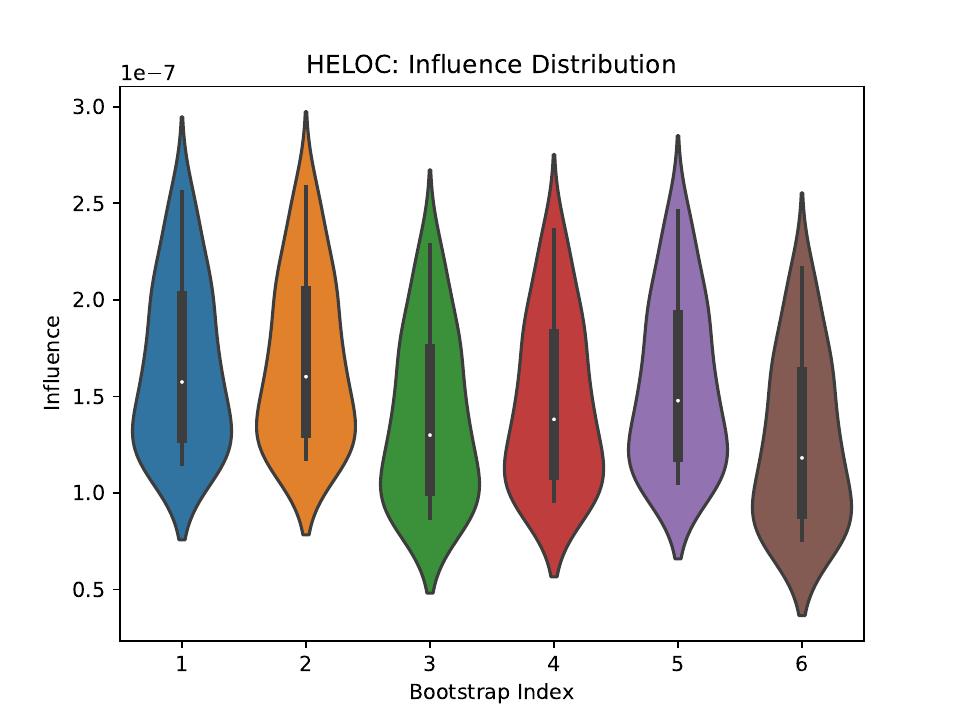}
    \caption{Influence distribution of $50$ examples in the Adult and HELOC datasets with the highest influence scores. We see that this remains consistently stable across bootstraps, implying that our explanations will also remain stable.}
    \label{fig:influence_distribution}
\end{figure}

\begin{figure}[H]
    \centering
    \includegraphics[width=0.27\linewidth]{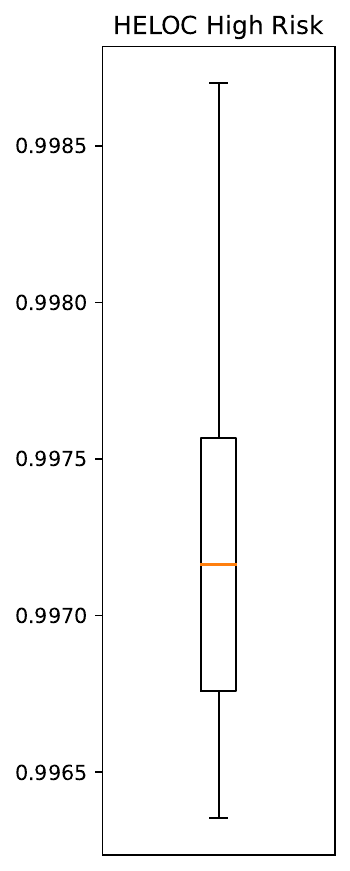}
    \vspace{0.1cm}
    \includegraphics[width=0.26\linewidth]{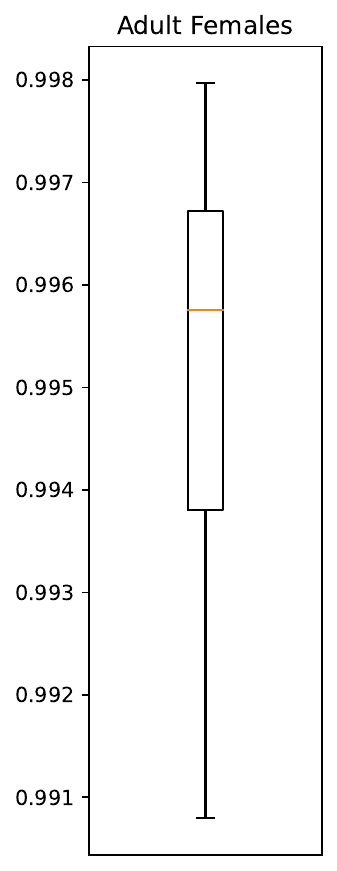}
    \caption{Kendall Tau Ranking Similarity of influential examples in $\mD$ across different bootstraps of $\mD^\prime$. For HELOC, we show how influential examples in the high-risk HELOC dataset (as defined as in Section~\ref{sec:fi_exp_heloc}) are affected as we bootstrap the low risk HELOC dataset. For Adult, we show how influental examples in the Female dataset are affected as we bootstrap the Male dataset. Ultimately, we see that the similarity across bootstraps is high, suggesting that the identified influential examples in $\mD$ are consistently recognized across different realizations of the dataset $\mD^\prime$. This shows that our explanations are robust to dataset perturbations.}
    \label{fig:kendall_tau_heloc}
\end{figure}

In this section, we evaluate whether our influential example explanations are sensitive to perturbations of the datasets. At a high level, we have two desiderata: 
\begin{itemize}
    \item The most influential examples exhibit similar influence scores across bootstrapped datasets. We show this in Figure \ref{fig:influence_distribution}.
    \item The ranking of the most influential examples is similar across bootstrapped datasets (i.e. the same originally influential examples remain influential). We show this in Figure \ref{fig:kendall_tau_heloc}. 
\end{itemize}
To do this, we perform the following procedure: 
\begin{itemize}
    \item We first fix one of the datasets - call this $\mD$. 
    \item Next, we generate $N = 5$ bootstrapped datasets from $\mD^\prime$ by sampling with replacement. 
    \item For each bootstrapped dataset $\mD^\prime_b$, we compute the influential example explanation between $\mD$ and $\mD^\prime_b$. This involves storing the $50$ most influential examples from $\mD$ using Algorithm~\ref{alg:fi_alignment} and recording their influence scores. 
\end{itemize}
Let the set of the $k$ most influential examples in $\mD$ explaining the difference in feature importances between $\mD$ and $\mD^\prime$ be denoted as $E_\mD(\mD^\prime) = \{e_1, e_2, \ldots, e_k\}$. For each bootstrapped dataset $\mD^\prime_b$, we now compute the Kendall-Tau ranking similarity between $E_\mD(\mD^\prime)$ and $E_\mD(\mD^\prime_b)$, defined as:
\begin{equation}
    K(E_1,E_2) = \frac{C-D}{\binom{n}{2}}
\end{equation}
where $C$ is the number of pairs in $E_1, E_2$ that have the same relative order in both rankings and $D$ is the converse. A value of $K(E_1,E_2)$ close to $1$ means that the ranking order is preserved, implying that the identified influential examples in $\mD$ are consistently recognized across different realizations of the dataset $\mD^\prime$. Figure \ref{fig:kendall_tau_heloc} shows that the Kendall-Tau similarity values remain high (i.e. $\sim 0.9975$) across multiple bootstrap iterations for both Adult and HELOC datasets, indicating a robust identification of influential examples.

\subsection{Influential Example Explanations: Effect of Rashomon Set}
We evaluate how our influential example explanations change as we change the size of the Rashomon set, i.e. the set of all near-optimal models. We can change the size of the Rashomon set using the $\epsilon$ parameter, which sets the maximum allowed sub-optimality gap for a model to enter the set. For a given epsilon value, we use the Rashomon Importance Distribution method from \cite{donnelly2023rashomon} to generate intrinsic feature importances for $\mD$ and $\mD^\prime$. We then see how the resulting set of influential examples in either dataset changes as we vary the epsilon parameter. We also examine how long it takes to generate the explanation as we vary the size of the Rashomon set.

\begin{figure}[H]
    \centering
    \includegraphics[width=0.9\linewidth]{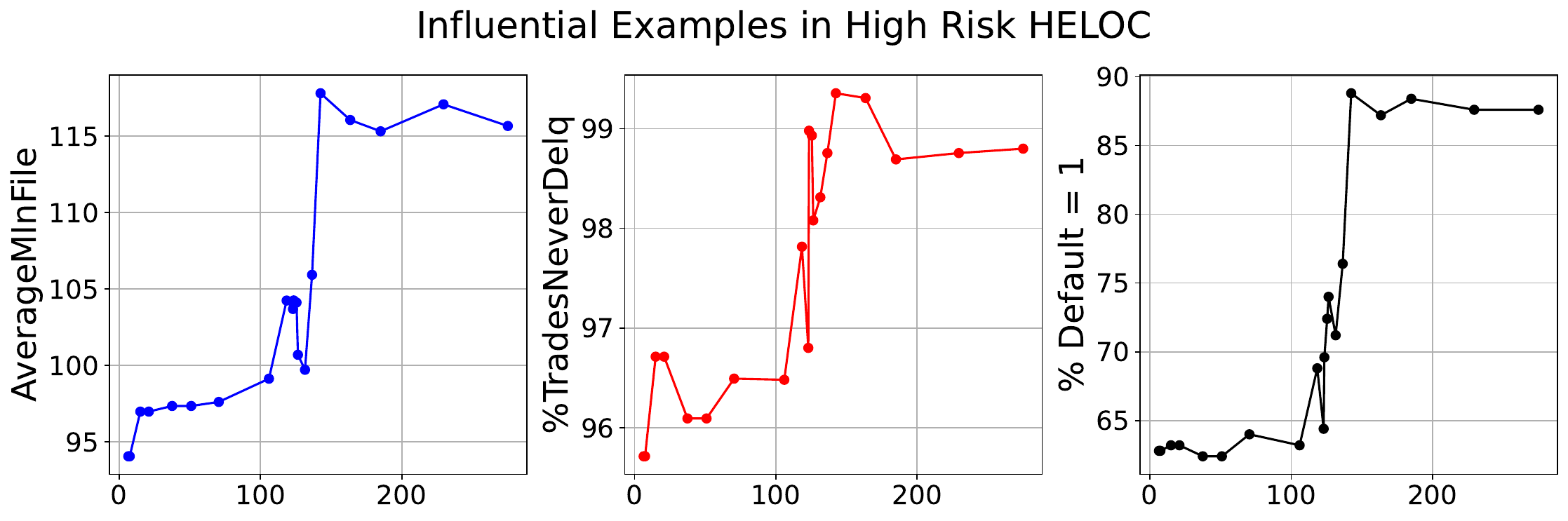}
    \includegraphics[width=0.91\linewidth]{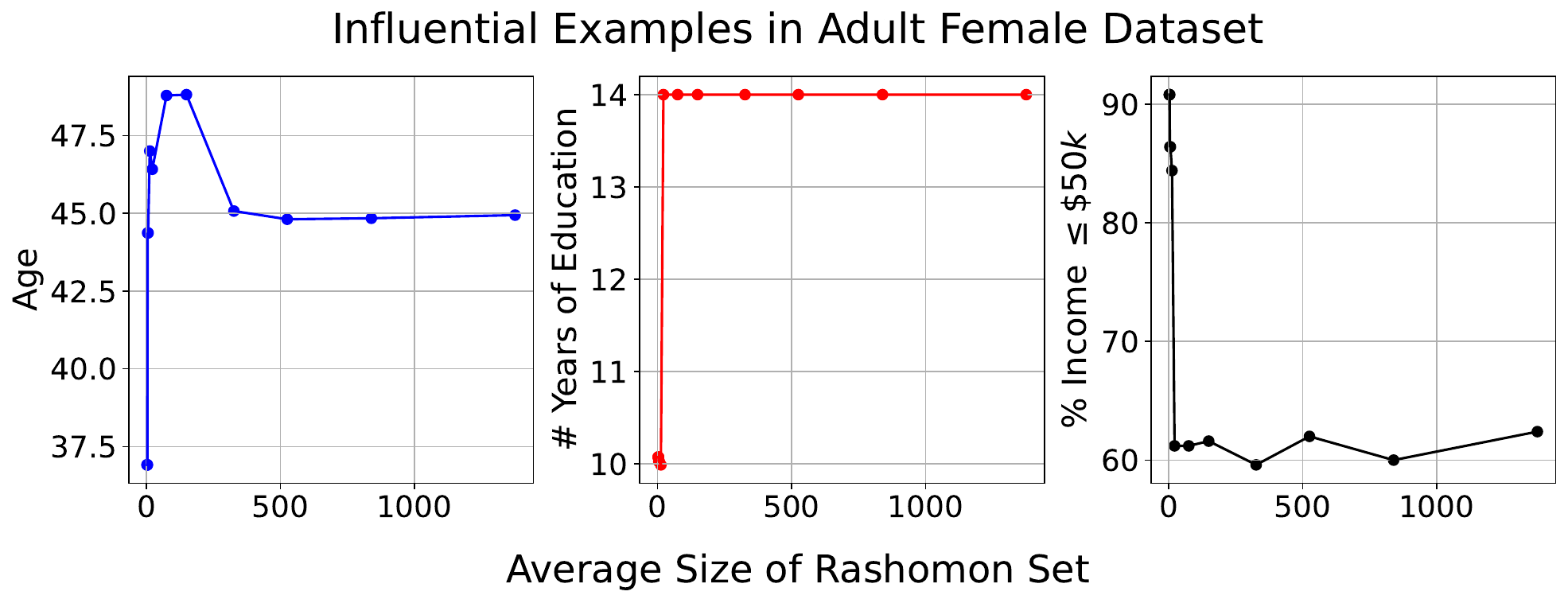}
    \caption{Properties of the 50 most influential examples in $\mD$ that explain differences between $\mD$ and $\mD^\prime$. As the Rashomon set size increases, the feature importances output by RID will stabilize, resulting in a corresponding stabilization in the influential examples output by our explanation.}
    \label{fig:heloc_influential_examples_rset_size}
\end{figure}

\begin{figure}[H]
    \centering
    \includegraphics[width=0.44\linewidth]{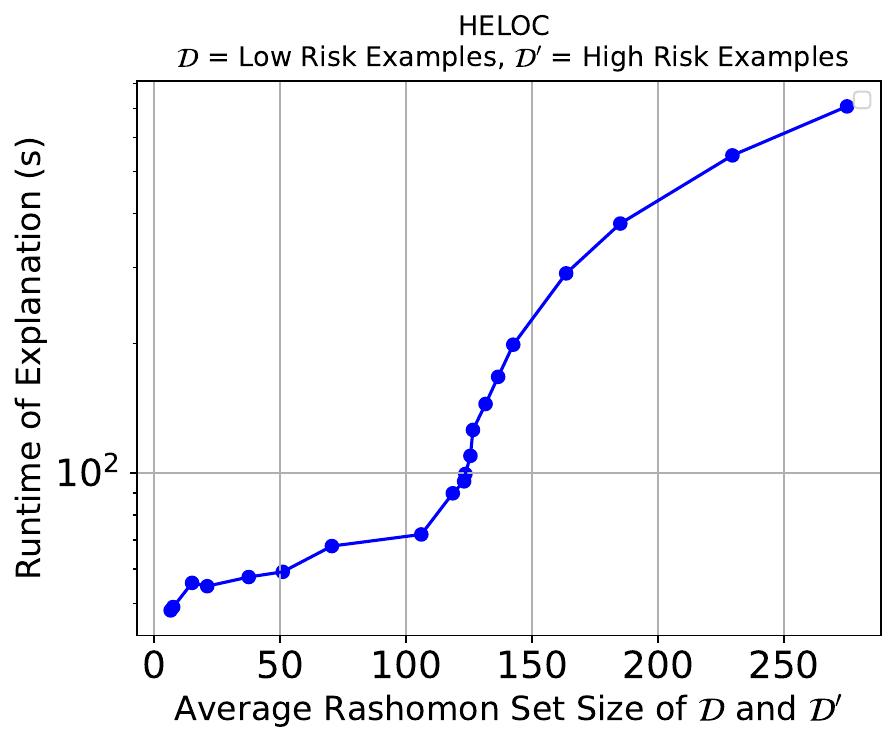}
    \includegraphics[width=0.45\linewidth]{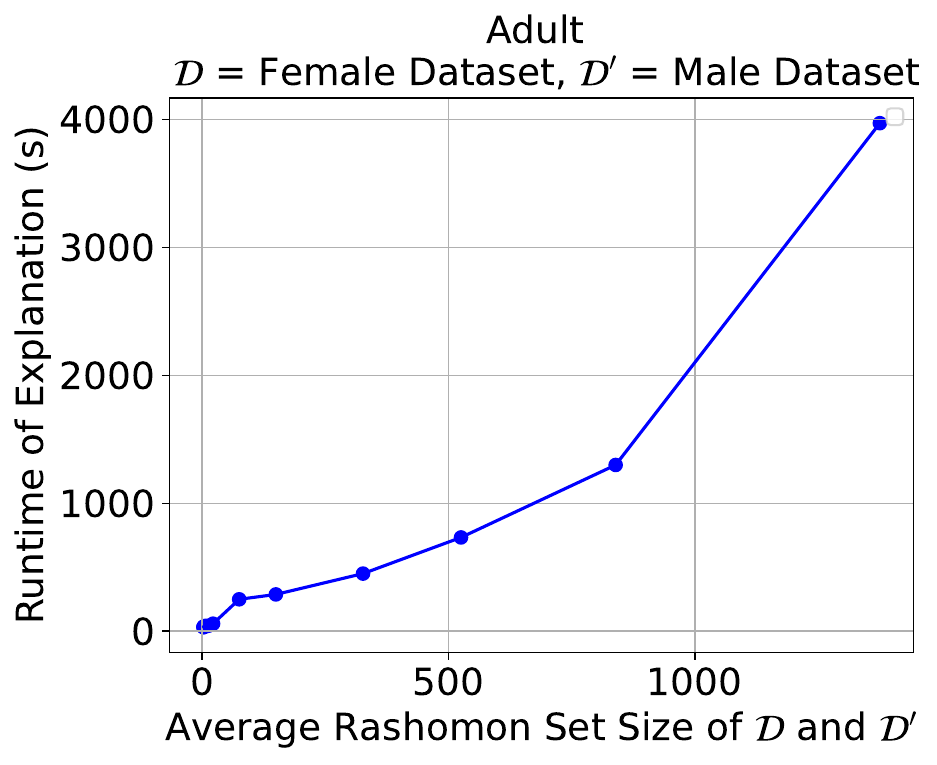}
    \caption{Size of the Rashomon set used to generate feature importances vs explanation runtime.}
    \label{fig:heloc_rset_size_runtime}
\end{figure}

\section{Illustrating a Failure Mode of \citet{towards_explain}}
\label{sec:failure}
\subsection{Methodology}

We now illustrate an example where the distribution shift explanation of \citet{towards_explain} is incoherent, but our prototype-based explanations are able to accurately capture dataset differences. In particular, we simulate the mean shift of a mixture of Gaussians -- Case $1$ below is the same setup as \citet{towards_explain}. Because the cluster centres are shifted by the same amount, we call the cluster centres of $X$ and $Y$ \textit{paired}.
\paragraph{Case 1:}
\begin{itemize}
    \item We first sample $k = 6$ points uniformly from the circumference of a circle of given radius $r_x = 10$. These points are the cluster centres of a mixture of Gaussians with isotropic covariances and equal cluster proportions. We sample $60$ points around each cluster centre. Call this resulting dataset of $360$ points $X$. 
    \item We then repeat this procedure, but with a circle radius of $r_y = 20$. Call this resulting dataset $Y$. 
\end{itemize}
\paragraph{Case 2:}
\begin{itemize}
    \item Dataset $X$ is generated in the same manner with the same parameters as above.
    \item We then sample cluster centres with a circle radius of $r_y = 20$. The resulting mixture of Gaussians still has isotropic covariances, but we now change the cluster proportions. To generate cluster proportions, we sample a $6$ dimensional probability vector from a Dirichlet distribution with parameters $\alpha_1 = ... = \alpha_6 = 1$. This is a distribution with the following pdf:
    \begin{equation}
        f(x_1,...x_6;\boldsymbol{\alpha}) = \frac{1}{\beta(\boldsymbol{\alpha})} \prod_{i=1}^6 x_i^{\alpha_i-1}
    \end{equation}
    with $\beta(\boldsymbol{\alpha})$ is the Beta function serving as the normalizing constant. This setup ensures that the generated vector satisfies $\sum_{i=1}^6 x_i = 1$. We then sample from the mixture of Gaussians according to these cluster proportions, generating $360$ points. Call this dataset $Y$.
\end{itemize}
\paragraph{Explanation computed by \citet{towards_explain}:} The explanation finds $k$ clusters in $X$ and illustrates how they shift from $X$ to $Y$ using an optimal transport formulation. The final output is the shifted cluster center and the map from $X$ to $Y$ - this is illustrated in the Figures below. 
% % % % % % %
\subsection{Case 1: Our explanation and \citet{towards_explain} is coherent}
\begin{figure}[H]
    \centering
    \includegraphics[width=0.495\textwidth]{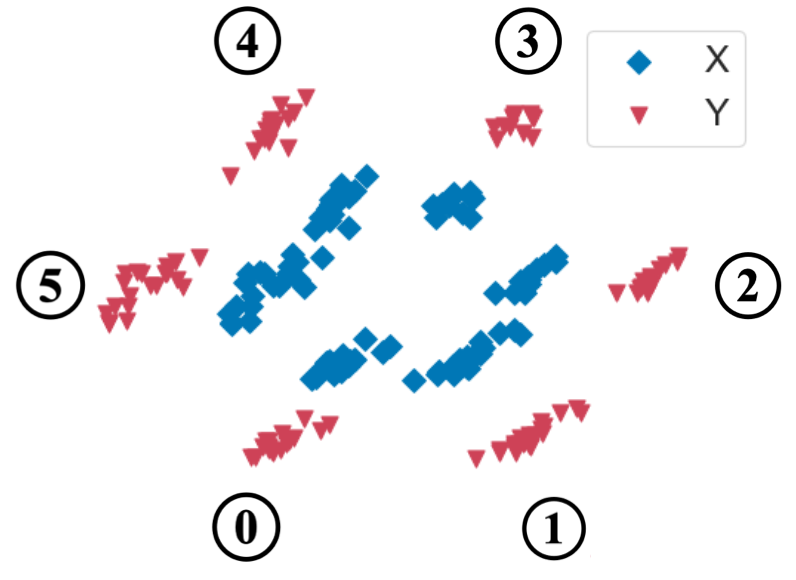}
    \includegraphics[width=0.495\textwidth]{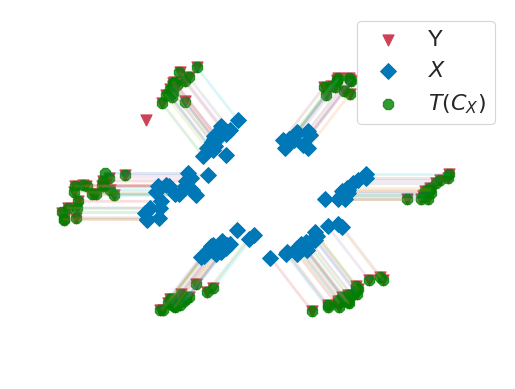}\\
        \includegraphics[width=0.7\textwidth]{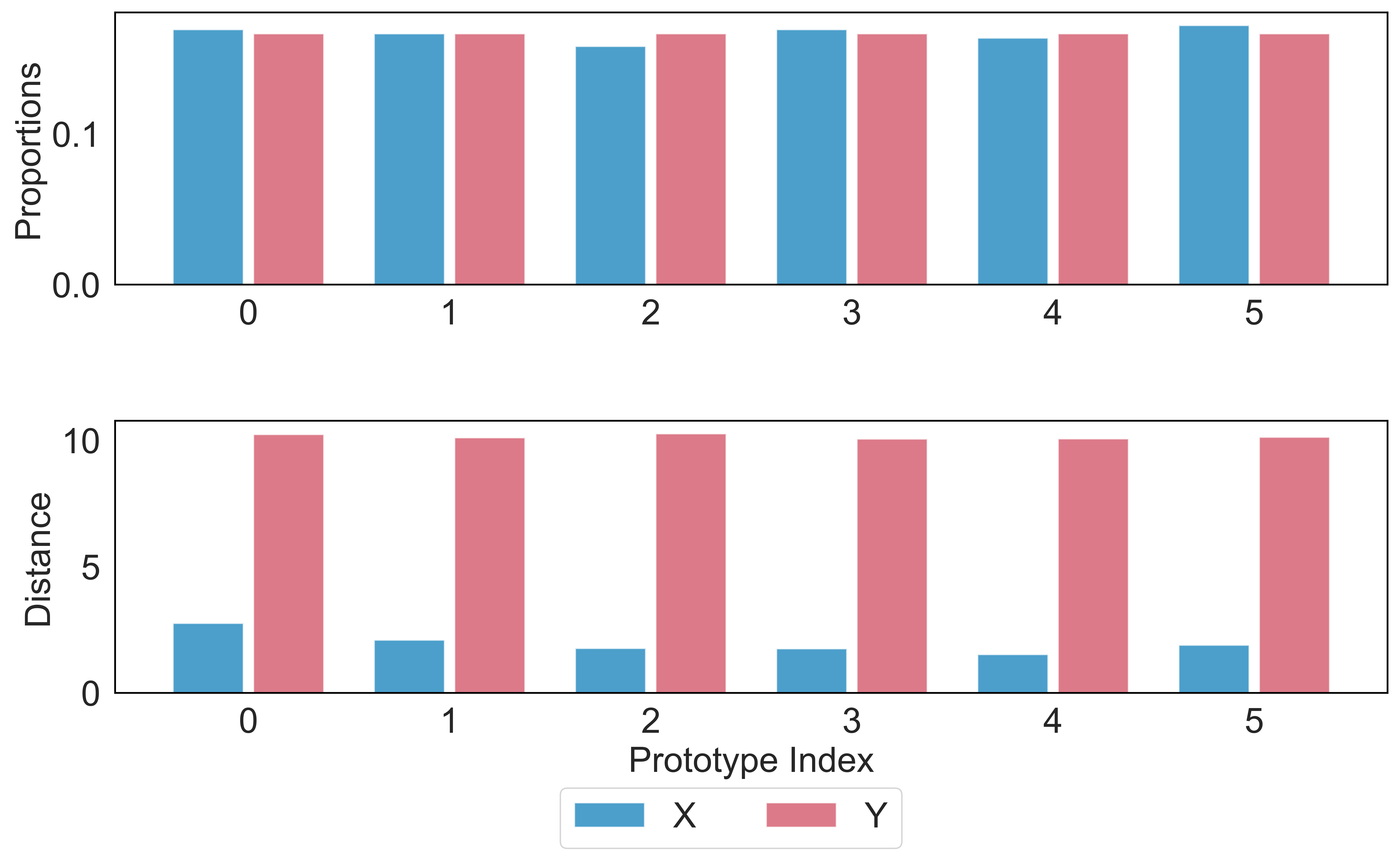}
    \caption{\textbf{Top left} - The two datasets $X$ and $Y$ have clusters with equal proportions. The paired cluster centres are labelled appropriately. \\
    \textbf{Top right} - The explanation map showing the shift of each cluster in $X$ to the corresponding cluster in $Y$ computed by \citet{towards_explain}. Their method almost perfectly maps each point from $X$ to $Y$ and the explanation is coherent.\\
    \textbf{Top bar plot} - (Our Explanation) Proportion of points in the neighborhood of each prototype (cluster centre) in $X$ for both datasets. The proportions are the same for all clusters.\\ 
    \textbf{Bottom bar plot} - (Our Explanation) Average distance of points to the closest prototype (cluster centre) in $X$ for both datasets. This is constantly high for dataset $Y$, suggesting a constant shift in clusters has occurred. }
    \label{fig:adult_mog_explanation}
\end{figure}
\subsection{Case 2: Our explanation is coherent but \citet{towards_explain} is not}
\begin{figure}[H]
    \centering
    \includegraphics[width=0.45\textwidth]{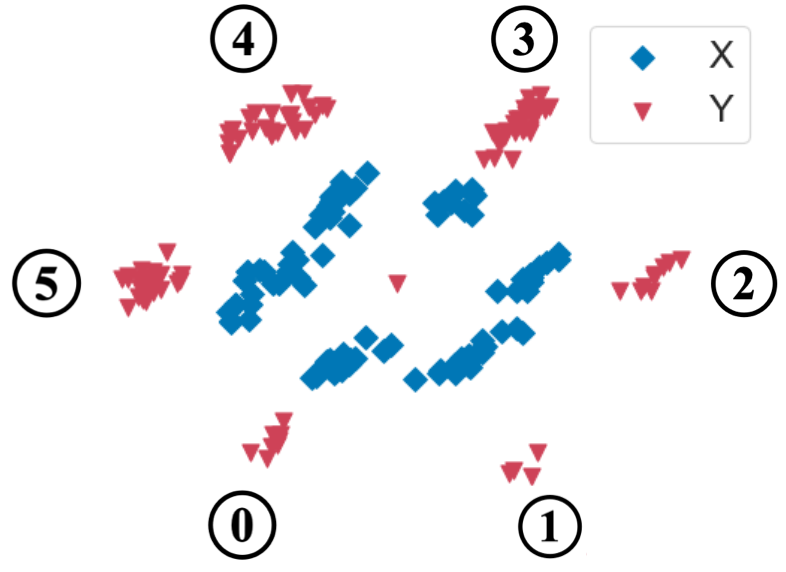}
    \includegraphics[width=0.45\textwidth]{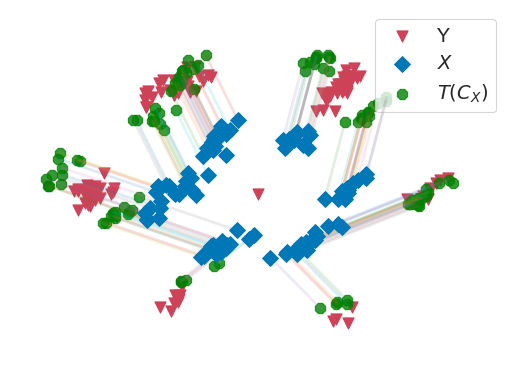}
    \includegraphics[width=0.7\textwidth]{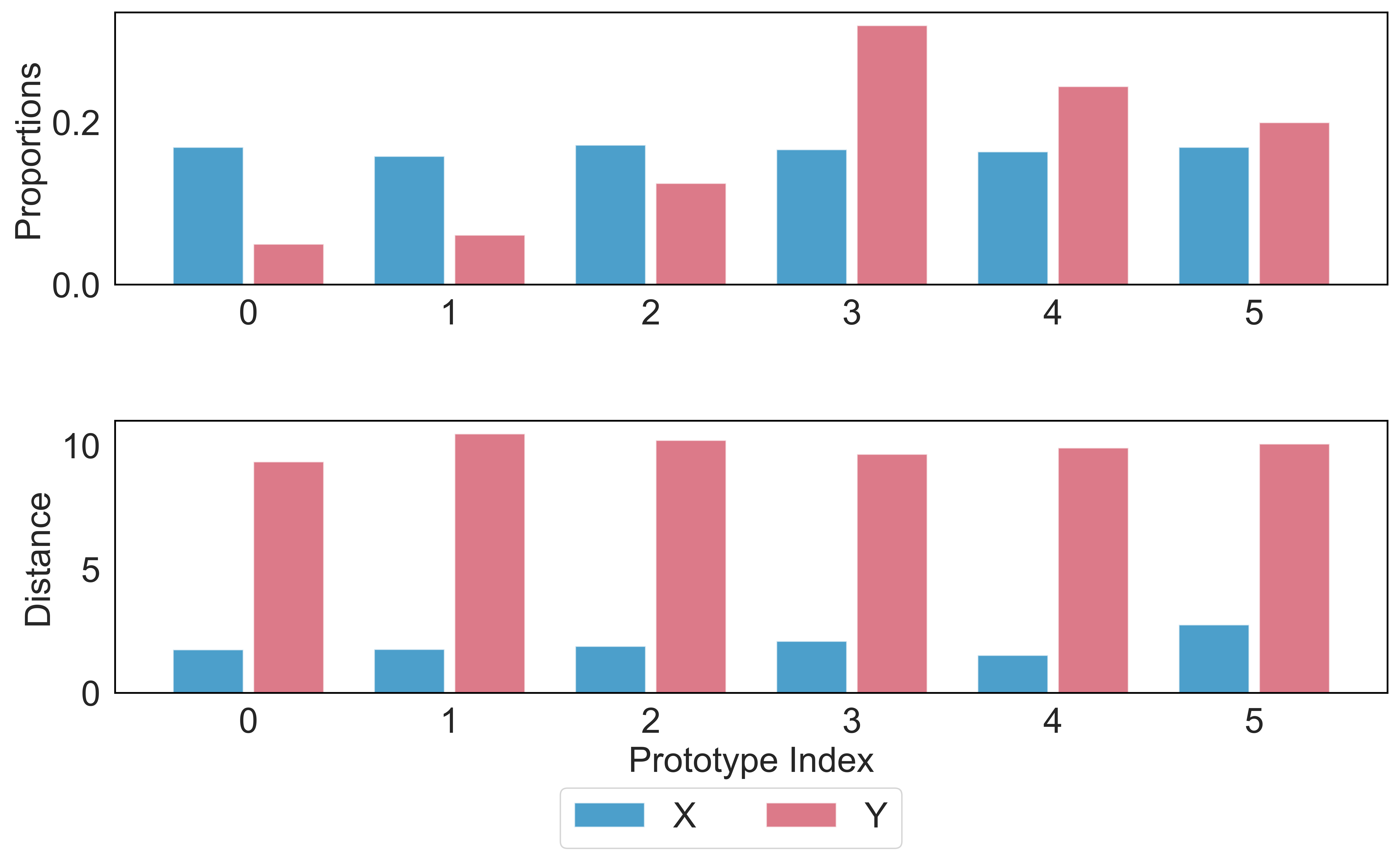} \\
    \caption{\textbf{Top left} - The two datasets $X$ and $Y$ have the same cluster centres as in Figure~\ref{fig:adult_mog_explanation}, but $Y$ has unequal cluster proportions sampled using the methodology above. The paired cluster centers are labeled appropriately. \\
    \textbf{Top right} - The explanation from \citet{towards_explain} is now unable to map the shift in clusters from $X$ to $Y$ as there is no longer a no one-to-one mapping between points in the clusters. This renders the explanation uninterpretable.\\
    \textbf{Top bar plot} - (Our Explanation) Proportion of points in the neighborhood of each prototype (cluster centre) in $X$ for both datasets. The proportions are now different across all clusters and can be visually validated from the diagram, which is what we want to see from our explanation.\\
    \textbf{Bottom bar plot} - (Our Explanation) Average distance of points closest prototype (cluster centre) in $X$ for both datasets. This is constantly high for dataset $Y$, suggesting only a constant shift in cluster centres has occurred.\\
    Our prototype-based explanations can, therefore, quantify exactly how cluster proportions and distances have changed.
    }
    \label{fig:prototype_explanations_mog}
\end{figure}

\section{Text Data Difference Analysis with ``\textit{Exact Counts}''}
\label{sec:llmexp}
In this section, we briefly consider the idea of explaining two text datasets. We consider this to be a significantly more challenging (and general) problem compared to other modalities, owing to various pathologies exhibited by text data. In particular, prototype based methods fail here because it is hard to define an underlying latent space where distances between textual examples make sense (e.g., what does the `distance' between a sentence prototype and another example mean?). Thus, pending future work, we provide a simple method to summarize two text datasets by extracting interpretable attributes from the datasets. Meanwhile, we consider that the approach in \cite{elazar2023whats} to be a useful starting point for understanding a given text dataset (and indeed drawing points of comparison with other similar datasets).
\paragraph{Dataset $\mD$ and $\mD^\prime$} We first define datasets $\mD$ and $\mD^\prime$. In this example, we used the HC-3-English dataset \citep{hc3} as a sample dataset for text dataset difference explanations. HC-3-English contains answers to questions from both humans ($\mD$) and ChatGPT (2022 version) ($\mD^\prime$) collected from several domains and tasks, including open-domain question-answering (QA), financial, medical, legal, and psychological areas. Human and ChatGPT answers are not compared pairwise (i.e., we did not assume a one-to-one mapping between the answers). Instead, we analyzed the differences between all human answers against all ChatGPT answers. We hypothesize that the datasets can be meaningfully compared using the following attributes: 
  
\begin{enumerate}[itemsep=0mm]
    \item Have consistent writing structure
    \item Use formal language
    \item Have a neutral tone
    \item Show subjective opinion
    \item Use of technical references
\end{enumerate}
These attributes can fairly reliably be determined by querying language models. \\

For each input $X$ in $\mD$ and $\mD^\prime$, we query GPT-3.5 Turbo using the following prompt:
\FloatBarrier
\begin{tcolorbox}
\textbf{Prompt:}\begin{verbatim}
Analyze the following text by answering the following questions including: \end{verbatim} 
$\{A_1, A_2, ... A_n\}$
\begin{verbatim}
For each question provide "YES OR NO" answer only.
\end{verbatim}
$X_i$
\end{tcolorbox}
Where each $A_i$ is an attribute.

\begin{figure}[hbt!]
    \begin{subfigure}[b]{0.5\textwidth}
        \centering
        \includegraphics[width=\textwidth]{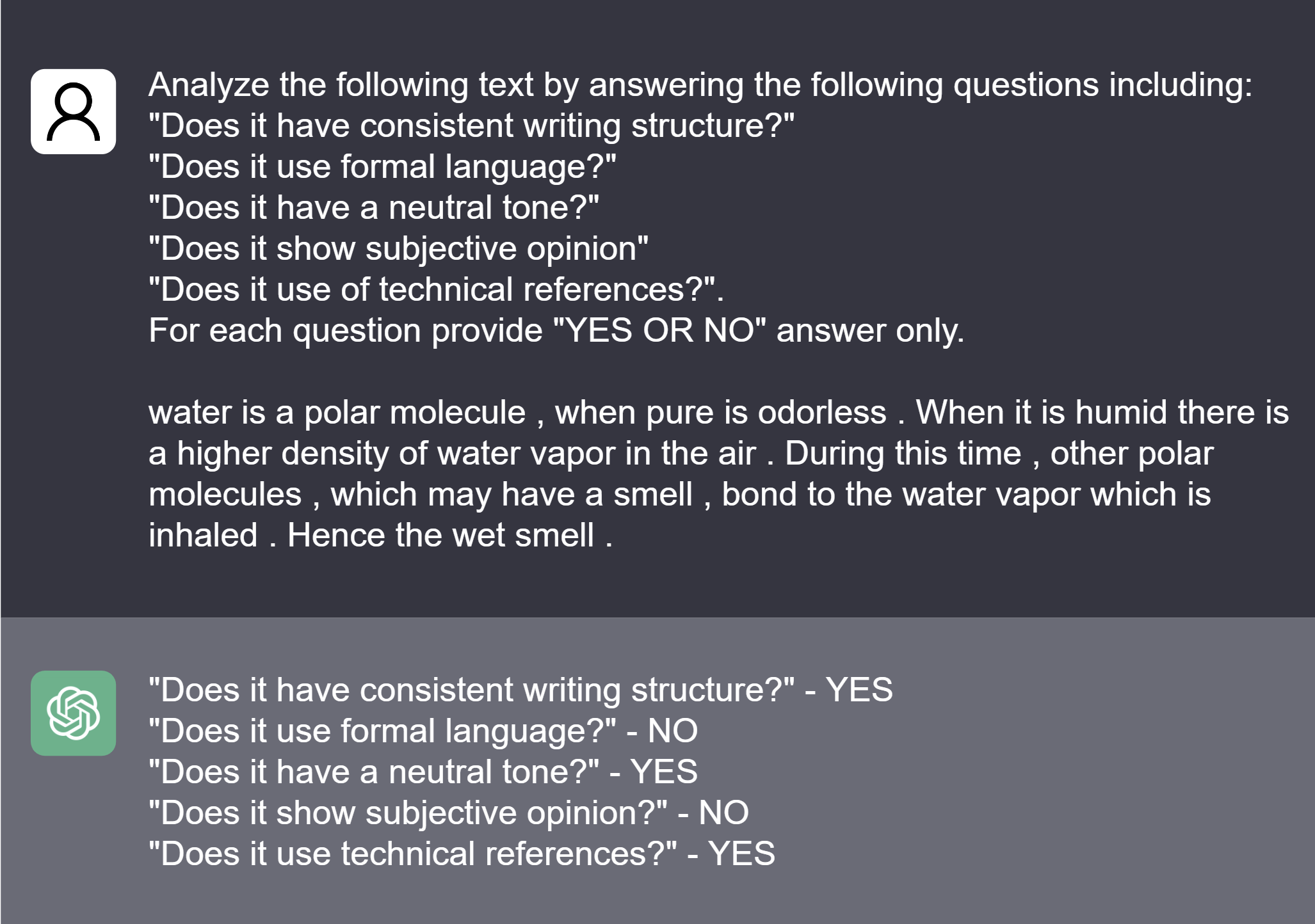}
        \caption{Human answer analysis.}
        \label{fig:ucla_nonoise}
    \end{subfigure}
    \begin{subfigure}[b]{0.5\textwidth}
        \centering
        \includegraphics[width=\textwidth]{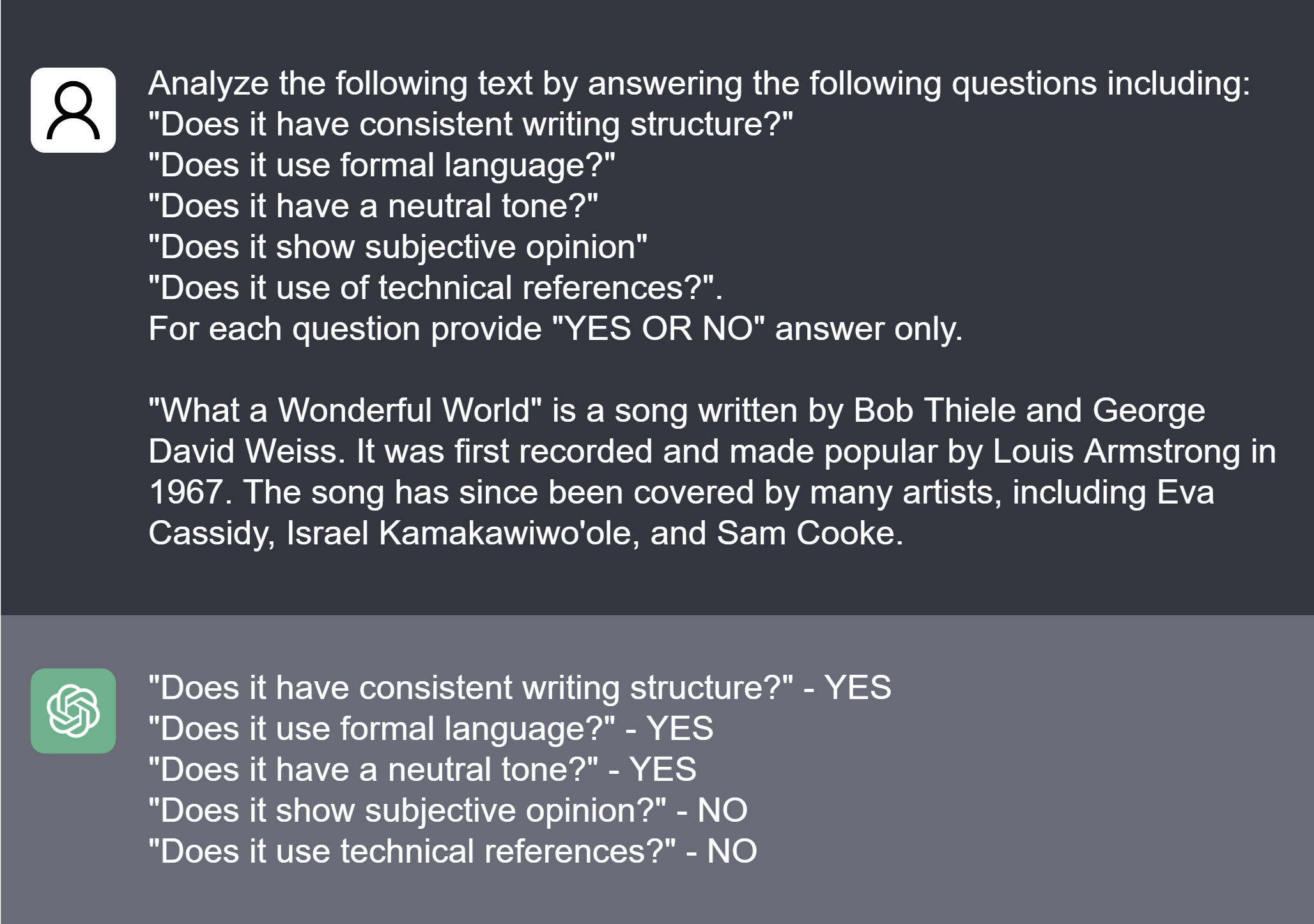}
        \caption{ChatGPT answer analysis.}
    \end{subfigure}
    \caption{Human and ChatGPT answer analysis example}
    \label{fig:llmexample}
\end{figure}

\paragraph{Forming the explanation} Using the prompt template defined in above, we show two query-answer examples in Figure~\ref{fig:llmexample}. After answers were collected from ChatGPT for all samples in the dataset, we formed explanations based on the results shown in Table~\ref{tab:llmanalysis} -- on these question-answering tasks, among the attributes analyzed, humans and ChatGPT answers mainly differ in \textit{use of formal language usage}, \textit{use of subjective opinion} and \textit{writing structure consistency}. That is, we found that humans tend to show more subjectivity in their answers and tend to use more informal language than ChatGPT. Also, humans are much less likely to use consistent writing structure throughout each writing sample. 

A simple logistic regression model was able to reach 85\% accuracy in predicting whether the text is produced by humans or ChatGPT with the five attributes we collected for each text sample. This indicates that the selected attributes are often able to correctly identify differentiating attributes between the two datasets. 
  
\begin{table}[hbt!]
\begin{adjustbox}{max width=\textwidth}
\begin{tabular}{rccccc}
\hline
        & \begin{tabular}[c]{@{}c@{}}\textbf{Have consistent} \\ \textbf{writing structure}\end{tabular} & \begin{tabular}[c]{@{}c@{}}\textbf{Use} \\ \textbf{formal language}\end{tabular} & \begin{tabular}[c]{@{}c@{}}\textbf{Have} \\ \textbf{a neutral tone}\end{tabular} & \begin{tabular}[c]{@{}c@{}}\textbf{Show} \\ \textbf{subjective opinion}\end{tabular} & \begin{tabular}[c]{@{}c@{}}\textbf{Use of }\\ \textbf{technical references}\end{tabular} \\ \hline
Human   & 59.4\%                                                                       & 22.6\%                                                         & 77.5\%                                                         & 26.9\%                                                             & 35.0\%                                                                 \\ \hline
ChatGPT & 99.2\%                                                                       & 93.2\%                                                         & 99.6\%                                                         & 0.9\%                                                              & 35.2\%                                                                 \\ \hline

ChatGPT \\ (Informal) & 69.9\%                                                                       & 10.1\%                                                         & 68.9\%                                                         & 17.5\%                                                              & 18.3\%                                                                 \\ \hline

\end{tabular}
\end{adjustbox}
\caption{Human and ChatGPT answer analysis results.}
\label{tab:llmanalysis}
\end{table}

\paragraph{Improving $\mD^\prime$} Following the conclusions that we have reached, to improve the similarity between $\mD$ and $\mD^\prime$, we aimed to make each sample in $\mD^\prime$ appear more informal and colloquial. Here we again leveraged GPT-3.5 Turbo to ``humanize'' the ChatGPT text samples. For each text $X_i$ in $\mD^\prime$, we used the following prompt:

\begin{tcolorbox}
\textbf{Prompt:}\begin{verbatim}
Make the following context sound less formal, paraphrase using some colloquial
language. 
\end{verbatim}
\end{tcolorbox}

The attribute calculated for the improved $\mD^\prime$ is shown in the last row of Table~\ref{tab:llmanalysis} (ChatGPT (informal)). Using the attributes, on the same test split, the same logistic regression classifier was only able to achieve 45\% accuracy. The decrease in the model's predictive performance indicates an increase in similarity between the text samples in $\mD$ and $\mD^\prime$.

\paragraph{Results} By utilizing these human-interpretable attributes and a straightforward counting method, we effectively identified key differences between human- and machine-generated text. Building on this understanding, we quantitatively improved the similarity between the two by addressing these differences. Without this dataset-level explanation, achieving this result would have been both time-consuming and subject to considerable uncertainty.

\end{document}